\PassOptionsToPackage{table,dvipsnames}{xcolor}
\documentclass[]{style}
\usepackage{microtype}
\usepackage{amsfonts}
\usepackage{xcolor}
\usepackage{graphicx}
\usepackage{booktabs}
\usepackage{wrapfig}
\usepackage{float}
\usepackage{amsmath}
\usepackage{amsthm}
\usepackage{subcaption}
\usepackage{makecell}
\theoremstyle{plain}

\theoremstyle{definition}
\theoremstyle{remark}

\usepackage{enumitem}
\usepackage{pifont}
\usepackage[edges]{forest}
\usetikzlibrary{arrows.meta}
\usepackage{tikz}
\usepackage{graphicx}     
\usepackage{booktabs}    
\usepackage{enumitem}   
\usepackage[most]{tcolorbox}
\usepackage{fontawesome5}
\usepackage{booktabs} 
\usepackage{url}
\usepackage{wrapfig}
\usepackage{longtable}
\usepackage{tabularx}
\usepackage{longtable}
\usetikzlibrary{
  positioning,
  calc,
  shapes.symbols,
  shapes.geometric,
  shapes.misc
}
\usepackage[T1]{fontenc}

\usepackage{svg}

\usepackage{xltabular}
\usepackage{booktabs}
\usepackage{caption}
\usepackage{longtable}
\usepackage{fontawesome5}

\newcommand{\githublink}[1]{\faIcon{github}\,\href{#1}{Code}}
\newcommand{\paperlink}[1]{\faIcon{book}\,\href{#1}{Paper}}

\newcommand\blfootnote[1]{%
  \begingroup
  \renewcommand\thefootnote{}\footnote{#1}%
  \addtocounter{footnote}{-1}%
  \endgroup
}

\title{{\raisebox{-1.ex}{\includegraphics[height=2em]{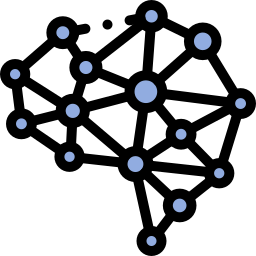}}}\; The Latent Space: Foundation, Evolution, Mechanism, Ability, and Outlook
}
\author{Xinlei Yu$^{*,\star}$}
\author{Zhangquan Chen$^{*}$}
\author{Yongbo He$^{*}$}
\author{Tianyu Fu$^{*}$}
\author{Guanting Dong$^{*}$}
\author{Cheng Yang$^{*}$}
\author{Chengming Xu$^{*}$}
\author{Yue Ma$^{*}$}
\author{Xiaobin Hu$^{*,\dagger}$}

\author{Zhe Cao}
\author{Jie Xu}
\author{Guibin Zhang}
\author{Jiale Tao}
\author{Jiayi Zhang}
\author{Siyuan Ma}
\author{Kaituo Feng}
\author{Haojie Huang}
\author{Youxing Li}
\author{Ronghao Chen}
\author{Huacan Wang}
\author{Chenglin Wu}
\author{Zikun Su}
\author{Xiaogang Xu}
\author{Kelu Yao}

\author{Kun Wang}
\author{Chen Gao}
\author{Yue Liao}
\author{Ruqi Huang}
\author{Tao Jin}
\author{Zhucun Xue}
\author{Cheng Tan$^{\dagger}$}
\author{Jiangning Zhang$^{\dagger}$}
\author{Wenqi Ren}
\author{Yanwei Fu}
\author{Yong Liu}
\author{Yu Wang}
\author{Xiangyu Yue$^{\dagger}$}
\author{Yu-Gang Jiang$^{\dagger}$}
\author{Shuicheng Yan$^{\dagger}$}

\affiliation{
\vspace{0.5em}
{\small$*$\;Core Contributors\;\;$\dagger$\;Core Supervisors \;\;$\star$\;Organizer}

{\small National University of Singapore, Fudan University, Tsinghua University, Zhejiang University, Shanghai Artificial Intelligence Laboratory, Renmin University of China, The Chinese University of Hong Kong, The Hong Kong University of Science and Technology, DeepWisdom, Nanjing University, Shanghai Jiatong University, Nanyang Technological University, Tencent Hunyuan, QuantaAlpha, Beijing University of Posts and Telecommunications, Zhejiang Lab, University of Chinese Academy of Sciences, Hong Kong University of Science and Technology (Guangzhou), Sun Yat-sen University}

\vspace{1em}

{\url{https://github.com/YU-deep/Awesome-Latent-Space}}
}

\newcommand{\xhdr}[1]{\noindent{\bfseries #1.}}

\usepackage{todonotes}
\vspace{1em}

\abstract{
\begin{abstract}
LLatent space is rapidly emerging as a native substrate for language-based models. While modern systems are still commonly understood through explicit token-level generation, an increasing body of work shows that many critical internal processes are more naturally carried out in continuous latent space than in human-readable verbal traces. This shift is driven by the structural limitations of explicit-space computation, including linguistic redundancy, discretization bottlenecks, sequential inefficiency, and semantic loss. As a result, research on latent space has expanded from early latent reasoning into a broader landscape spanning planning, modeling, perception, memory, collaboration, and embodiment. However, the literature remains fragmented across mechanisms, modalities, and tasks, lacking a unified perspective on how latent space is defined, classified, and studied.

This survey aims to provide a unified and up-to-date landscape of latent space in language-based models. We organize the survey into five sequential perspectives: \textbf{Foundation}, \textbf{Evolution}, \textbf{Mechanism}, \textbf{Ability}, and \textbf{Outlook}. We begin by delineating the scope of latent space, distinguishing it from explicit or verbal space and from the latent spaces commonly studied in generative visual models. We then trace the field's evolution from early exploratory efforts to the current large-scale expansion. To organize the technical landscape, we examine existing work through the complementary lenses of mechanism and ability. From the perspective of \textbf{Mechanism}, we identify four major lines of development: \textbf{Architecture}, \textbf{Representation}, \textbf{Computation}, and \textbf{Optimization}. From the perspective of \textbf{Ability}, we show how latent space supports a broad capability spectrum spanning \textbf{Reasoning}, \textbf{Planning}, \textbf{Modeling}, \textbf{Perception}, \textbf{Memory}, \textbf{Collaboration}, and \textbf{Embodiment}. Beyond consolidation, we discuss the key open challenges, and outline promising directions for future research. We hope this survey serves not only as a reference for existing work, but also as a foundation for understanding latent space as a general computational and systems paradigm for next-generation intelligence.
\end{abstract}

}

\begin{document}

\maketitle

\blfootnote{We will continue to update this paper and its associated GitHub repository, and we warmly welcome contributions from the community. Should you identify any related work that has not been included, please feel free to notify us via Github or email: xinleiyu88@gmail.com, czq23@mails.tsinghua.edu.cn, ben0xiaobin0hu1@nus.edu.sg.}

\clearpage
\tableofcontents
\clearpage

\begin{figure*}[t]
  \centering
    \includegraphics[width=1\linewidth]{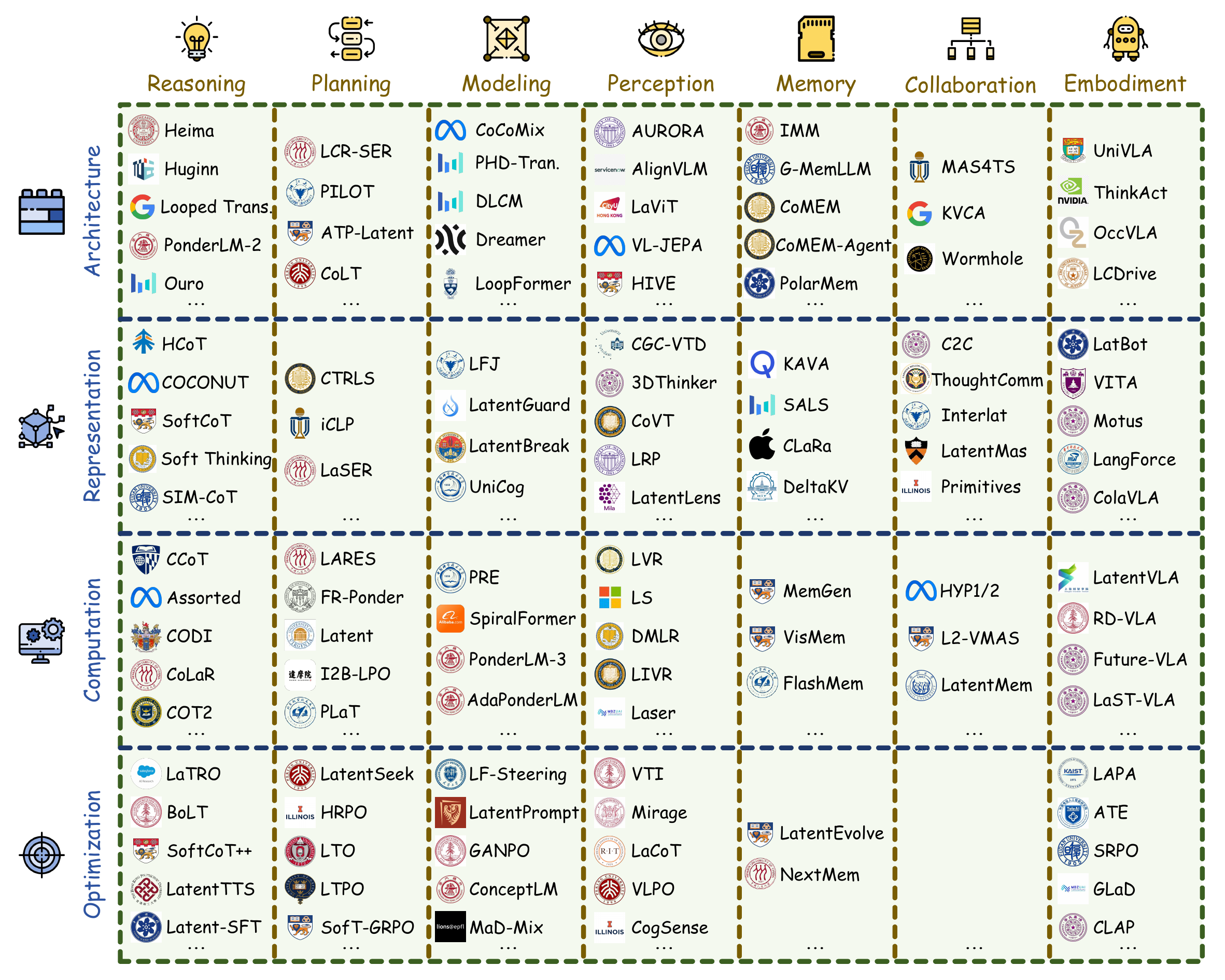}
    \caption{Overview of the latent space methods classified by two axes: four main \textbf{Mechanisms} (Section~\ref{sec:mechanism}) and seven key \textbf{Abilities} (Section~\ref{sec:ability}). Within our classification system, a single method may be affiliated with one or more mechanisms and capabilities. For the visualization in this figure, we adopt the most appropriate classification for each method; a comprehensive elaboration of these categories will be presented in the main text.}
    \label{fig:grid}
\end{figure*}

\section{Introduction}
\label{sec:intro}
Recent advances in language-based models, including Large Language Models (LLMs), Vision-Language Models (VLMs), Vision-Language-Action models (VLAs), and agentic systems built on language backbones, are still commonly understood through explicit token-level generation, where inputs, outputs, and even intermediate reasoning are expressed in human-readable form~\cite{vaswani2017attention,wei2022chain,yao2023react}. Yet this token-centric framing is increasingly insufficient~\cite{hao2024training,perception2025bigverdi,jihoon2025llm}. Because computation in such models fundamentally unfolds through continuous activations, latent space is increasingly being reconceived not as a hidden implementation detail, but as a machine-native substrate, such as reasoning~\cite{hao2024training,zhu2025soft,xu2025softcot}, perception~\cite{perception2025bigverdi,ahmed2025alignvlm}, memory~\cite{zhang2025memgen,yu2025vismem}, communication~\cite{zheng2025thought,zou2025latent}, and action~\cite{huang2025thinkact,ni2025swiftvla}. This shift is driven in part by the structural limitations of explicit space, its redundancy, discretization bottleneck, sequential decoding cost, and potential loss of fine-grained information, especially in complex, multimodal, or long-horizon settings. By contrast, latent-space computation offers a more continuous, compact, and expressive medium that can support higher-fidelity representations and more flexible allocation of computation. 

\begin{figure*}[!t]
  \centering
    \includegraphics[width=0.85\linewidth]{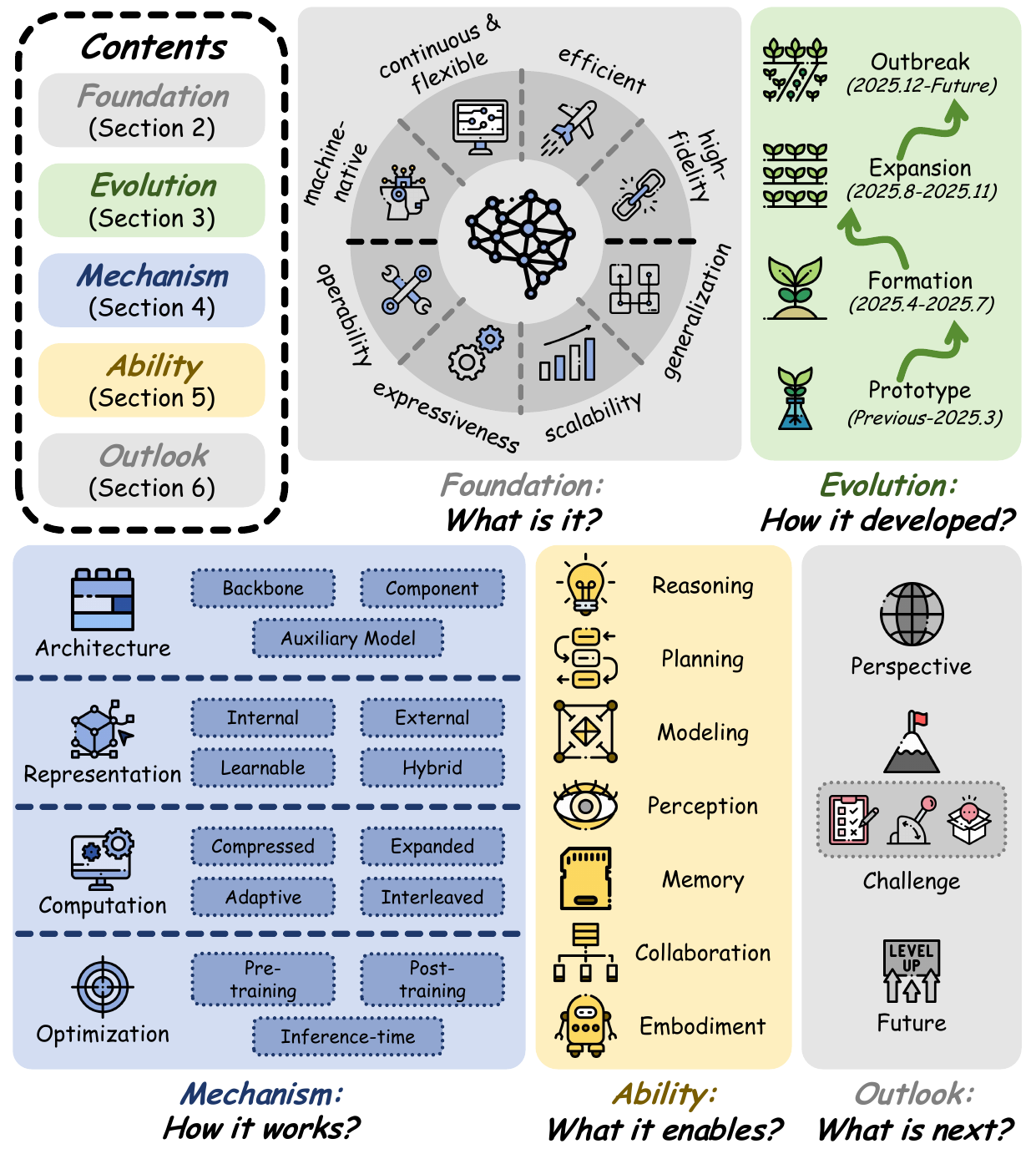}
    \caption{Outline of the survey, including five sections and sequential questions: \textcolor{secgray}{\textbf{Foundation}}\textbf{: What is Latent Space?} (Section~\ref{sec:foundation_sec2}), \textcolor{secgreen}{\textbf{Evolution}}\textbf{: How Did Latent Space Develop?} (Section~\ref{sec:evolution}), \textcolor{secblue}{\textbf{Mechanism}}\textbf{: How Does Latent Space Work?} (Section~\ref{sec:mechanism}), \textcolor{secyellow}{\textbf{Ability}}\textbf{: What Does Latent Space Enable?} (Section~\ref{sec:ability}), and \textcolor{secgray}{\textbf{Outlook}}\textbf{: What is Next?} (Section~\ref{sec:outlook})}
    \label{fig:intro}
\end{figure*}

Research has therefore moved far beyond the initial framing of latent space as latent reasoning alone.What began as an attempt to internalize chain-of-thought into continuous states has rapidly expanded into a broader systems paradigm spanning new modalities, new interaction settings, and new design choices~\cite{reasoning2025chen,survey2025zhu,implicit2025li}. However, this growth has also fragmented the literature in at least three ways: by application object, \textit{e.g.}, latent reasoning, visual understanding, and embodied action; by mechanism, \textit{e.g.}, architecture design, representation choice, computation pattern, and optimization strategy; and by scenario, spanning text, vision, multi-agent systems, and embodied environments. Existing reviews mainly focus on latent reasoning or implicit reasoning as a reasoning-specific phenomenon. What remains missing is a \textit{unified perspective} that treats latent space as a broader computational and systems paradigm across modalities, paradigms, mechanisms, and capabilities.

To address this gap, we organize the survey around five sequential questions that move from conceptual grounding to future outlook, as illustrated in Figure~\ref{fig:intro}: \textbf{What is latent space? How did it develop? How does it work? What does it enable? What is next?} These questions define the macro-level narrative of the paper: \textcolor{secgray}{\textbf{Foundation}} (Section~\ref{sec:foundation_sec2}) delineates the \textbf{concept of latent space} and clarifies its relation to explicit space and to latent space in generative visual models; \textcolor{secgreen}{\textbf{Evolution}} (Section~\ref{sec:evolution}) traces how the field \textbf{progressed} from prototype exploration to rapid outbreak; \textcolor{secblue}{\textbf{Mechanism}} (Section~\ref{sec:mechanism}) explains how latent space is \textbf{instantiated and operationalized}; \textcolor{secyellow}{\textbf{Ability}} (Section~\ref{sec:ability}) examines what latent computation enables across \textbf{downstream capabilities}; and \textcolor{secgray}{\textbf{Outlook}} (Section~\ref{sec:outlook}) synthesizes \textbf{open challenges and future directions}. This five-question narrative is intentionally sequential. This organization allows us to preserve a clear narrative while also comparing diverse methods through shared principles and  capability outcomes, rather than through task-specific labels alone.

Within this sequential narrative, our technical synthesis is anchored by a two-dimensional taxonomy shown in Figure~\ref{fig:grid}. The first axis, \textbf{Mechanism}, asks how latent space is built and used, and covers four major lines: \textbf{Architecture}, \textbf{Representation}, \textbf{Computation}, and \textbf{Optimization}. The second axis, \textbf{Ability}, asks what latent space enables, and covers seven major capability domains: \textbf{Reasoning}, \textbf{Planning}, \textbf{Modeling}, \textbf{Perception}, \textbf{Memory}, \textbf{Collaboration}, and \textbf{Embodiment}. This design lets us preserve a clear survey-level storyline while also comparing diverse methods through shared design principles and shared capability outcomes, rather than through task-specific labels alone.

\vspace{5pt}
\begin{tcolorbox}[
  colback=secgray!5,
  colframe=secgray!50,
  colbacktitle=secgray!50,
  coltitle=black,
  title={\textbf{Contributions}},
  boxrule=5pt,
  arc=5pt,
  drop shadow,
  parbox=false,
  before skip=5pt,
  after skip=10pt,
  left=5pt,   
  right=20pt,
]
\begin{itemize}[leftmargin=1.4em,itemsep=0.25em,topsep=0.35em]
    \item We clarify the conceptual scope of latent space in language-based models, distinguishing it from explicit or verbal space and from the latent spaces commonly studied in generative visual models.
    \item We provide a unified review of how latent space has evolved from early latent reasoning into a broader multimodal and systems-level research paradigm.
    \item We introduce a two-dimensional taxonomy across \textbf{Mechanism} and \textbf{Ability}, offering a common framework for organizing otherwise fragmented methods and applications.
    \item We provide a comprehensive collection of resources, including illustrative figures, structured tables, accessible links, and repositories, to facilitate further research and community engagement.
\end{itemize}
\end{tcolorbox}

\section{\textcolor{secgray}{Foundation:} What is Latent Space?}
\label{sec:foundation_sec2}

As an emerging paradigm, the exploration of the latent space within language models has exhibited immense potential and vast room for further development. Accordingly, this section provides a preliminary elaboration covering its formal definition and comparisons with existing paradigms.

\subsection{Concept}
In language-based models, \textit{i.e.}, classic autoregressive models, encompassing LLMs, VLMs, VLAs, and derivative systems with language backbone, the entire operational process unfolds explicitly within the \textbf{explicit space}, or the \textbf{verbal space}~\cite{vaswani2017attention}.
The most immediate and externally observable domain is: the discrete space of linguistic symbols in which inputs and outputs are expressed. Formally, this space is defined by a vocabulary that specifies the language model's next-token predictions. In this space, language is represented as overt verbalized units, \textit{i.e.}, subword-level tokens~\cite{sennrich2016neural}, that are directly interpretable by humans. The training objective of a language model is typically formulated in this explicit space, since the model learns to assign probabilities to symbol sequences and to predict the next token given a textual prefix.

However, a language-based model does not operate solely on discrete symbols. To compute over language, it first maps tokens into internal continuous representations and transforms them through multiple layers of nonlinear computation. This gives rise to what is commonly called the \textbf{latent space}: the continuous, learned representational space in which the model encodes and manipulates information that is not explicitly verbalized at the token level. More precisely, the latent space of a language model can be understood as the family of hidden-state spaces, within which contextual, semantic, syntactic, and relational features of an input are jointly represented in this space. A token sequence in the explicit space is thus mapped to a trajectory of latent space, and these latent states are subsequently projected back into the verbal space to yield a probability distribution over possible next tokens. 
Furthermore, this latent space could be expanded into a unified space beyond language that maps modality-specific inputs into continuous internal representations.
We further provide a formal definition and formulation in Section~\ref{sec:mechanism}.

\subsection{Comparison with Explicit Space}
\begin{figure*}[t]
  \centering
    \includegraphics[width=1\linewidth]{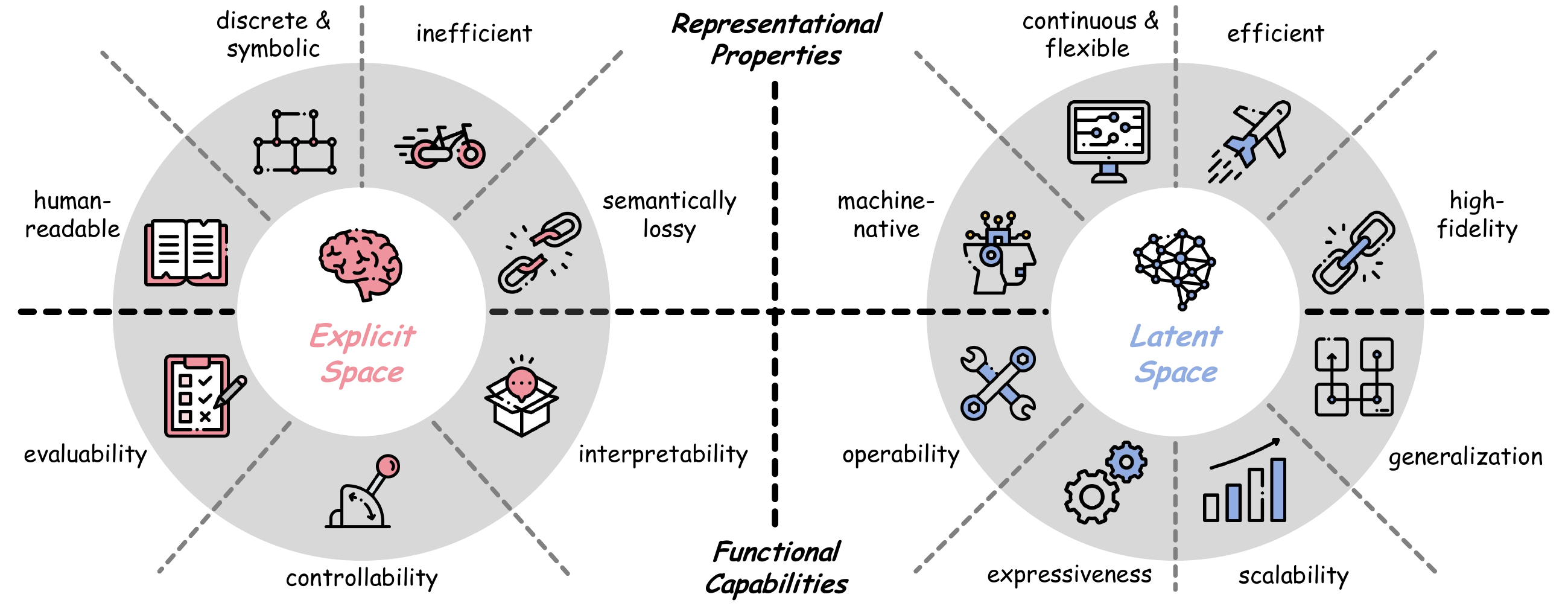}
    \caption{Comparison of the \textbf{explicit space} and \textbf{latent space} of the language models, including their representational properties and functional capabilities.}
    \label{fig:com}
\end{figure*}

To further elucidate the unique characteristics of the latent space, we conduct a comparative analysis with the traditional explicit space, or verbal space,  across several critical dimensions~\cite{xi2025rise,humze2025comprehensive,wu2025evolutionary}. As demonstrated in Figure~\ref{fig:com}, this comparison highlights a paradigm shift in the representational properties and functional capabilities of the two language models.

\subsubsection{Representational Properties} 
The two distinct paradigms of language models, \textit{i.e.}, explicit space and latent space, exhibit fundamental differences in their representational forms, information processing modes, and practical performances. The latent space uses a \textbf{Machine-native} vectorized representation that is \textbf{Continuous}, \textbf{Flexible}, \textbf{Efficient}, and capable of preserving \textbf{High-fidelity} semantic information.

\xhdr{Human-readable \textit{v.s.} Machine-native}
In the explicit space, every state is expressed as a sequence of natural language tokens drawn from a finite vocabulary. This endows the generated trajectories with direct human readability and verifiability~\cite{wei2022chain,yao2023react}. 

By contrast, the latent space is a machine-native representational paradigm that lacks direct human legibility. Its core representations are high-dimensional real-valued vectors, where individual dimensions do not have a straightforward correspondence to human-interpretable semantic, structural, or perceptual features~\cite{xu2025formal,jos2025beyond}. 
Tailored for the intrinsic operational logic of language models, the latent representation and operations can be processed by the model, without the need for transformations of human-readable signals, reducing computational redundancy from additional encoding/decoding overhead~\cite{hao2024training,saunshi2025reasoning,zhang2025memgen}.

\xhdr{Discrete \& Symbolic \textit{v.s.} Continuous \& Flexible}
Explicit-space representations are redundant, discrete, and symbolic. For instance, a chain-of-thought trace for a complex reasoning problem may span thousands of tokens~\cite{wei2022chain}. Each token is a discrete symbol whose meaning is grounded in linguistic convention, and the relationship to other tokens is expressed through positional and grammatical structure~\cite{xu2024symbol,lanham2023measuring}. This decoding pattern is, in part, a redundancy and inflexibility: the majority of tokens serve textual coherence rather than substantive semantic contents~\cite{xia2025tokenskip,dong2025rethinking}. They ensure grammaticality, maintain topic continuity, and satisfy the autoregressive constraint that each token be a plausible continuation of the preceding sequence and be mapped into a finite discrete token, obligations that have no intrinsic connection to the logical structure of the autoregressive generation~\cite{tang2024language,clara2021revisiting,xu2025formal}.

Conversely, latent space exhibits inherent continuity and flexibility. It encapsulates core semantic information in a continuous form, discarding superficial linguistic redundancies and symbolic mapping~\cite{hao2024training,shen2025codi,zhu2025reasoning}. Liberated from the constraints of discrete tokenization and autoregressive linguistic conventions, this nature confers upon latent space distinct advantages in bolstering representational capacity: in reasoning models, it elevates explicit reasoning trajectories onto continuous manifolds~\cite{zhuang2026geometric,su2025token}; in visual scenarios, it enables smoother multimodal operations (\textit{e.g.}, fusion, alignment, and interaction) within latent spaces marked by a narrower modality gap~\cite{sun2025latent,yu2025vismem,wu2026lavit}; in collaboration, it inherently serves as a more fitting medium for information storage and transmission~\cite{fu2025cache,zou2025latent}.
In essence, the continuous and flexible properties enable language-based models to prioritize the intrinsic semantic essence, unlocking potential across diverse tasks.

\xhdr{Inefficient \textit{v.s.} Efficient}
Conventional autoregressive generation operates entirely within an explicit discrete space, and this paradigm inherently introduces at least three fundamental inefficiencies~\cite{xia2024unlocking,zhang2024uncovering}:
First, linguistic redundancy: as the main medium of explicit space, natural language introduces unavoidable structural and semantic redundancy, further reducing the efficiency~\cite{tang2024language,lanham2023measuring};
Second, representational transformation inefficiency: the model is forced to transfer representations through a narrow, explicit channel at every generation step, including intermediate steps~\cite{pfau2024let,liang2026do}. This mandatory conversion introduces unnecessary and high representational conversion costs.
Third, sequential decoding overhead: discrete tokenization locks generation into a strictly sequential pipeline: each token requires a full model forward pass and vocabulary-wide probability calculation. This sequential paradigm not only has low computational efficiency but also increases computational burden~\cite{leviathan2023fast,arip2025your}; 
In contrast, latent space methods bypass these inefficiencies, \textit{e.g.}, removing mandatory representation conversion in inter-agent communication~\cite{zou2025latent,fu2025cache,zheng2025thought}, and efficient recurrent or looped computation patterns~\cite{jonas2025scaling,saunshi2025reasoning}.

\xhdr{Semantically-lossy \textit{v.s.} High-fidelity}
Explicit-space representations are prone to semantic loss. When a model externalizes its internal continuous state as a token sequence, the mapping from latent activations to discrete symbols imposes a quantization bottleneck: a finite vocabulary and the combinatorial constraints of natural language delimit what can be expressed~\cite{wei2022outlier,alec2021learning}. As a result, fine-grained uncertainty, intermediate computational traces, cross-modal alignments, and other structures that are difficult to render in language may be compressed, distorted, or discarded. In this sense, natural language constitutes a semantically lossy transformation of the underlying computational state.

Latent-space representations, by contrast, preserve information with higher fidelity. By avoiding discretization and linguistic rendering, latent variables can carry rich, continuous information between computational steps, including content that is inexpressible in natural language and representations that naturally support multimodal structure. This perspective motivates a growing line of work directly in latent space, such as continuous thoughts~\cite{hao2024training,shen2025codi,zhu2025reasoning}, latent memory~\cite{zhang2025memgen,yu2025vismem,hou2026flashmem}, and latent visual reasoning~\cite{li2025latent,yang2025machine}, \textit{etc.}

\subsubsection{Functional Capabilities}
The latent space, as a machine-native representational space, possesses multiple key functional capabilities that distinguish it from the explicit space, including \textbf{Operability}, \textbf{Expressiveness}, \textbf{Scalability}, \textbf{Generalization}, as well as Evaluability, controllability, and interpretability, which collectively underpin its utility in various advanced computational and representational tasks.

\xhdr{Operability} 
As a machine-native representational space, the operability of latent space characterizes its utility as a structured, differentiable manifold that serves as several internal computational substrates. This operability seamlessly enables direct calculations (such as concatenation, linear combination, \textit{etc.}) and also advanced operations, \textit{e.g.}, controllable semantic steering~\cite{yang2025lfsteering,kazama2025geosteer}, active intervention~\cite{zhu2025soft,li2025latentseek,fu2025tah}, iterative interleaving~\cite{liu2025reason,chen2025reasoning}, and visual latent thinking~\cite{zhang2025latentsketchpad,wang2025monet,sun2025latent}. 
On the contrary, the discrete tokens in the explicit space are inherently non-differentiable and lack the support of a continuous, structured manifold, rendering the aforementioned fine-grained, advanced semantic operations infeasible and permitting only limited, indirect token-level operations.

\xhdr{Expressiveness} 
It serves as a core capacity for internalizing and manipulating complex, high-dimensional, and even non-linguistic information. In contrast to natural language, which is constrained by a discrete vocabulary and grammatical conventions, representations within latent space provide a substantially richer representational substrate. In principle, it can express the representation including but not limited to the whole explicit space, enabling efficient communication~\cite{zou2025latent,zheng2025thought,yang2025internal}, visual perception~\cite{yang2025machine,perception2025bigverdi,chen2025think}, embodied action planning~\cite{bu2025univla,huang2025thinkact}, latent memory formation~\cite{zhang2025memgen,yu2025vismem}, \textit{etc}.

\xhdr{Scalability} 
It follows naturally from the compactness and parallelizability of vectorized representations. Latent-space approaches are therefore well-positioned to benefit from continued scaling of longer reasoning trajectories~\cite{liu2025deliberation,ruan2025reasoning,tan2025think,wei2025simcot}, deeper agent interaction turns~\cite{zheng2025thought,yu2026dual}, and test-time compute~\cite{you2025parallel,jonas2025scaling,zhang2025latentevolve}.

\xhdr{Generalization}
It further bolsters the ability to generalize effectively to inputs distinct from those encountered during training, which capture abstract semantic structures rather than superficial linguistic patterns. By embedding abstract semantic concepts into a latent space, models gain improved cross-domain transfer and zero-shot generalization, enabling the transfer of learned abstractions to previously unseen tasks and domains. Transfer learning~\cite{zhan2025l2vcot,wezuke2025transfer}, and cross-domain robustness~\cite{zhang2025memgen,shi2025bridging,deng2025latent,yu2025vismem} have all been empirically shown to benefit from such informative latent representations.

\xhdr{Evaluability \& Controllability \& Interpretability} 
This denotes the ability of humans or automated systems to evaluate, control, observe, interpret, and audit the autoregressive generation process. For explicit generation, the resulting generation traces are evaluable, controllable, and interpretable, as each intermediate step is represented in a human-readable format~\cite{lightman2024verify,korbak2025chain,ball2025human}. In principle, systems built upon explicit reasoning mechanisms can integrate human-aligned verification or automated consistency checks. In contrast, machine-native latent space representations make it inherently difficult for humans to perform granular, direct evaluation, control, and interpretation of the generation process, posing potential challenges to model evaluability, controllability, and interpretability (Section~\ref{sec:challenges}).

\subsection{Comparison with Generative Visual Models}
The latent space of generative visual models is derived from a VAE-style framework~\cite{kingma2014auto}, which learns a probabilistic mapping from high-dimensional observations to a compact continuous representation. Subsequent work introduced discrete latent codes via VQ-VAE~\cite{van2017neural}, and the decisive step toward scalable synthesis was taken by latent diffusion models~\cite{rombach2022high}, which demonstrated that diffusion processes operating in a perceptually compressed latent space could achieve both computational efficiency and high sample fidelity. Extensions to video generation, including VideoLDM~\cite{andreas2023stable} and related architectures, further organized this space along a spatiotemporal axis, encoding appearance and motion as jointly structured representations.

Despite the shared reliance on learned continuous representations, the latent spaces of generative visual models and large language models differ fundamentally in their geometric structure, representational organization, and conditioning regimes.

\xhdr{Training Objective} The latent space of generative visual models is explicitly shaped by a reconstruction objective, which anchors the learned geometry to the statistical structure of the visual signal~\cite{arash2021score,andreas2023stable}. This produces a relatively smooth, locally Euclidean manifold in which linear interpolation between encoded points yields perceptually coherent intermediates, and in which distances carry interpretable perceptual meaning. Rather than a reconstruction objective, language model hidden states are organized by a predictive criterion, next-token prediction in autoregressive models~\cite{tom2020language},  with no explicit constraint on the geometry of the space.

\xhdr{Structure} Latent space of visual generative models maintains an explicit spatiotemporal structure: the latent tensor of an image is a spatial grid of patch-level codes~\cite{esser2021taming,peebles2023scarable}, and that of a video extends this grid along the temporal axis, with motion treated as a first-class representational element. This structural regularity reflects the continuous, compositional nature of visual scenes, in which spatial proximity and temporal coherence are strong inductive priors. On the contrary, the latent space of a language model focuses on the linguistic semantics and is devoid of spatial topology or physical dynamics.

\xhdr{Controllability} Precise control over visual generation is exercised through dedicated architectural pathways integrated into the model itself~\cite{zhang2023adding}. These signals include pose sequences, depth maps, segmentation masks, and reference images, affording fine-grained, spatially localized control over the generated output by conditioning applied directly within the representational substrate.

\section{\textcolor{secgreen}{Evolution:} How Did Latent Space Develop? }
\label{sec:evolution}

\begin{figure*}[t]
  \centering
    \includegraphics[width=1\linewidth]{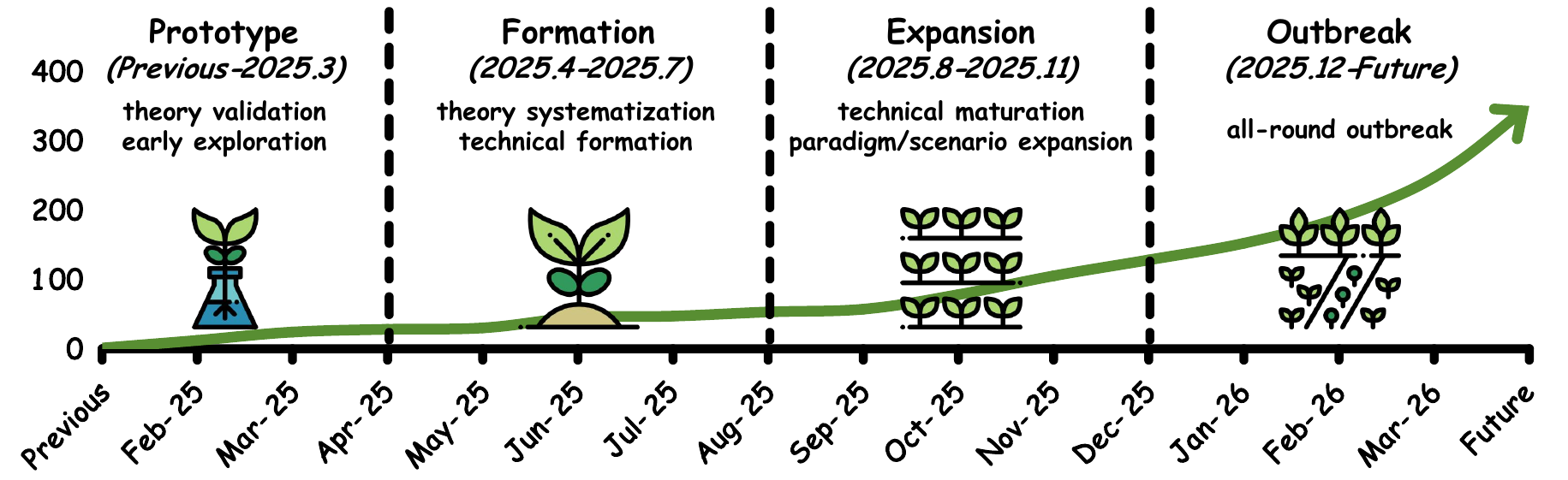}
    \caption{Timeline of representative works in the evolution of latent space research, organized into four developmental stages: \textbf{Prototype} (Section~\ref{sec:embryonic}), \textbf{Formation} (Section~\ref{sec:foundation}), \textbf{Expansion} (Section~\ref{sec:expansion}), and \textbf{Outbreak} (Section~\ref{sec:outbreak}) stages, where the horizontal axis denotes the month, and vertical axis indicates the number of the latent-level works.}
    \label{fig:timeline}
\end{figure*}

As large language models have demonstrated outstanding performance across a range of natural language understanding and reasoning tasks, the exploration of their latent spaces has undergone a remarkably rapid and transformative evolution.
In this section, we trace the developmental trajectory of this emerging paradigm, organizing the literature into four chronologically and thematically coherent stages (Figure~\ref{fig:timeline}): \textbf{(1) Prototype} (Previous -- Mar 2025), where pioneering works first demonstrated the feasibility of moving reasoning from discrete token sequences into continuous latent representations; \textbf{(2) Formation} (Apr -- Jul 2025), where theoretical foundations were established and systematic evaluations emerged, with research primarily centered on textual latent reasoning and initial explorations into multimodal settings; \textbf{(3) Expansion} (Aug -- Nov 2025), where the field rapidly diversified into visual tasks, embodied action, multi-agent collaboration, and other application paths; and \textbf{(4) Outbreak} (Dec 2025 -- Present), where explosive growth drove architectural innovation, advanced optimization, rigorous theoretical analysis, and cross-modal integration toward maturity.

\subsection{Prototype}
\label{sec:embryonic}
The prototype stage marks the genesis of latent space reasoning, where researchers first question whether large language models must articulate every intermediate reasoning step in natural language. In this stage, \textbf{Theory Validation} and \textbf{Early Exploration} begin to use continuous representations as an alternative substrate.

\xhdr{Theory Validation}
Before the first latent systems appeared, several works laid essential groundwork by revealing that reasoning-like behavior is already encoded in the internal representations of language models.
Represented by HCoT~\cite{liu2024expediting}, early precursors show that a full CoT trace can be compressed into a compact special-token representation via contrastive semantic alignment, suggesting that much of the information in explicit CoT is redundant for the model itself.
Meanwhile, \citet{zhang2024uncovering} discovers latent thinking vectors by extracting steering vectors from model activations, and demonstrates that injecting these vectors at inference time can elicit CoT-like reasoning without any fine-tuning or explicit prompting.
From a theoretical perspective, \citet{hu2024understanding} offers a Hopfieldian interpretation, framing reasoning as transitions across representation spaces with grounding in cognitive neuroscience.
Furthermore, Latent Space Chain-of-Embedding~\cite{wang2025latent} demonstrates that LLMs can perform self-evaluation through latent embeddings rather than explicit verbal outputs.
Together, these precursors establish a critical insight: the capacities of language models are not fundamentally bound to discrete token sequences, but are already substantially encoded in their latent spaces.

\xhdr{Early Exploration}
Building on these insights, the initial stage produced a series of high-impact works that gradually shaped the initial direction of latent space reasoning.
COCONUT~\cite{hao2024training} proposes the first complete framework for reasoning in continuous latent space, feeding the last hidden state back as the next input embedding to form a loop of continuous thoughts that bypasses the discrete vocabulary bottleneck. Its key finding is that continuous thought vectors can encode superpositions of multiple potential next steps, enabling emergent breadth-first search in latent space.
Along a similar compression-based direction, CCoT~\cite{cheng2024compressed} introduces contemplation tokens that compress explicit reasoning chains into dense latent form. At the same time, \citet{liu2025deliberation} augments the KV cache with latent embeddings via an offline coprocessor, keeping the decoder entirely frozen.
Both demonstrate that latent representation can be injected into existing models through parameter-efficient adaptation without inference latency overhead.

Subsequently, Huginn~\cite{jonas2025scaling} proposes using recurrent depth to scale test-time compute in latent space, iterating a shared transformer block a variable number of times to perform all reasoning implicitly without specialized reasoning data.
SoftCoT~\cite{xu2025softcot} provides the first plug-in approach to latent space, projecting instance-specific soft thought tokens into the frozen backbone's representation space to avoid catastrophic forgetting.

Taken together, the prototype stage establishes the feasibility of latent-space reasoning, but it also exposes its first major bottleneck: the field still lacks a systematic account of \emph{why} latent reasoning works, \emph{when} it outperforms explicit CoT, and \emph{how} it should be evaluated beyond isolated proof-of-concept demonstrations. In other words, the central challenge at this stage is not capability alone, but interpretability, formalization, and comparability. These limitations directly motivate the next stage, where the community moves from scattered prototypes toward theoretical systematization, benchmark construction, and more principled technical design.

\subsection{Formation}
\label{sec:foundation}
The formation stage advances from the prototype-stage insights toward building a more practically meaningful research program. This period is primarily centered on textual latent reasoning and is characterized by two advances: \textbf{Theoretical Systematization} that explains why it works, \textbf{Technical Formation} that works on how well it works, and initial explorations of extension into multimodal and embodied settings.

\xhdr{Theory Systematization}
The most transformative contribution of this stage is the theoretical understanding of continuous thought mechanisms. These theoretical works provide formal guarantees for the expressiveness and computational advantages of latent space, grounding the empirical successes of the prototype stage in rigorous mathematical frameworks.
\citet{zhu2025reasoning} (Reasoning by Superposition) provide the first formal complexity analysis, proving that continuous thought vectors acting as superposition states can encode multiple search frontiers simultaneously, offering a rigorous explanation for COCONUT's empirically observed behavior.
CoT2~\cite{gozeten2025continuous} further quantifies the relationship between parallelism and embedding dimension and introduces continuous supervision and reinforcement learning for continuous thought optimization.
\citet{saunshi2025reasoning} prove that looped transformers with latent iterations can express strictly more complex computations than standard transformers, establishing theoretical separation results for recurrent-depth architectures.

\xhdr{Technical Formation}
The formation stage also witnesses important methodological innovations that translate the prototypes into more practical and versatile latent techniques. These advances span both representation design and optimization strategies, progressively expanding the toolkit available.
On the representation front, a series of works explored latent representation design: Assorted~\cite{su2025token} proposes mixing latent discrete tokens with text tokens to achieve shorter traces with improved accuracy; CODI~\cite{shen2025codi} formalizes a self-distillation procedure where the same model serves as both teacher and student to generate reasoning chains in explicit and latent spaces respectively; and BoLT~\cite{ruan2025reasoning} shifts attention to pretraining, treating web text as compressed outcomes of thought processes and using bootstrapping for data-efficient pretraining.

On the optimization front, represented by HRPO~\cite{yue2025hybrid}, System-1.5~\cite{wang2025system}, and CoLaR~\cite{tan2025think}, a family of methods respectively introduced latent reinforcement methods, adaptive computation allocation between language and latent spaces, and dynamic inference-time controllable compression. These works demonstrate that latent optimization can be optimized end-to-end and adaptively allocated across different modalities.

Another characteristic of this stage is that latent-level models begin extending beyond text-only LLMs. While the primary focus remains on textual reasoning, several pioneering efforts demonstrate the broader applicability of the latent space paradigm across modalities.
Mirage~\cite{yang2025machine} enables VLMs to think visually by recasting hidden states as latent visual tokens interleaved with text, implementing internal visual manipulation analogous to human mental imagery.
UniVLA~\cite{bu2025univla} introduces task-centric latent actions for cross-embodiment robot policies, learning latent action representations from internet-scale video.
These early multimodal explorations demonstrate that the latent space paradigm is not confined to textual reasoning but may constitute a general framework applicable across modalities and embodiment types.

Despite these advances, the formation stage remains constrained by a second bottleneck: most methods are still developed and evaluated within relatively narrow settings, with text-centric assumptions, limited downstream diversity, and weak integration across memory, planning, communication, and action. As a result, the field has learned how to make latent reasoning more principled, but not yet how to make it broadly useful across heterogeneous scenarios. This gap naturally drives the next stage, where the main question shifts from ``can latent reasoning be formalized?'' to ``\textit{how far can the latent-space paradigm generalize across domains, modalities, and system settings?}''

\subsection{Expansion}
\label{sec:expansion}
The expansion stage transforms latent space research from a text-centric landscape into a multi-modal, multi-domain ecosystem. This stage witnesses rapid diversification along concurrent dimensions. The period is marked by the \textbf{Technical Maturation} of domain-specific innovations and \textbf{Paradigm/Scenario Expansion} of cross-cutting themes such as latent memory, test-time scaling, and RL-based optimization.

\xhdr{Technical Maturation}
In the LLM domain, latent methods evolve from proof-of-concept to more sophisticated systems that address challenges in memory, scalability, optimization, and domain-specific applications. The research landscape broadens considerably, with multiple sub-directions developing in parallel.
Represented by MemGen~\cite{zhang2025memgen}, a line of works pioneers latent memory for agents, interweaving reasoning and memory so that planning, procedural, and working memory types emerge without explicit supervision.
On the test-time scaling front, LTPO~\cite{ye2025thinking} treats latent thought vectors as optimizable parameters with online policy gradient; Ouro~\cite{zhu2025scaling} moves optimization into pretraining via looped language models; and \citet{you2025parallel} enables parallel test-time scaling through stochastic sampling strategies with a latent reward model for trajectory selection.
For RL-driven optimization, SofT-GRPO~\cite{zheng2025softgrpo} solves the differentiability challenge of applying RL to continuous latent reasoning through Gumbel-reparameterized policy optimization.
In the interleaving and hybrid direction, SpiralThinker~\cite{piao2025spiralthinker} and CLaRa~\cite{he2025clara} propose text-latent iterative interleaving and unified retrieval-augmented generation in shared continuous spaces, respectively.
Additionally, domain-specific applications emerge, including search-and-recommendation unification~\cite{shi2025bridging}, code language model interpretability~\cite{sharma2025analyzing}, and System 1/2 dual-architecture communication~\cite{codaforno2025exploring}.

\xhdr{Paradigm/Scenario Expansion}
The expansion stage sees an explosion of visual latent-level methods, establishing the paradigm of thinking in visual space as a complement to textual reasoning. These methods enable VLMs to perform internal visual manipulation and reasoning directly in latent representations, bypassing the information loss incurred by converting visual content into discrete text tokens.
Represented by LVR~\cite{li2025latent} and Monet~\cite{wang2025monet}, a family of works introduces autoregressive reasoning in the visual embedding space, generating latent states interleaved with text for fine-grained visual understanding.
3DThinker~\cite{chen2025think} further extends this to 3D mental simulation from limited 2D views by aligning latent representations with 3D foundation models.
Another line address latent visual tasks: VisMem~\cite{yu2025vismem} proposes cognitively inspired short-term and long-term latent vision memory modules, while CoMEM~\cite{wu2025autoscaling} scales continuous memory for visual agents.
Works such as Latent Sketchpad~\cite{zhang2025latentsketchpad} and LaCoT~\cite{sun2025latent} further explore visual scratchpads for planning and variational inference for visual reasoning, respectively.

The expansion stage gives birth to latent communication as a new paradigm for multi-agent systems. Unlike traditional text-based inter-agent communication, latent communication enables direct exchange of continuous representations, offering higher bandwidth and lower latency.
C2C~\cite{fu2025cache}  introduces direct semantic communication between LLMs via KV-cache projection and fusion.
\cite{zheng2025thought} develops a theoretical framework for mind-to-mind latent thought communication, while LatentMAS~\cite{zou2025latent} demonstrates latent collaboration through shared latent working memory, significantly reducing output tokens while improving accuracy.

The embodied domain also expands significantly during this stage, with latent representations becoming a central tool for learning and transferring robotic manipulation and navigation skills. Self-supervised pretraining from unlabeled video emerged as a key methodology.
Represented by LAPA~\cite{ye2025latent} and LAWM~\cite{tharwat2025latent}, a line of self-supervised methods pioneers latent action pretraining from unlabeled video data through world modeling.
OccVLA~\cite{liu2025occvla} integrates implicit 3D occupancy supervision for interpretable trajectory planning, while SRPO~\cite{fei2025srpo} applies RL with latent world representations for VLA training.
ATE~\cite{zhang2025alignthensteer} enables data-efficient VLA adaptation through unified latent guidance, achieving substantial real-world cross-embodiment gains.

However, rapid expansion also introduces a new bottleneck: the field becomes increasingly fragmented. Once latent-space methods spread across language, vision, multi-agent systems, and embodiment, the main challenge is no longer lack of applications, but lack of unification. Different works adopt different architectural assumptions, optimization objectives, evaluation criteria, and latent interfaces, making it difficult to compare methods or identify stable design principles. This fragmentation sets the stage for the outbreak period, in which the research frontier shifts toward architectural specialization, optimization sophistication, and more mature attempts to consolidate latent space as a first-class computational paradigm.

\subsection{Outbreak}
\label{sec:outbreak}
The outbreak stage represents an explosive acceleration of the field, characterized by the \textbf{All-round Outbreak} of all research threads. The maturity of this stage is reflected in many hallmarks:
architectural and representational specialization, with purpose-built models designed specifically for latent representation;
computation and optimization sophistication, with methods addressing fine-grained challenges such as exploration collapse and latent reward encoding;
Moreover, multi-scenario surge, with unified frameworks spanning language, vision, action, and multi-agent systems.

\xhdr{All-round Outbreak}
A defining feature of this stage is the increasing specialization of model architectures. Rather than merely adapting standard transformer backbones via shallow recurrence or iterative decoding, recent work has developed architectures explicitly designed to support latent computation as a first-class mechanism. These models generally aim to improve the controllability, efficiency, and expressiveness of reasoning in latent space.

Representative examples include Dreamer~\cite{knupp2026depthrecurrent} and LoopFormer~\cite{jeddi2026loopformer}, which introduce depth-recurrent designs that combine sequence attention with depth-wise computation and elastic looping, thereby enabling budget-aware reasoning and more flexible compute allocation. MLRA~\cite{liu2026multi} further revises the attention mechanism through low-rank projections, while DLCM~\cite{qu2025dynamic} shifts the granularity of latent computation from token-level operations toward concept-level reasoning with adaptive conceptual boundaries. Collectively, these studies suggest that architectural development is moving beyond incremental modifications to sequence models toward dedicated latent-space systems.

Alongside architectural specialization, optimization strategies for latent space also become substantially more sophisticated. 
For example, ReLaX~\cite{zhang2025relax} and Active Latent Planning~\cite{zheng2026beyond} advance latent-space exploration by moving beyond imitation-based learning toward more explicit reinforcement-learning-based planning. At the same time, \citet{enes2025reinforcement} demonstrate, through a systematic analysis, that reinforcement learning remains sensitive to design choices and continues to face persistent optimization challenges. LED~\cite{xiaomi2026led} addresses post-training exploration collapse by leveraging entropy variation across recurrent depth. In contrast, Latent Thinking Optimization~\cite{du2025latent} shows that latent thoughts can themselves encode reward-relevant information, thereby opening the possibility of directly optimizing latent trajectories without relying exclusively on external reward models. Overall, these developments indicate that optimization is becoming an independent axis of progress rather than a secondary concern.

In visual tasks, recent studies demonstrate that latent space supports increasingly complex forms of multi-step and interleaved inference. ILVR~\cite{dong2025interleaved} and CrystaL~\cite{yang2026crystal}, for instance, explore visual-text interleaved reasoning and report the emergence of visual latent representations during the reasoning process. LIVR~\cite{li2025latent1} and Mull-Tokens~\cite{ray2025mulltokens} further extend this line of work by pushing visual reasoning more fully into latent space and by proposing modality-agnostic latent thinking mechanisms. Related efforts such as VL-JEPA~\cite{chen2025vljepa} and DMLR~\cite{liu2025reason} further expand the design space of visual reasoning and highlight the growing diversity of methodological formulations in this area.

A similar expansion is also evident in multi-agent systems, where latent communication evolves from preliminary demonstrations into more structured frameworks for coordination and representation sharing. \citet{dery2026latent} propose latent-space communication via K-V cache alignment with lightweight adapters, enabling translation between heterogeneous internal states. L2-VMAS~\cite{yu2026dual} and Wormhole~\cite{liu2026vision} extend this perspective to visual and heterogeneous multi-agent settings, while LatentMem~\cite{fu2026latentmem} introduces shared latent memory mechanisms for multi-agent experience accumulation. These works suggest that latent communication is becoming an important mechanism for scalable inter-agent coordination.

The embodied VLA setting provides another major area of expansion. Here, latent representations increasingly serve as a unifying interface for perception, generation, and action, and latent action modeling is emerging as a central design paradigm. Motus~\cite{bi2025motus} and VLA-JEPA~\cite{ginwind2026vlajepa} exemplify this trend by developing unified latent action world models that integrate action generation and environment understanding within a shared latent space. Villa-X~\cite{chen2025villax} and JALA~\cite{jala2026joint} further improve the expressiveness and scalability of latent action modeling, while CoWVLA~\cite{fx2026cowvla} introduces world-model reasoning in latent motion space. Additional work, including WholeBodyVLA~\cite{jiang2025wholebodyvla}, SwiftVLA~\cite{ni2025swiftvla}, and LoLA~\cite{wang2025lola}, reinforces the view that latent action representations are becoming increasingly central to VLA pretraining and deployment.

At the same time, the outbreak stage surfaces the next-order bottleneck for the field as a whole: once latent-space methods become specialized, powerful, and widespread, the hardest problem is no longer demonstration but consolidation. The open questions now concern standardization of latent interfaces, principled evaluation across modalities, alignment between latent efficiency and interpretability, and the integration of latent computation into broader agentic systems. In this sense, the outbreak stage does not mark the end of the story; rather, it reveals that the next phase of progress will likely depend on turning a rapidly growing collection of techniques into a coherent science and harness engineering for latent computation.

\begin{table}[t]
    \centering
    \caption{General notations used throughout the paper.}
    \label{tab:notation}
    \setlength{\tabcolsep}{2mm}
    \begin{tabular}{l|ll}
        \toprule
        \textbf{Symbol} & \textbf{Space} & \textbf{Description} \\
        \midrule
        $\mathcal{V}$ & token space & discrete vocabulary/token IDs \\
        $\mathcal{H}$ & latent space & continuous hidden space $\mathbb{R}^d$ \\
        $\mathcal{M}$ & model space & set of models/modules/components \\
        $\mathcal{W}$ & parameter space & parameters within the model \\
        $\mathcal{D}$ & data space & the whole training and test data \\
        \midrule
        $\mathbf{x}$ & $\mathcal{V}$ & input \\
        $\mathbf{y}$ & $\mathcal{V}$ & output \\
        $\mathbf{r}$ & $\mathcal{V}$ & generation trajectory  \\
        $\mathbf{E}$ & $\mathcal{V}$ & matrix embedding \\
        $\mathbf{W}$ & $\mathcal{V}$ & projection matrix \\
        $\mathbf{q,k,v}$ & $\mathcal{V}$ & query, key, value \\
        $\mathbf{h}$ & $\mathcal{H}$ & hidden state of one token \\
        $\mathbf{H}$ & $\mathcal{H}$ & hidden state of token sequence \\
        $\mathbf{z}$ & $\mathcal{H}$ & latent representation \\
        $\Phi$ & $\mathcal{M}$ & whole framework \\
        $\Phi^{back}$ & $\mathcal{M}$ & backbone \\
        $\Phi^{\mathrm{comp}}$ & $\mathcal{M}$ & functional component \\
        $\Phi^{\mathrm{aux}}$ & $\mathcal{M}$ & auxiliary model \\
        $\theta$ & $\mathcal{W}$ & model parameters \\
        $R$ & -- & reward function  \\
        $\mathcal{L}$ & -- & loss function \\
        \bottomrule
    \end{tabular}
\end{table}

\section{\textcolor{secblue}{Mechanism:} How Does Latent Space Work?}
\label{sec:mechanism}

\begin{figure*}[t]
  \centering
    \includegraphics[width=1\linewidth]{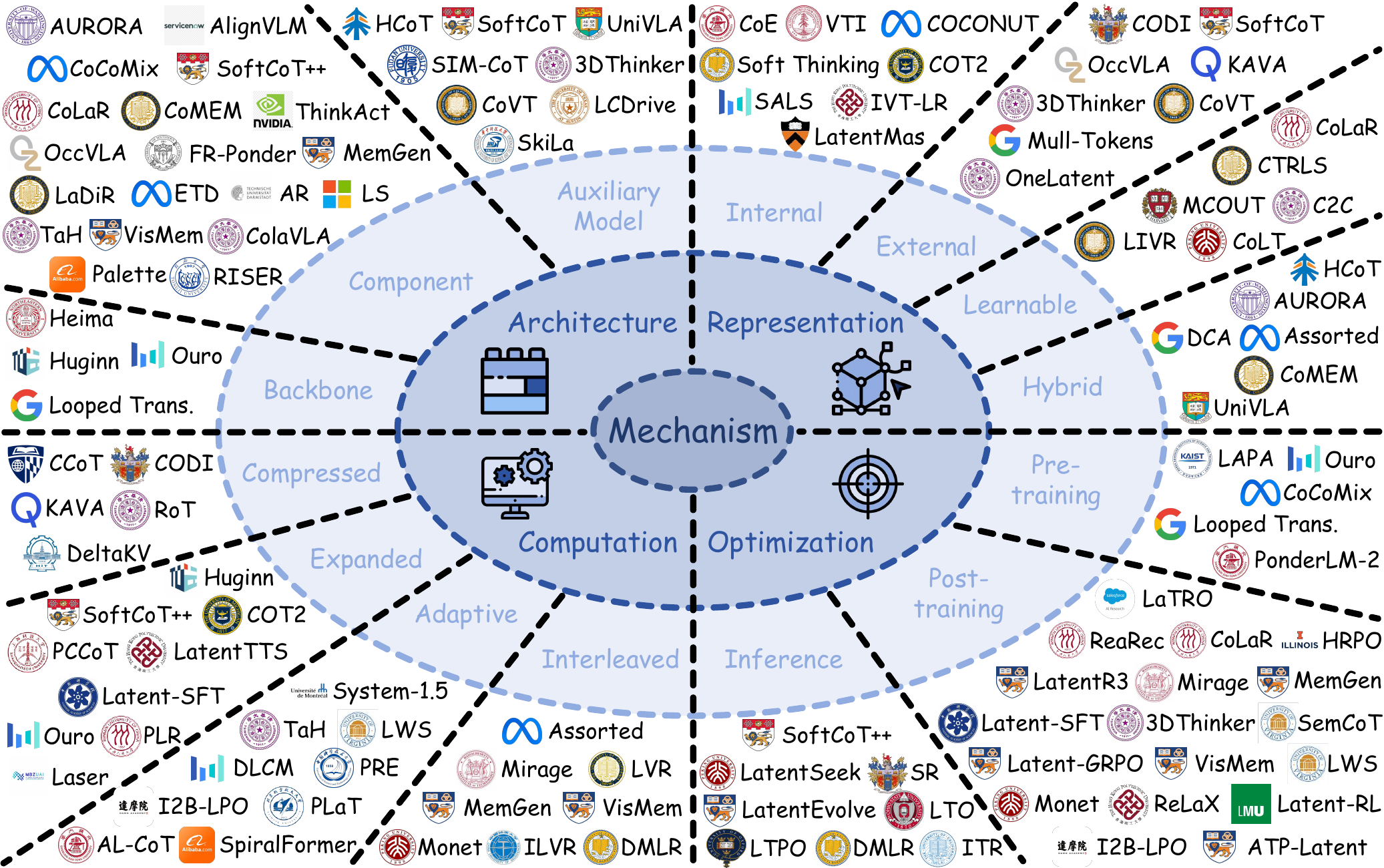}
    \caption{Representative works operate in accordance with latent space mechanisms. We classify all methods into four lines based on diverse ways of utilizing the latent space, including: \textbf{Architecture} (Section~\ref{sec:architecture}), \textbf{Representation} (Section~\ref{sec:representation}), \textbf{Computation} (Section~\ref{sec:computation}), and \textbf{Optimization} (Section~\ref{sec:optimization}).}
    \label{fig:mechanism}
\end{figure*}

Mechanism concerns how latent space is instantiated, structured, and operationalized within a model, beyond the high-level question of why it is useful. Existing methods differ not only in where latent variables are introduced, but also in how they are represented, how computation is carried out through them, and at which stage they are shaped or exploited. To organize this design space, we categorize latent-space mechanisms along four complementary axes: 
\textbf{Architecture} (Section~\ref{sec:architecture}) characterizes the structural role of latent space in the model, including whether it is embedded in the backbone, realized as a dedicated component, or supported by an auxiliary model;
\textbf{Representation} (Section~\ref{sec:representation}) describes the form of latent variables, distinguishing internal, external, learnable, and hybrid representations. 
\textbf{Computation} (Section~\ref{sec:computation}) captures how the latent space participates in information processing, including compressed, expanded, adaptive, or interleaved computation. 
\textbf{Optimization} (Section~\ref{sec:optimization}) focuses on when and how latent space is induced, aligned, or refined, spanning pre-training, post-training, and inference-time mechanisms. 
These four mechanistic types provide a unified lens for comparing diverse approaches and clarify the major design choices that govern how latent spaces are constructed and used in modern language-based systems.

\xhdr{General Notation and Formalization} Based on the notations in Table~\ref{tab:notation}, we first formalize the standard autoregressive generation paradigm. Given an input sequence $\mathbf{x}\in\mathcal{V}$, a model $\Phi_\theta$ defines a conditional distribution over the output sequence $\mathbf{y}\in\mathcal{V}$:
\begin{equation}
    \mathbf{y} \sim \Phi_{\theta}(\cdot \mid \mathbf{x}).
\end{equation}
where generation is performed purely in token space: the model predicts each output token conditioned on the input and the previously generated context. Although the computation is internally carried out through continuous hidden states $\mathbf{h}\in\mathcal{H}$, the generation interface itself remains token-to-token.

Latent-space methods extend this formulation by introducing an additional latent representation $\mathbf{z}\in\mathcal{H}$, such that generation is conditioned not only on the observable input $\mathbf{x}$ but also on a continuous latent variable:
\begin{equation}
    \mathbf{y} \sim \Phi_{\theta}(\cdot \mid \mathbf{x}, \mathbf{z}),
    \label{eq:latent_autogressive}
\end{equation}
where $\mathbf{z}$ denotes a latent representation in the latent space $\mathcal{H}$. Compared with standard autoregressive generation, the latent variable $\mathbf{z}$ provides an additional channel for encoding information that may be difficult to express directly in token space, such as global semantics, multimodal features, intermediate reasoning states, structural constraints, or other task-relevant factors.

Under this perspective, the central question is not merely whether a method uses latent variables, but how latent space is instantiated and integrated into the generation process. This motivates the mechanism-oriented taxonomy in this survey.

\subsection{Architecture}
\label{sec:architecture}

Recent advances in latent-level methods have catalyzed a profound rethinking of the established architectural design paradigms for models operating in explicit representational spaces~\cite{hu2024understanding}. Departing from the exclusive reliance on conventional autoregressive frameworks, an increasing number of studies have pioneered innovative mechanisms that enable core computations within latent spaces, where high-level cognitive processes~\cite{li2026dynamics}, \textit{e.g.}, reasoning, perception, and planning, can be conducted with substantially enhanced efficiency and expressiveness. This transformative shift extends far beyond a mere change of representational domain; it encapsulates a fundamental evolution in the architectural philosophy of modern neural models~\cite{zhu2025reasoning}.

This emerging paradigm reveals that latent space is evolving from an isolated technical heuristic into a general, foundational architectural principle. To answer the core question: \textit{what architectures are utilized to learn the latent space?},
we distinguish these methods by the location where latent space $\mathcal{H}$ is integrated into $\Phi$, with a focus on the \textbf{structure} of the latent system within model space $\Phi=\left\{\Phi^{back},\Phi^{comp},\Phi^{aux}\right\}$.
As reported in Table~\ref{tab:architecture_backbone} and Table~\ref{tab:architecture_module}, we classify all existing architecture-driven methods into three categories:
\vspace{5pt}
\begin{tcolorbox}[
  colback=secblue!5,
  colframe=secblue!50,
  colbacktitle=secblue!50,
  coltitle=black,
  title={\textbf{\textcolor{secblue}{Mechanism:} Architecture}},
  boxrule=5pt,
  arc=5pt,
  drop shadow,
  parbox=false,
  before skip=5pt,
  after skip=10pt,
  left=5pt,   
  right=20pt,
]
\begin{itemize}
    \item \textbf{Backbone} (Section~\ref{sec:backbone}): endows the main model with native latent capacity through recurrent, looping, recursive structures, thereby making an architectural primitive.
    \item \textbf{Component} (Section~\ref{sec:component}): employs generation, projection heads, alignment, control, storage, or other components, which allow latent functions while preserving the main model skeleton.
    \item \textbf{Auxiliary Model} (Section~\ref{sec:auxiliary_model}): utilizes an extra model to provide supervision signals or intermediate features to guide or supplement the host model.
\end{itemize}
\end{tcolorbox}

\begin{table*}[t]
\centering
\caption{Overview of the \textbf{Backbone} (Section~\ref{sec:backbone}) based architecture. We compare the hidden dimension, layer, size, and architectural feature of these backbones.}
\setlength{\tabcolsep}{0.9mm}
\resizebox{1\textwidth}{!}{
\begin{tabular}{l|l|ll|llll}
\toprule
\textbf{Date} & \textbf{Backbone} & \textbf{Paper} & \textbf{Code} & \textbf{Dimension} & \textbf{Layer} & \textbf{Size}  & \textbf{Feature}    \\ \midrule
01/25 & Heima~\cite{shen2025efficient} & \paperlink{https://arxiv.org/abs/2501.19201} & \githublink{https://github.com/shawnricecake/Heima} & 4096 & 72 & 19B & encoder-decoder/progressive/adaptive decoding \\
02/25 & Huginn~\cite{jonas2025scaling} & \paperlink{https://arxiv.org/abs/2502.05171} & \githublink{https://github.com/seal-rg/recurrent-pretraining} & 5280 & 8 & 3.5B & decoder-only/recurrent depth/shared recurrent block/test-time \\
02/25 & Looped Trans.~\cite{saunshi2025reasoning} & \paperlink{https://arxiv.org/abs/2502.17416} & \qquad - & 5120 & 24 & 1.5B & decoder-only/looped model/looping-based regularization \\
03/25 & MoLAE~\cite{liu2025molae} & \paperlink{https://arxiv.org/abs/2503.23100} & \qquad - & 512 & 12 & 0.1B & mixture of latent experts/shared projection/lower dimension \\
04/25 & PHD-Trans.~\cite{wu2025efficient} & \paperlink{https://arxiv.org/abs/2504.14992} & \qquad - & 2048 & 16 & 1.2B & decoder-only/cache management/sliding window attention \\
09/25 & PonderLM2~\cite{zeng2025pretraining} & \paperlink{https://arxiv.org/abs/2509.23184} & \githublink{https://github.com/LUMIA-Group/PonderLM-2} & 2048 & 24 & 0.5B/1.4B & decoder-only/iterative refinement/Jacobi-style parallel updates \\
10/25 & Ouro~\cite{zhu2025scaling} & \paperlink{https://arxiv.org/abs/2510.25741} & \qquad - & 2048 & 24/48 & 1.4B/2.6B & decoder-only/recursive inference/parameter-shared loop \\
12/25 & DLCM~\cite{qu2025dynamic} & \paperlink{https://arxiv.org/abs/2512.24617} & \qquad - & 1536 & 32 & 2.3B & encoder-decoder/large concept model/hierarchical/heterogeneous \\
01/26 & Dreamer~\cite{knupp2026depthrecurrent} & \paperlink{https://arxiv.org/abs/2601.21582} & \qquad - & 1024 & 16/32 & 1B/2B & depth-recurrent/sequence-depth-sparse attention mixture \\
02/26 & LoopFormer~\cite{jeddi2026loopformer} & \paperlink{https://arxiv.org/abs/2602.11451} & \githublink{https://github.com/armenjeddi/loopformer} & 2048 & - & - & decoder-only/elastic-depth looped transformer \\
03/26 & MLRA~\cite{liu2026multi} & \paperlink{https://arxiv.org/abs/2603.02188} & \githublink{https://github.com/SongtaoLiu0823/MLRA} & 3072 & 24 & 2.9B & multi-head low-rank attention/four-way tensor parallelism decoding \\
\bottomrule
\end{tabular}}
\label{tab:architecture_backbone}
\end{table*}

\subsubsection{Backbone}
\label{sec:backbone} 
In this category, latent computation is intrinsically embedded in the primary generative architecture rather than introduced via an external auxiliary module. Formally, the backbone itself carries out an iterative or structured transition over latent states:
\begin{equation}
\mathbf{h}_{t+1} = \Phi^{back}(\mathbf{h}_{1:t}, \mathbf{x}, \mathbf{y}_{1:t}),
\end{equation}
where each subsequent output token is produced based on the updated hidden state, under this formulation, the latent operation constitutes a native operational mechanism of $\Phi^{back}(\cdot)$ itself, meaning that the reasoning or intermediate transformation process is realized internally within the backbone, without requiring any additional component.

Existing backbone-based methods can be broadly understood from three complementary perspectives: \textbf{Parameter-shared}, \textbf{Iterative Refinement}, and \textbf{Augmented}. Rather than simply following the explicit-level counterparts, these works revisit the backbone design itself to offer a promising path.

\xhdr{Parameter-shared Backbone} From the perspective of parameter sharing, a representative line of work replaces a deep stack of distinct layers with a smaller set of reusable modules applied repeatedly. However, in this paradigm, the recurrent depth or times is typically fixed. Huginn~\cite{jonas2025scaling}, for example, adopts a decoder-only architecture, but compensates for the shallow explicit depth by introducing a shared recurrent block that is reused across multiple depth steps. This design substantially reduces the number of unique parameters while enabling test-time scaling, where additional recurrent steps can be applied during inference to trade compute for performance. Looped Trans.~\cite{saunshi2025reasoning} further develops this idea by enforcing an explicit layer-looping mechanism. A key technical feature is its looping-based regularization, which stabilizes the hidden-state dynamics under repeated application of the same transformation and encourages convergence of iterative representations.
In a related efficiency-oriented setting, PHD-Trans.~\cite{wu2025efficient} explores how such recurrent computation can remain practical under long-context decoding. It integrates cache management with sliding-window attention, reducing memory overhead while preserving the benefits of repeated updates.

\xhdr{Iterative Backbone} This paradigm highlights the iterative process and dynamic updating, with variable or learnable depth allocation.
For instance, Ouro~\cite{zhu2025scaling} introduces a recursive inference framework whose iterations can serve as an alternative scaling axis, analogous to increasing architectural depth, showing that repeated application of parameter-shared transformations can yield consistent gains. LoopFormer~\cite{jeddi2026loopformer} introduces an elastic-depth looped transformer, where the number of loop iterations is not fixed but can vary across inputs or computational budgets. This makes recurrent-depth execution more flexible and better aligned with the complexity of individual examples.
PonderLM2~\cite{zeng2025pretraining} is a representative method in this category. Built as a decoder-only model, it employs iterative refinement via Jacobi-style parallel updates. Instead of strictly following the standard autoregressive regime, where each forward pass advances the prediction by one token, it also performs multi-step hidden-state evolution in parallel, enabling richer internal computation while retaining decoding efficiency.

\xhdr{Augmented Backbone} Beyond these two designs, several works explore augmented architectures at different granularities.
For example, Heima~\cite{wu2025efficient} and DLCM~\cite{qu2025dynamic}, unlike most encoder-only designs, employ hierarchical encoder-decoder architectures to organize latent computation. The latter one also shifts computation from tokens to a compressed concept space, with more efficient semantic operations.
Dreamer~\cite{knupp2026depthrecurrent} and MLRA~\cite{liu2026multi} design sequence-depth sparse attention mixtures and low-rank attention, respectively, making multi-step latent transitions computationally tractable and efficient. 
In addition, MoLAE~\cite{liu2025molae} designs an architecture of reformulating the mixture of latent experts in lower dimension.

\xhdr{Summary} Overall, backbone-oriented methods represent the most intrinsic form of architecture-level latent modeling.
Parameter-sharing approaches improve efficiency by reusing subsets of layers or models; iterative-refinement methods further enhance flexibility by enabling dynamic, adaptive iterations; and augmentation-based designs provide a broader view of architectural shifts.
This shift provides the foundation for more flexible, computation-aware, and cognitively expressive generative systems.

\begin{table*}[!t]
\centering
\caption{Overview of the \textbf{Component} (Section~\ref{sec:component}) and \textbf{Auxiliary Model} (Section~\ref{sec:auxiliary_model}) based architecture, respectively. We compare the modality, backbone, the type of the component module or external model, function, and scenario. Here, (VQ)-AE, SAE, MLP, Q-Former, LoRA, and JEPA denote (vector quantized) variational autoencoder, multilayer perceptron, querying transformer, low-rank adaptation, and joint embedding predictive architecture, respectively.}
\setlength{\tabcolsep}{0.9mm}
\resizebox{1\textwidth}{!}{
\begin{tabular}{l|l|ll|lllll}
\toprule
\textbf{Date} & \textbf{Method} & \textbf{Paper} & \textbf{Code} & \textbf{Modality} & \textbf{Backbone}  & \textbf{Module/Model} & \textbf{Function} & \textbf{Scenario}   \\ \midrule
\multicolumn{7}{l}{\textbf{Component}} \\\midrule
12/24 & AURORA~\cite{perception2025bigverdi} & \paperlink{https://arxiv.org/abs/2412.03548} & \githublink{https://github.com/mahtabbigverdi/Aurora-perception} & vision & LLaVA1.5-13B & VQ-VAE & vision generation & visual perception \\
01/25 & LF-Steering~\cite{yang2025lfsteering} & \paperlink{https://arxiv.org/abs/2501.11036} & \qquad - & text & LLaMA2-7B &  SAE & semantic projection & semantic consistency \\
02/25 & AlignVLM~\cite{ahmed2025alignvlm} & \paperlink{https://arxiv.org/abs/2502.01341} & \qquad - & vision & LLaMA3.2-1B/3B & mixed layers & vision alignment & document understanding \\
02/25 & CoCoMix~\cite{jihoon2025llm} & \paperlink{https://arxiv.org/abs/2502.08524} & \githublink{https://github.com/facebookresearch/RAM/tree/main/projects/cocomix} & text & GPT2-0.1B/0.4B/1.4B & SAE & concept extraction & interpretability/efficiency \\
02/25 & IMM~\cite{jos2025beyond} & \paperlink{https://arxiv.org/abs/2502.21030} & \qquad - & text & GPT2-0.1B & vector library & memory bank & long-horizon/safety \\
05/25 & SoftCoT++~\cite{xu2025softcot1} & \paperlink{https://arxiv.org/abs/2505.11484} & \githublink{https://github.com/xuyige/SoftCoT} & text & LLaMA3.1-8B & linear layer & semantic projection & exploration/zero-shot/scaling \\
05/25 &  CoLaR~\cite{tan2025think} & \paperlink{https://arxiv.org/abs/2505.16552} & \githublink{https://github.com/xiaomi-research/colar} & text & LLaMA3.2-1B & MLP & embedding generation & dynamic reasoning/efficiency \\
05/25 & CoMEM~\cite{wu2025towards} & \paperlink{https://arxiv.org/abs/2505.17670} & \githublink{https://github.com/WenyiWU0111/CoMEM/tree/main} & vision & Qwen2/2.5-VL-7B & Q-Former & memory generation & visual reasoning/long-horizon/efficiency \\
07/25 & ThinkAct~\cite{huang2025thinkact} & \paperlink{https://arxiv.org/abs/2507.16815} & \githublink{https://jasper0314-huang.github.io/thinkact-vla} & action & Qwen2.5-VL-7B & Q-Former & action projection & embodied manipulation/long-horizon \\
09/25 & OccVLA~\cite{liu2025occvla} & \paperlink{https://arxiv.org/abs/2509.05578} & \qquad - & action & PaliGemma2-3B & transformer & feature injection & spatial understanding \\
09/25 & FR-Ponder~\cite{he2025learning} & \paperlink{https://arxiv.org/abs/2509.24238} & \qquad - & text & LLaMA3-8B/other 4 & MLP &  trigger & dynamic reasoning/efficiency/generalization  \\
09/25 & MemGen~\cite{zhang2025memgen} & \paperlink{https://arxiv.org/abs/2509.24704} & \githublink{https://github.com/bingreeky/MemGen}  & text & SmolLM3-3B/Qwen3-8B & LoRA  & trigger/generation & experimental memory/generalization \\
10/25 & HYP1/2~\cite{codaforno2025exploring} & \paperlink{https://arxiv.org/abs/2510.00494} & \qquad - & text & GPT2-0.1B/Qwen3-0.6B & transformer & messages exchange & latent reasoning/communication \\
10/25 & LaDiR~\cite{kang2025ladir} & \paperlink{https://arxiv.org/abs/2510.04573} & \githublink{https://github.com/mk322/LaDiR} &  text & LLaMA3.1-8B & VAE decoder & semantic encoding & exploration/diffusion-augmented \\
10/25 & ETD~\cite{koishekenov2025encode} & \paperlink{https://arxiv.org/abs/2510.07358} & \qquad - & text &  OLMo-2-1B & encoder-decoder & token generation & multi-step reasoning/zero-shot \\
10/25 & CoMEM-Agent~\cite{wu2025autoscaling} & \paperlink{https://arxiv.org/abs/2510.09038} & \githublink{https://github.com/WenyiWU0111/CoMEM-Agent} & vision & Qwen2.5-VL-7B/other 1 & Q-Former & memory generation & GUI agent/long-horizon \\
10/25 & Kelp~\cite{li2025kelp} & \paperlink{https://arxiv.org/abs/2510.09694} & \githublink{https://github.com/Alibaba-AAIG/Kelp} & text & Qwen3-8B & MLP & trigger & risk detection/efficiency \\
10/25 & AR~\cite{lukas2025activationreasoning} & \paperlink{https://arxiv.org/abs/2510.18184} & \qquad -  & text & LLaMA3.1-8B/Gemma2-9B & SAE & feature generation & safety/multi-hop reasoning \\
10/25 & LS~\cite{zhang2025latentsketchpad} & \paperlink{https://arxiv.org/abs/2510.24514} & \githublink{https://github.com/hwanyu112/Latent-Sketchpad} & vision & Gemma3-12B/other 2 & VAE decoder & vision generation & visual imagination/visual understanding \\
11/25 & TaH~\cite{fu2025tah} & \paperlink{https://arxiv.org/abs/2511.08577} & \githublink{https://github.com/thu-nics/TaH} & text & Qwen3-0.6B/1.7B & MLP & decider & dynamic reasoning/complex reasoning \\
11/25 & SpiralThinker~\cite{piao2025spiralthinker} & \paperlink{https://arxiv.org/abs/2511.08983} & \qquad - & text & LLaMA3.2-7B & MLP & semantic projection & latent reasoning/complex reasoning \\
11/25 & VisMem~\cite{yu2025vismem} & \paperlink{https://arxiv.org/abs/2511.11007} & \githublink{https://github.com/YU-deep/VisMem} & vision & Qwen2.5-VL-7B/other 8 & LoRA  & memory generation & visual understanding/visual reasoning \\
11/25 & VITA~\cite{ma2025unifying} & \paperlink{https://arxiv.org/abs/2511.19859} & \githublink{https://github.com/vita-cvpr26/vita} & action & PaliGemma-3B & encoder-decoder & feature encoding & embodied manipulation \\
12/25 & LiteReason~\cite{gurung2025lightweight} & \paperlink{https://arxiv.org/abs/2512.02240} & \qquad - & text & Qwen2.5-7B & MLP & semantic projection & latent reasoning/efficiency \\
12/25 & Palette~\cite{long2025reasoning} & \paperlink{https://arxiv.org/abs/2512.17206} & \qquad - & vision & Qwen3-1.7B/other 2 & VAE decoder & token generation & exploration/latent reasoning \\
12/25 & JEPA-Reasoner~\cite{liu2025jepareasoner} & \paperlink{https://arxiv.org/abs/2512.19171} & \qquad - & text & transformer-based & JEPA & chains generation & complex reasoning/robustness \\
12/25 & LoLA~\cite{wang2025lola} & \paperlink{https://arxiv.org/abs/2512.20166} & \qquad - & action & Qwen2.5-VL-7B & transformer & feature alignment & embodied manipulation/long-horizon \\
12/25 & ColaVLA~\cite{peng2025colavla} & \paperlink{https://arxiv.org/abs/2512.22939} & \githublink{https://github.com/pqh22/ColaVLA} & action & LLaVA-v1.5-7B & forward layer & vision projection & planning/autonomous driving \\
12/25 & iCLP~\cite{chen2025iclp} & \paperlink{https://arxiv.org/abs/2512.24014} & \githublink{https://github.com/AgenticFinLab/latent-planning} & text & Qwen2.5-0.5B/3B/7B & VAE & token generation & latent planning/efficiency  \\
01/26 & LaST0~\cite{liu2026last0} & \paperlink{https://arxiv.org/abs/2601.05248} & \qquad - & action & DeepSeek-LLM-1.5B & MLP & action projection & embodied manipulation/efficiency  \\
01/26 & RB-CoT~\cite{he2026reasoning} & \paperlink{https://arxiv.org/abs/2601.08058} & \qquad - & text & LLaMA3.1-8B/other 5 & SAE & feature generation & multi-step reasoning/prompting \\
01/26 & RISER~\cite{ye2026riser} & \paperlink{https://arxiv.org/abs/2601.09269} & \githublink{https://github.com/gooogleshanghai/RISER-Orchestrating-Latent-Reasoning-Skills-for-Adaptive-Activation-Steering} & text & Qwen2.5-7B/other 3 & vector library & reusable chains & zero-shot/generalization \\
01/26 & Fast-ThinkAct~\cite{huang2026fastthinkact} & \paperlink{https://arxiv.org/abs/2601.09708} & \qquad - & action & Qwen2.5-VL-3B & MLP & action projection & embodied manipulation/efficiency \\
01/26 & GeoSteer~\cite{kazama2025geosteer} & \paperlink{https://arxiv.org/abs/2601.10229} & \qquad - & text & Qwen3-8B/other 3 & VAE & projection & complex reasoning/consistency \\
01/26 & PREGEN~\cite{serussi2026pregen} & \paperlink{https://arxiv.org/abs/2601.13797} & \qquad - & vision & Qwen2.5-VL-7B/other 3 & mixed layers & embedding generation & video retrieval/zero-shot \\
01/26 & PILOT~\cite{zheng2026pilot} & \paperlink{https://arxiv.org/abs/2601.19917} & \qquad - & text & Qwen2.5-1.5B/other 2 &  MLP & guidance synthesis & robustness/generalization \\
01/26 & PLaT~\cite{wang2026latent} & \paperlink{https://arxiv.org/abs/2601.21358} & \githublink{https://github.com/yunsaijc/PLaT} & text & GPT2-0.1B & linear layer & semantic projection & exploration/efficiency/generalization \\
01/26 & ATP-Latent~\cite{zheng2026beyond} & \paperlink{https://arxiv.org/abs/2601.21598} & \githublink{https://github.com/zz1358m/ATP-Latent-master} & text & LaMA3.2-1B & VAE decoder & token generation & latent reasoning/generalization \\
01/26 & ReGuLaR~\cite{wang2026regular} & \paperlink{https://arxiv.org/abs/2601.23184} & \githublink{https://github.com/FanmengWang/ReGuLaR} & text & LLaMA-3.2-1B & VAE & chain rendering & latent reasoning/efficiency \\
02/26 & CoLT~\cite{zhu2026colt} & \paperlink{https://arxiv.org/abs/2602.04246} & \qquad - & text & LLaMA3.2-1B & transformer decoder & state unpacking & tool calling/efficiency \\
02/26 & PolarMem~\cite{chen2026polarmem} & \paperlink{https://arxiv.org/abs/2602.00415} & \githublink{https://github.com/czs-ict/PolarMem} & vision & Qwen2.5-VL-7B/other 7 & graph topology & state storage & long-horizon/hallucination mitigation \\
02/26 & G-MemLLM~\cite{xu2026gmemllm} & \paperlink{https://arxiv.org/abs/2602.00015} & \qquad - & text & GPT2-0.1B/LLaMA3.1-8B & gated logic & memory bank & long-horizon/multi-hop reasoning \\
02/26 & L2-VMAS~\cite{yu2026dual} & \paperlink{https://arxiv.org/abs/2602.00471} & \githublink{https://github.com/YU-deep/L2-VMAS} & vision & Qwen3-VL-8B/other 8 & vector library & dual vision memory & multi-agent collaboration/long-horizon \\
02/26 & LatentMem~\cite{fu2026latentmem} & \paperlink{https://arxiv.org/abs/2602.03036} & \githublink{https://github.com/KANABOON1/LatentMem} & text & Qwen3-4B/LLaMA3.1-8B & LoRA  & memory generation & multi-agent collaboration \\
02/26 & Interpeter~\cite{kamai2026talking} & \paperlink{https://arxiv.org/abs/2602.09670} & \qquad - & text & LLaMA3.2-1B/other 2 & LoRA  & ability transfer & domain adaptation \\
02/26 & Wormhole~\cite{liu2026vision} & \paperlink{https://arxiv.org/abs/2602.15382} & \githublink{https://github.com/xz-liu/heterogeneous-latent-mas} & vision & Qwen3-VL-2B/other 6 & encoder-decoder & vision projection & multi-agent collaboration \\

\midrule
\multicolumn{7}{l}{\textbf{Auxiliary Model}} \\\midrule
09/24 & HCoT~\cite{liu2024expediting} & \paperlink{https://arxiv.org/abs/2409.08561} & \qquad - & text & LLaMA2-7B/13B & LLM & chain generation & efficiency/generalization \\
02/25 & SoftCoT~\cite{xu2025softcot} & \paperlink{https://arxiv.org/abs/2502.12134} & \githublink{https://github.com/xuyige/SoftCoT} & text & LLaMA3.1-8B & LLM & chain generation & efficiency/generalization/zero-shot \\
05/25 & UniVLA~\cite{bu2025univla} & \paperlink{https://arxiv.org/abs/2505.06111} & \githublink{https://github.com/OpenDriveLab/UniVLA} & action & Prismatic-7B & vision encoder & feature generation & embodied manipulation/scaling \\
07/25 & CTRLS~\cite{wu2025ctrls} & \paperlink{https://arxiv.org/pdf/2507.08182} & \qquad - & text & LLaMA3.2-3B/Qwen2.5-3B & external model & state generation & exploration/latent reasoning \\
09/25 & SIM-CoT~\cite{wei2025simcot} & \paperlink{https://arxiv.org/abs/2509.20317} & \githublink{https://github.com/InternLM/SIM-CoT} & text & GPT2-0.1B/other 3 & LLM & semantic alignment & latent reasoning/efficiency \\
09/25 & SSM-VLA~\cite{cai2025seeing} & \paperlink{https://arxiv.org/abs/2509.26251} & \qquad - & action & 	LLaVA1.5-7B & latent action model & feature generation & embodied manipulation/decision  \\
09/25 & LatentEvolve~\cite{zhang2025latentevolve} & \paperlink{https://arxiv.org/abs/2509.24771} & \githublink{https://github.com/jins7/LatentEvolve} & text & LLaMA3.2-3B/other 3 & LLM & token generation &  experimental memory \\
10/25 & 3DThinker~\cite{chen2025think} & \paperlink{https://arxiv.org/abs/2510.18632} & \githublink{https://github.com/zhangquanchen/3DThinker} & vision & Qwen2.5-VL-7B/other 7 & vision model & feature generation & spatial understanding/visual reasoning \\
10/25 & SemCoT~\cite{he2025semcot} & \paperlink{https://arxiv.org/abs/2510.24940} & \githublink{https://github.com/YinhanHe123/SemCoT}  & text &  LLaMA2-7B/Mistral-7B & LLM & alignment/generation & complex reasoning/efficiency \\
11/25 & LaRe~\cite{ma2025multimodal} & \paperlink{https://arxiv.org/abs/2511.02360} & \qquad - & vision &  Qwen2.5-7B/Vicuna-7B & diffusion model & vision reconstruction & visual understanding/efficiency \\
11/25 & CoVT~\cite{qin2025chainofvisualthought} & \paperlink{https://arxiv.org/abs/2511.19418} & \githublink{https://github.com/Wakals/CoVT} & vision & Qwen2.5-VL-7B & vision model & feature generation & visual understanding/visual reasoning \\
12/25 & SwiftVLA~\cite{ni2025swiftvla} & \paperlink{https://arxiv.org/pdf/2512.00903} & \githublink{https://github.com/GigaAI-research/SwiftVLA} & action & SmolVLM-0.5B & vision model & feature generation & embodied manipulation/efficiency \\
12/25 & LCDrive~\cite{tan2025latent} & \paperlink{https://arxiv.org/abs/2512.10226} & \qquad - & action & Qwen3-0.5B & world model & feature generation & autonomous driving/efficiency \\
12/25 & VL-JEPA~\cite{chen2025vljepa} & \paperlink{https://arxiv.org/abs/2512.10942} & \qquad - & vision & LLaMA3.2-1B & JEPA & vision projection & classification/retrieval \\
12/25 & WholeBodyVLA~\cite{jiang2025wholebodyvla} & \paperlink{https://arxiv.org/abs/2512.11047} & \githublink{https://github.com/OpenDriveLab/WholebodyVLA} & action & Prismatic-7B & latent action model & feature encoding & embodied manipulation \\
12/25 & SkiLa~\cite{tong2025sketchinlatents} & \paperlink{https://arxiv.org/abs/2512.16584} & \githublink{https://github.com/TungChintao/SkiLa} & vision & Qwen2.5-VL-7B & vision encoder & vision generation  & visual imagination/visual understanding \\
01/26 & LatentVLA~\cite{xie2026latentvla} & \paperlink{https://arxiv.org/abs/2601.05611} & \qquad - & action & Qwen2.5-VL-3B & latent action model & feature generation & planning/autonomous driving \\
01/26 & LaViT~\cite{wu2026lavit} & \paperlink{https://arxiv.org/abs/2601.10129} & \githublink{https://github.com/Svardfox/LaViT} & vision & Qwen2.5-VL-3B & vision teacher model & chain generation & visual reasoning/efficiency/robustness \\
01/26 & RoT~\cite{wang2026renderofthought} & \paperlink{https://arxiv.org/abs/2601.14750} & \githublink{https://github.com/TencentBAC/RoT} & vision & Qwen3-VL-4B/other 2 & vision encoder & vision projection & visual reasoning/accelerating/efficiency \\
02/26 & MM-CoT~\cite{shao2026learning} & \paperlink{https://arxiv.org/abs/2602.00574} & \qquad - & vision & Qwen2.5-VL-7B & diffusion model & vision reconstruction & visual imagination/visual understanding  \\
02/26 & VaLR~\cite{jeon2026vision} & \paperlink{https://arxiv.org/abs/2602.04476} & \qquad - & vision & Qwen2.5-VL-7B & vision encoder & vision alignment & visual imagination/visual reasoning \\
02/26 & HIVE~\cite{zhang2026multimodal} & \paperlink{https://arxiv.org/abs/2602.05359} & \qquad - & vision & Huginn-3.5B & vision encoder & vision injection & visual reasoning/efficiency \\
02/26 & DW-VLA~\cite{liu2026driveworld} & \paperlink{https://arxiv.org/abs/2602.06521} & \qquad - & action & InternVL3-2B & extra encoder & vision projection & planning/autonomous driving \\
02/26 & OneLatent~\cite{lv2026onelatent} & \paperlink{https://arxiv.org/abs/2602.13738} & \qquad - & vision & DeepSeek-MoE-3B & vision encoder & vision projection & visual reasoning/efficiency \\
02/26 & Future-VLA~\cite{fan2026future} & \paperlink{https://arxiv.org/abs/2602.15882} & \qquad - & action & Qwen3-VL-4B & vision encoder & feature generation & embodied planning/efficiency \\
03/26 & LaST-VLA~\cite{luo2026last} & \paperlink{https://arxiv.org/abs/2603.01928} & \githublink{https://github.com/luo-yc17/LaST-VLA} & action & InternVL3-2B/8B & vision model & feature generation & spatial-temporal reasoning \\
\bottomrule
\end{tabular}}
\label{tab:architecture_module}
\end{table*}

\subsubsection{Component}
\label{sec:component}
This paradigm preserves the original backbone architecture but augments it with functional modules that construct, transform, store, or retrieve latent representations. Formally, a component produces and is then injected into the backbone decoding process:
\begin{equation}
    \mathbf{z} = \Phi^{\mathrm{comp}}\left(\mathbf{h}, \mathbf{x}\right),
\end{equation}
where the backbone $\Phi^{back}(\cdot)$ remains the principal generator, while the extra component $\Phi^{\mathrm{comp}}(\cdot)$ acts as a plug-in operator over latent space, and the output $\mathbf{z}$ will be used in Equation~\ref{eq:latent_autogressive}.

Such a design preserves the backbone architecture while equipping it with a latent mechanism to facilitate or guide downstream generation.
It covers a broad family of modules that operate on latent spaces while leaving the backbone largely frozen. We categorize existing approaches into five component families: \textbf{Generation}, \textbf{Projection}, \textbf{Alignment}, \textbf{Control}, and \textbf{Storage}, in view of their functional characteristics.

\xhdr{Generation Component} 
A major line of this part aims to construct intermediate latent states in hidden space, allowing the model to synthesize new implicit objectives, subgoals, or reasoning states that are not previously available as explicit symbols. 
A representative paradigm is realized as discrete tokens or a soft chain evolving rather than through explicit verbalized chains. For instance, ETD~\cite{koishekenov2025encode} introduces an encode–think–decode mechanism that shifts part of the process into latent computation without requiring explicit verbalization. At the same time, Palette~\cite{long2025reasoning}, iCLP~\cite{chen2025iclp}, ATP-Latent~\cite{zheng2026beyond}, and ReGuLaR~\cite{wang2026regular} employ VAE-style components to modulate high-level contexts and encourage diverse exploration. At a coarser granularity, JEPA-Reasoner~\cite{liu2025jepareasoner} formulates reasoning as a chain of latent predictions under a JEPA-style framework, thereby decoupling latent reasoning from surface token generation.

Beyond latent trajectory generation, another line of work leverages non-trajectory latent signals, \textit{e.g.}, compressed embeddings, activation-space features, or learned steering vectors. CoLaR~\cite{tan2025think}, for example, enables more efficient silent reasoning through embedding prediction and control. More broadly, AR~\cite{lukas2025activationreasoning} and RB-CoT~\cite{he2026reasoning} both employ SAE to generate interpretable feature directions in activation space, while PILOT~\cite{zheng2026pilot} synthesizes latent guidance vectors via an MLP to steer decoding. Additionally, MemGen~\cite{zhang2025memgen}, VisMem~\cite{yu2025vismem}, and LatentMem~\cite{fu2026latentmem} also adopt LoRA attached to the backbone to weave latent memories as an injection.

This idea has also been extended to the multimodal setting, where latent variables serve not only as semantic states but also as vision carriers. Recent works, such as AURORA~\cite{perception2025bigverdi}, introduce a variational autoencoder to synthesize perception tokens, enabling richer visual understanding. CoMEM~\cite{wu2025towards} and its agent variant CoMEM-Agent~\cite{wu2025autoscaling} deploy querying transformers to produce compact visual memory tokens, showing potential in long-horizon tasks.
In addition, SteerVAD~\cite{cai2026steering} generates related latent frame-level vision embeddings from mixed layers for video-based tasks, and Latent Sketchpad~\cite{zhang2025latentsketchpad} uses a VAE decoder to generate intermediate visual imaginations during multi-step visual reasoning.

\xhdr{Projection Component}
This category does not primarily synthesize new latent content, but instead improves reasoning or control by projecting existing internal representations into a different target space.
Several works attach lightweight linear layers, \textit{e.g.}, SoftCoT++~\cite{xu2025softcot1} and PLaT~\cite{wang2026latent}, or MLPs, such as SpiralThinker~\cite{piao2025spiralthinker} and LiteReason~\cite{gurung2025lightweight}, that project hidden states into a target semantic space. Besides, LF-Steering~\cite{yang2025lfsteering} trains an SAE to project activations into a semantic subspace to ensure semantic consistency.

In another typical path, projection serves as a bridge across modalities or agents. Wormhole~\cite{liu2026vision} connects heterogeneous visual agents through an encoder–decoder bridge that projects each agent's visual representation into a shared latent space. OccVLA~\cite{liu2025occvla} projects 3D spatial features via a transformer adapter, improving unified multimodal understanding.
Furthermore, as embodied scenarios advance, more researchers are focusing on action projection. For instance, LCLA~\cite{subedi2026lcla} and LaST0~\cite{liu2026last0} attach MLP projectors in embodied manipulation.
ThinkAct~\cite{huang2025thinkact} and Fast-ThinkAct~\cite{huang2026fastthinkact} both attach Q-former and MLP heads to project visual reasoning traces into robot action spaces. 
Combining both, VITA~\cite{ma2025unifying} adopts an encoder–decoder architecture to encode visual-action context into one latent representation, unifying perception and control in manipulation, while ColaVLA~\cite{peng2025colavla} introduces a forward layer that projects visual features into the action-planning space.

\xhdr{Alignment Component} While generation creates and projection reshapes, alignment ensures the transformed latent is anchored to something meaningful. These components enforce correspondence between latent representations and external grounding signals, whether visual features, task semantics, or cross-domain knowledge. 
AlignVLM~\cite{ahmed2025alignvlm} redesigns the cross-modal fusion layers, introducing mixed alignment layers that more faithfully bind visual tokens to textual semantics, while PREGEN~\cite{serussi2026pregen} aligns generated video embeddings to retrieval-relevant textual semantics.
Aligned with generative priors, LaDiR~\cite{kang2025ladir} and LoLA~\cite{wang2025lola} align hidden states into diffusion-compatible or transformer-based semantic spaces, ensuring consistency with desirable grounding signals. Extending to cross-domain scenes, Interpreter~\cite {kamai2026talking} uses LoRA to transfer latent abilities across domains, aligning source-domain competencies with target-domain representations.

\xhdr{Control Component} It determines when and how the model enters, exits, or delegates latent modes. Rather than generating content, they act as meta-level switches or routers that adaptively modulate the generation process.
To provide switching signals, FR-Ponder~\cite{he2025learning} trains a small MLP gating head to predict whether a given input requires extended latent pondering, and TaH~\cite{fu2025tah} positions a small MLP-style decider at each layer that votes on whether to enter a latent deliberation phase, enabling dynamic budget allocation in latent manifolds. 
MemGen~\cite{zhang2025memgen} incorporates a trigger, implemented via LoRA or entropy signals, that determines when memory should be invoked, while Kelp~\cite{li2025kelp} similarly uses an MLP-based module for risk-sensitive inputs in safety scenarios.

\xhdr{Storage Component} Storage components maintain persistent latent states across steps, turns, or episodes, enabling models to accumulate, compress, and retrieve information without relying on the finite context window.
IMM~\cite{jos2025beyond} and L2-VMAS~\cite{yu2026dual} introduce a differentiable vector library that acts as a latent memory bank, and G-MemLLM~\cite{xu2026gmemllm} scales this with a gated write–read logic that selectively updates memory. Further, PolarMem~\cite{chen2026polarmem} replaces the flat vector store with a graph-topology structure that clusters and links visual memory across episodes.

\xhdr{Summary} Component-based methods preserve the backbone architecture while augmenting it with plug-in latent modules that construct, transform, align, control, or store internal representations for downstream generation. Rather than replacing the backbone, these components operate as functional operators over latent space, enhancing reasoning, grounding, controllability, and memory with minimal architectural disruption. Existing approaches can be broadly organized into five families: generation components that synthesize latent reasoning states; projection components that map hidden representations into task-relevant spaces; alignment components that anchor latents to external semantics or priors; control components that regulate latent computation adaptively; and storage components that maintain persistent latent storage.

\subsubsection{Auxiliary Model}
\label{sec:auxiliary_model}
Here, the latent guidance signal is introduced by an external auxiliary model, rather than being natively induced within the backbone or instantiated by an internal functional component. Concretely, an auxiliary model first produces a latent representation and then is used to condition, guide, or refine the generation process of the host model:
\begin{equation}
    \mathbf{z} = \Phi^{\mathrm{aux}}(\mathbf{x}),
\end{equation}
where the latent introduction is outsourced to auxiliary model $\Phi^{aux}(\cdot)$ to replenish the host model $\Phi^{back}(\cdot)$, then $\mathbf{z}$ will be injected into the autoregressive process in Equation~\ref{eq:latent_autogressive}.

This paradigm introduces a functional division of labour: the host model retains responsibility for downstream prediction, while the auxiliary model either shapes its learning objective or enriches its internal representations. The resulting works bifurcate cleanly into two families according to what the auxiliary model contributes: \textbf{Supervision-oriented} approaches that provide training signals, and \textbf{Feature-oriented} approaches that provide intermediate representations.

\xhdr{Supervision-oriented Auxiliary Model} For the methods in this paradigm, the auxiliary models supply signals that guide the host model with semantic constraints, regularization, and supervision. 
The most direct strategy is to leverage another language model serving as a teacher to generate expressive traces. For instance, HCoT~\cite{liu2024expediting} and LaViT~\cite{wu2026lavit} instantiate assistant models to generate explicit reasoning chains, whose internal representations are then distilled into the host hidden states. SoftCoT~\cite{xu2025softcot} extends this paradigm by projecting the auxiliary chain into a continuous latent embedding rather than discrete token sequences, yielding soft supervision compatible with gradient-based fine-tuning. 
For finer-grained supervision, some methods focus on aligning the host latent representations with those of an external reference model. SIM-CoT~\cite{wei2025simcot} and SemCoT~\cite{he2025semcot} couple addition models whose hidden states function as semantic anchors, and train the host to reproduce these representations within its own latent space via a contrastive objective.

Another group treats the auxiliary model as a state generator whose outputs define structured training targets over latent trajectories. CTRLS~\cite{wu2025ctrls} uses a small LLM to synthesise intermediate reasoning states that serve as exploration waypoints within the host latent space. LatentEvolve~\cite{zhang2025latentevolve} extends this idea to evolutionary optimisation, employing an auxiliary model to generate candidate token sequences that drive iterative refinement of memory representations. CoLT~\cite{zhu2026colt} takes a more task-specific stance, assigning a tiny-scale auxiliary model to decompose latent tool-calling states into interpretable intermediate representations that supervise the internal planning phase.

\xhdr{Feature-oriented Auxiliary Model} These auxiliary models shift the locus to features available for the host model, generating and injecting intermediate representations directly into the computation graph. 
This paradigm is especially prevalent in multimodal and embodied settings, where the heterogeneity between modalities, \textit{i.e.}, language, vision, and action, creates representational bottlenecks. 
A sizeable cohort of methods employs dedicated vision models whose outputs supplement or supplant the host's own visual representations. 
CoVT~\cite{qin2025chainofvisualthought} frames this injection as a chain of visual thought, wherein auxiliary vision models iteratively construct intermediate visual representations analogous to the token-level reasoning steps of textual CoT.
3DThinker~\cite{chen2025think} pairs a specialised 3D foundation model that provides spatially-grounded geometric priors. Further works, including LaRe~\cite{ma2025multimodal}, MM-CoT~\cite{shao2026learning}, and VaLR~\cite{jeon2026vision}, leverage diffusion-architected generative models. Given a visual input, they reconstruct or imagine visual states whose latent codes are fed to the host model as an enriched perceptual context, enabling forms of visual imagination. 
OneLatent~\cite{lv2026onelatent}  and RoT~\cite{wang2026renderofthought} use a vision encoder to render textual reasoning into latent visual spaces. 
In addition, SkiLa~\cite{tong2025sketchinlatents} and VL-JEPA~\cite{chen2025vljepa} further explore vision projection in generative and self-supervised regimes, respectively.

In embodied and autonomous systems, the role of the auxiliary model shifts to policy grounding and environment perception. UniVLA~\cite{bu2025univla} incorporates a dedicated vision encoder to generate task-specific feature sequences, and SSM-VLA~\cite{cai2025seeing} introduces a VLM-based auxiliary that bridges visual perception and motor action within a state-space framework. 
SwiftVLA~\cite{ni2025swiftvla} and LaST-VLA~\cite{luo2026last} introduce spatial-temporal information through vision auxiliary, proving particularly beneficial for long-horizon manipulation tasks that demand coherent temporal reasoning.
In autonomous driving, LCDrive~\cite{tan2025latent} and DW-VLA~\cite{liu2026driveworld} incorporate world models and extra vision encoders, respectively, as auxiliary feature sources, grounding consistent representations of scene dynamics. Future-VLA~\cite{fan2026future} takes this further by conditioning action generation on auxiliary-predicted future visual features, effectively rendering the auxiliary model as a predictive look-ahead mechanism.
In addition, WholeBodyVLA~\cite{jiang2025wholebodyvla} and LatentVLA~\cite{xie2026latentvla} extend this to whole-body humanoid control by employing a latent action model as a feature encoder whose representations capture joint-level coordination structure.

\xhdr{Summary} The auxiliary-model paradigm introduces latent guidance through an external model rather than the backbone itself. Such auxiliary models either provide supervision signals to shape the backbone’s latent space or supply intermediate features that enrich its internal computation, making this paradigm particularly effective for complex reasoning, multimodal understanding, and embodied decision-making.
\subsection{Representation}
\label{sec:representation}

The transition from discrete token sequences to the continuous latent space $\mathcal{H}$ necessitates a precise definition of the core information carrier: the latent representation $\mathbf{z} \in \mathcal{H}$. Unlike discrete tokens $\mathbf{x} \in \mathcal{V}$, which are constrained to a predefined vocabulary, latent representations reside on a continuous, high-dimensional manifold, enabling substantially richer semantic expressivity~\cite{yoshua2013representation,hao2024training,xu2025softcot}.

\begin{figure*}[t]
  \centering
    \includegraphics[width=0.9\linewidth]{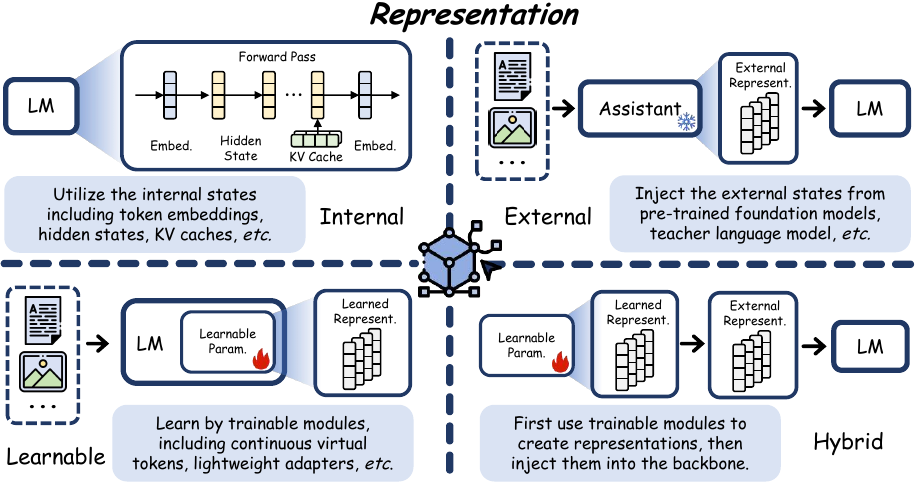}
    \caption{The schematic diagram of \textbf{Representation} mechanism, including four sub-types: \textbf{Internal} (Section~\ref{sec:internal}), \textbf{External} (Section~\ref{sec:external}), \textbf{Learnable} (Section~\ref{sec:learnable}), and \textbf{Hybrid} (Section~\ref{sec:hybrid}).}
    \label{fig:representation}
\end{figure*}

Two central questions motivate the study of these methods: \textit{what information does a latent representation encode, and how is it integrated into the generative pipeline?} The answers fundamentally determine the representational capacity, training dynamics, and generalization behavior of the resulting system. To provide a coherent organizational framework, we propose a taxonomy that classifies existing methods along two orthogonal axes: the \textbf{subject} of the representation (how $\mathbf{z}$ is structurally constructed, \textit{i.e.}, whether it is computed natively within the backbone or generated by a structurally independent module) and its \textbf{parameterization} (whether the construction process relies on fixed model states or incorporates dedicated trainable modules). As illustrated in Figure~\ref{fig:representation} and Table~\ref{tab:representation}, the intersection of these axes yields a comprehensive taxonomy comprising four distinct paradigms, detailed as follows:

\vspace{5pt}
\begin{tcolorbox}[
  colback=secblue!5,
  colframe=secblue!50,
  colbacktitle=secblue!50,
  coltitle=black,
  title={\textbf{\textcolor{secblue}{Mechanism:} Representation}},
  boxrule=5pt,
  arc=5pt,
  drop shadow,
  parbox=false,
  before skip=5pt,
  after skip=10pt,
  left=5pt,
  right=20pt,
]
\begin{itemize}
    \item \textbf{Internal} (Section~\ref{sec:internal}): operates directly on activations produced during the backbone's forward pass, including token embeddings, intermediate hidden states, and KV caches.
    \item \textbf{External} (Section~\ref{sec:external}): derived from a structurally independent auxiliary system (\textit{e.g.}, a pre-trained encoder), and injects these exogenous signals into the backbone as conditioning inputs or supervision targets while the auxiliary source remains frozen.
    \item \textbf{Learnable} (Section~\ref{sec:learnable}): constructed by dedicated trainable modules (\textit{e.g.}, continuous virtual tokens or lightweight adapters) that are embedded directly into the backbone and optimized end-to-end under specific task objectives.
    \item \textbf{Hybrid} (Section~\ref{sec:hybrid}): combines the Learnable and External paradigms sequentially by first using trainable modules to create specialized representations, then injecting these states as exogenous signals into the backbone for conditioning or latent supervision.
\end{itemize}
\end{tcolorbox}

\begin{table*}[!t]
\centering
\caption{Overview of the \textbf{Internal} (Section~\ref{sec:internal}), \textbf{External} (Section~\ref{sec:external}), \textbf{Learnable} (Section~\ref{sec:learnable}), and \textbf{Hybrid} (Section~\ref{sec:hybrid}) representation. We compare the modality, backbone, representation subject, and scenario.}
\setlength{\tabcolsep}{0.9mm}
\resizebox{1\textwidth}{!}{
\begin{tabular}{l|l|ll|llll}
\toprule
\textbf{Date} & \textbf{Method} & \textbf{Paper} & \textbf{Code} & \textbf{Modality} & \textbf{Backbone}  & \textbf{Subject} & \textbf{Scenario}   \\ \midrule

\multicolumn{7}{l}{\textbf{Internal}} \\\midrule
10/24 & CoE~\cite{wang2025latent} & \paperlink{https://arxiv.org/abs/2410.13640} & \githublink{https://github.com/Alsace08/Chain-of-Embedding} & text & LLaMA2-7B/other 6 & embedding/hidden state & label-free self-evaluation \\
10/24 & VTI~\cite{liu2024reducing} & \paperlink{https://arxiv.org/abs/2410.15778} & \githublink{https://github.com/shengliu66/VTI} & vision & LLaVA1.5-7B/other 2 & hidden state & hallucination mitigation \\
12/24 & COCONUT~\cite{hao2024training} & \paperlink{https://arxiv.org/abs/2412.06769} & \githublink{https://github.com/facebookresearch/coconut} & text & GPT2-0.1B & the last hidden state & latent reasoning/efficiency \\
05/25 & Soft Thinking~\cite{zhang2025soft} & \paperlink{https://arxiv.org/abs/2505.15778} & \qquad - & text & LLaMA3.1-8B/other 3 & weighted embedding & latent reasoning/efficiency \\
05/25 & CoT2~\cite{gozeten2025continuous} & \paperlink{https://arxiv.org/abs/2505.23648} & \githublink{https://github.com/alperengozeten/CoT2} & text & GPT2-0.1B & weighted embedding & exploration/complex reasoning \\
05/25 & CGC-VTD~\cite{wang2025image} & \paperlink{https://arxiv.org/abs/2505.21547} & \githublink{https://github.com/weixingW/CGC-VTD/tree/main} & vision & Chameleon-7B/other 2 & hidden state & hallucination mitigation \\
08/25 & LFJ~\cite{xing2025latent} & \paperlink{https://arxiv.org/abs/2508.10029} & \qquad - & text & Vicuna-7B/other 4 & hidden state & jailbreak attack \\
09/25 & SIM-CoT~\cite{wei2025simcot} & \paperlink{https://arxiv.org/abs/2509.20317} & \githublink{https://github.com/InternLM/SIM-CoT} & text & LaMA3.2-1B/3B/8B & hidden state & latent reasoning/efficiency \\
10/25 & LatentBreak~\cite{mura2025latentbreak} & \paperlink{https://arxiv.org/abs/2510.08604} & \qquad - & text & LLaMA2-7B/other 8 & hidden state & jailbreak attack \\
10/25 & SALS~\cite{mu2025sals} & \paperlink{https://arxiv.org/abs/2510.24273} & \qquad - & text & LLaMA2-7B/Mistral-7B & compressed kv-cache & accelerating/efficiency \\
10/25 & IVT-LR~\cite{chen2025reasoning} & \paperlink{https://arxiv.org/abs/2510.12603} & \githublink{https://github.com/FYYDCC/IVT-LR} & vision & Qwen2-VL-7B/Chameleon-7B & image embedding/hidden state & visual reasoning/efficiency \\
11/25 & LRP~\cite{yao2025reading} & \paperlink{https://arxiv.org/abs/2511.19806} & \qquad - & vision & Qwen-VL-2.5/Gemma3-12B & hidden state/attention weight & abstention/generalization \\
11/25 & LatentMAS~\cite{zou2025latent} & \paperlink{https://arxiv.org/abs/2511.20639} & \githublink{https://github.com/Gen-Verse/LatentMAS} & text & Qwen3-4B/8B/14B & hidden state/kv-cache & multi-agent collaboration \\
01/26 & CLReg~\cite{tang2026from} & \paperlink{https://arxiv.org/abs/2601.22028} & \qquad - & text & LLaMA3.1-8B/other 2 & hidden state & forgetting mitigation \\
01/26 & CausalEmbed~\cite{huo2026causalembed} & \paperlink{https://arxiv.org/abs/2601.21262} & \qquad - & vision & PaliGemma-3B/Qwen2.5-VL-3B & the last hidden state & visual document retrieval \\
02/26 & LT-Tuning~\cite{liu2026latent} & \paperlink{https://arxiv.org/abs/2602.10229} & \githublink{https://github.com/NeosKnight233/Latent-Thoughts-Tuning} & text & LLaMA3.2-1B/3B/8B & weighted embedding/hidden state & latent reasoning/dynamic reasoning \\
02/26 & JLT~\cite{kadali2026jailbreaking} & \paperlink{https://arxiv.org/abs/2602.11495} & \qquad - & text & LLaMA3.1-8B & hidden state & jailbreak defense \\
02/26 & ThinkRouter~\cite{xu2026thinkrouter} & \paperlink{https://arxiv.org/abs/2602.11683} & \qquad - & text & Qwen3-8B/other 3 & weighted embedding & dynamic reasoning/efficiency \\
\midrule

\multicolumn{7}{l}{\textbf{External}} \\\midrule
02/25 & CODI~\cite{shen2025codi} & \paperlink{https://arxiv.org/abs/2502.21074} & \githublink{https://github.com/zhenyi4/codi} & text & GPT2-0.1B/LLaMA3.2-1B & teacher hidden state & latent reasoning/efficiency \\
02/25 & SoftCoT~\cite{xu2025softcot} & \paperlink{https://arxiv.org/abs/2502.12134} & \githublink{https://github.com/xuyige/SoftCoT} & text & LLaMA3.1-8B & assistant hidden state & efficiency/generalization/zero-shot \\
07/25 & GoK~\cite{bystronski2025large} & \paperlink{https://arxiv.org/abs/2507.13874} & \qquad - & text & Mistral-7B & pretrained embedding & exploration/diverse generation \\
09/25 & OccVLA~\cite{liu2025occvla} & \paperlink{https://arxiv.org/abs/2509.05578} & \qquad - & action & PaliGemma2-3B & pretrained  3D occupancy token & spatial understanding \\
10/25 & KaVa~\cite{kuzina2025kava} & \paperlink{https://arxiv.org/abs/2510.02312} & \qquad - & text & LLaMA3.2-1B/other 2 & teacher compressed kv-cache & latent reasoning/efficiency \\
10/25 & 3DThinker~\cite{chen2025think} & \paperlink{https://arxiv.org/abs/2510.18632} & \githublink{https://github.com/zhangquanchen/3DThinker} & vision & Qwen2.5-VL-7B/other 7 & pretrained 3D token  & spatial understanding/visual reasoning \\
11/25 & COVT~\cite{qin2025chainofvisualthought} & \paperlink{https://arxiv.org/abs/2511.19418} & \githublink{https://github.com/Wakals/CoVT} & vision & Qwen2.5-VL-7B & pretrained visual token & visual understanding/visual reasoning \\
12/25 & Mull-Tokens~\cite{ray2025mulltokens} & \paperlink{https://arxiv.org/abs/2512.10941} & \qquad - & vision & Qwen2.5-VL-7B & pretrained multimodal token & latent reasoning \\
12/25 & LCDrive~\cite{tan2025latent} & \paperlink{https://arxiv.org/abs/2512.10226} & \qquad - & action & Qwen3-0.5B & pretrained action/world model token & autonomous driving/efficiency \\
12/25 & SkiLa~\cite{tong2025sketchinlatents} & \paperlink{https://arxiv.org/abs/2512.16584} & \githublink{https://github.com/TungChintao/SkiLa} & vision & Qwen2.5-VL-7B & pretrained sketch token & visual imagination/understanding \\
12/25 & VL-JEPA~\cite{chen2025vljepa} & \paperlink{https://arxiv.org/abs/2512.10942} & \qquad - & vision & LLaMA3.2-1B & pretrained text embedding & classification/retrieval \\
02/26 & LaRA-VLA~\cite{bai2026latent} & \paperlink{https://arxiv.org/abs/2602.01166} & \githublink{https://github.com/LoveJu1y/LaRA-VLA} & action & Qwen3-VL-8B & pretrained visual/action token & embodied manipulation \\
02/26 & OneLatent~\cite{lv2026onelatent} & \paperlink{https://arxiv.org/abs/2602.13738} & \qquad - & vision & DeepSeek-MoE-3B & pretrained hidden state & visual reasoning/efficiency \\
03/26 & LaSER~\cite{jin2026laser} & \paperlink{https://arxiv.org/abs/2603.01425} & \qquad - & text & Qwen3-0.6B/other 5 & teacher hidden state & complex reasoning/dense retrieval \\
\midrule

\multicolumn{7}{l}{\textbf{Learnable}} \\\midrule
05/25 & CoLaR~\cite{tan2025think} & \paperlink{https://arxiv.org/abs/2505.16552} & \githublink{https://github.com/xiaomi-research/colar} & text & LLaMA3.2-1B & compressed reasoning embedding & dynamic reasoning/efficiency \\
05/25 & Comma~\cite{sun2025enhancing} & \paperlink{https://arxiv.org/abs/2505.12629} & \qquad - & text & LLaMA3.2-1B & learnable latent token & generalization \\
07/25 & CTRLS~\cite{wu2025ctrls} & \paperlink{https://arxiv.org/pdf/2507.08182} & \qquad - & text & LLaMA3.2-3B/Qwen2.5-3B & iteratively updated latent states & exploration/latent reasoning \\
08/25 & MCOUT~\cite{pham2025multimodal} & \paperlink{https://arxiv.org/abs/2508.12587} & \qquad - & vision & LLaMA3.2-1B & fused  thought embedding & visual reasoning/latent reasoning \\
09/25 & LatentGuard~\cite{shu2025latentguard} & \paperlink{https://arxiv.org/abs/2509.19839} & \qquad - & text & Gemini2.5 Pro & disentangled latent code & security/attacks refusal \\
09/25 & MARCOS~\cite{liu2025marcos} & \paperlink{https://arxiv.org/abs/2509.25020} & \qquad - & text & Qwen2.5-0.5B/3B & iteratively updated latent states & latent reasoning/efficiency \\
10/25 & C2C~\cite{fu2025cache} & \paperlink{https://arxiv.org/abs/2510.03215} & \githublink{https://github.com/thu-nics/C2C} & text & Qwen3-0.6B/Qwen2.5-0.5B & fused kv-cache & multi-agent collaboration \\
11/25 & Interlat~\cite{du2025enabling} & \paperlink{https://arxiv.org/abs/2511.09149} & \qquad - & text & Qwen2.5-7B/other 2 & adapted hidden state & multi-agent collaboration \\
12/25 & LIVR~\cite{li2025latent1} & \paperlink{https://arxiv.org/abs/2512.21218} & \qquad - & vision & Qwen3-VL-4B & learnable latent token & visual reasoning/multi-task \\
01/26 & KVCA~\cite{dery2026latent} & \paperlink{https://arxiv.org/abs/2601.06123} & \qquad - & text & Gemma2-2B & aligned kv-cache & inter-agent communication \\
01/26 & TCLA~\cite{nikulin2026visionlanguage} & \paperlink{https://arxiv.org/abs/2601.22714} & \qquad - & action & Qwen2.5-VL-7B/other 23 & learnable latent action token & embodied manipulation \\
01/26 & UniCog~\cite{liu2026unicog} & \paperlink{https://arxiv.org/abs/2601.17897} & \githublink{https://github.com/milksalute/unicog} & text & Qwen3-8B/other 3 & learnable latent mind & cognitive reasoning/interpretability \\
02/26 & CoLT~\cite{zhu2026colt} & \paperlink{https://arxiv.org/abs/2602.04246} & \qquad - & text & LLaMA3.2-1B & latent tool calling & tool calling/efficiency \\
02/26 & DeltaKV~\cite{hao2026deltakv} & \paperlink{https://arxiv.org/abs/2602.08005} & \githublink{https://github.com/CURRENTF/Sparse-vLLM} & text & LLaMA3.1-8B/other 3 & compressed kv-cache & complex reasoning/efficiency \\
02/26 & MAS4TS~\cite{ruan2026visual} & \paperlink{https://arxiv.org/pdf/2602.03026} & \qquad - & text & Qwen3-VL-235B & reconstructed  latent trajectory & visual reasoning/time series analysis \\
\midrule

\multicolumn{7}{l}{\textbf{Hybrid}} \\\midrule
09/24 & HCoT~\cite{liu2024expediting} & \paperlink{https://arxiv.org/abs/2409.08561} & \qquad - & text & LLaMA2-7B/13B & special CoT token & efficiency/generalization \\
12/24 & AURORA~\cite{perception2025bigverdi} & \paperlink{https://arxiv.org/abs/2412.03548} & \githublink{https://github.com/mahtabbigverdi/Aurora-perception} & vision & LLaVA1.5-13B & discrete perception token & visual perception \\
12/24 & DCA~\cite{liu2025deliberation} & \paperlink{https://arxiv.org/abs/2412.17747} & \qquad - & text & Gemma2-2B & augmented kv-cache & latent reasoning/zero-shot \\
02/25 & Assorted~\cite{su2025token} & \paperlink{https://arxiv.org/abs/2502.03275} & \qquad - & text & LLaMA3.2-1B/3B/8B & compressed CoT token & latent reasoning/efficiency \\
05/25 & CoMEM~\cite{wu2025towards} & \paperlink{https://arxiv.org/abs/2505.17670} & \githublink{https://github.com/WenyiWU0111/CoMEM/tree/main} & vision & Qwen2-VL-7B/Qwen2.5-VL-7B & latent memory & visual reasoning/long-horizon/efficiency \\
05/25 & UniVLA~\cite{bu2025univla} & \paperlink{https://arxiv.org/abs/2505.06111} & \githublink{https://github.com/OpenDriveLab/UniVLA} & action & Prismatic-7B & latent action token & embodied manipulation/scaling \\
07/25 & DEP~\cite{qiu2025latent} & \paperlink{https://arxiv.org/abs/2507.20849} & \githublink{https://github.com/SnowCharmQ/DEP} & text & Qwen2.5-7B/32B & compressed embedding & personalization \\
09/25 & GainRouter~\cite{zheng2025fast} & \paperlink{https://arxiv.org/abs/2509.23633} & \qquad - & text & Qwen3-4B & latent codebook of strategy priors & latent reasoning/adaptive reasoning \\
09/25 & MemGen~\cite{zhang2025memgen} & \paperlink{https://arxiv.org/abs/2509.24704} & \githublink{https://github.com/bingreeky/MemGen} & text & SmolLM3-3B/Qwen3-8B & latent memory & experimental memory/generalization \\
10/25 & Latent-SFT~\cite{deng2025latent} & \paperlink{https://arxiv.org/abs/2510.15522} & \githublink{https://github.com/DJC-GO-SOLO/Latent-SFT} & text & LLaMA3.2-1B/other 2 & compressed CoT token & latent reasoning/efficiency\\
10/25 & ThoughtComm~\cite{zheng2025thought} & \paperlink{https://arxiv.org/abs/2510.20733} & \qquad - & text & Qwen3-0.6B/other 4 & latent thought & multi-agent collaboration \\
11/25 & CLaRa~\cite{he2025clara} & \paperlink{https://arxiv.org/abs/2511.18659} & \githublink{https://github.com/apple/ml-clara} & text & Mistral-7B/Phi4-Mini-3.8B & compressed representation & retrieval/efficiency \\
11/25 & EBM-CoT~\cite{chen2025think1} & \paperlink{https://arxiv.org/abs/2511.07124} & \qquad - & text & Qwen2.5-7B/other 5 & latent thought & latent reasoning/efficiency \\
11/25 & Monet~\cite{wang2025monet} & \paperlink{https://arxiv.org/abs/2511.21395} & \githublink{https://github.com/NOVAglow646/} & vision & Qwen2.5-VL-7B & latent thought & visual reasoning/generalization \\
11/25 & VisMem~\cite{yu2025vismem} & \paperlink{https://arxiv.org/abs/2511.11007} & \githublink{https://github.com/YU-deep/VisMem} & vision & Qwen2.5-VL-7B/other 8 & latent visual memory & visual understanding/visual reasoning \\
11/25 & LatBot~\cite{li2025latbot} & \paperlink{https://arxiv.org/abs/2511.23034} & \qquad - & action & InternVL3.5-2B & latent action token & embodied manipulation \\
11/25 & VITA~\cite{ma2025unifying} & \paperlink{https://arxiv.org/abs/2511.19859} & \githublink{https://github.com/vita-cvpr26/vita} & action & PaliGemma-3B & dynamics-unified token & embodied manipulation \\
12/25 & iCLP~\cite{chen2025iclp} & \paperlink{https://arxiv.org/abs/2512.24014} & \githublink{https://github.com/AgenticFinLab/latent-planning} & text & Qwen2.5-0.5B/3B/7B & latent plan & latent planning/efficiency \\
12/25 & Motus~\cite{bi2025motus} & \paperlink{https://arxiv.org/abs/2512.13030} & \githublink{https://github.com/thu-ml/Motus} & action &  Qwen3-VL-2B & latent action token & embodied manipulation \\
01/26 & RB-CoT~\cite{he2026reasoning} & \paperlink{https://arxiv.org/abs/2601.08058} & \qquad - & text & LLaMA3.1-8B/other 5 &  prompt-sensitive latent token & multi-step reasoning/prompting \\
02/26 & LatentChem~\cite{ye2026latentchem} & \paperlink{https://arxiv.org/abs/2602.07075} & \githublink{https://github.com/xinwuye/LatentChem} & text & Qwen3-8B & latent thought/molecular token & chemical reasoning \\
02/26 & UniLACT~\cite{govind2026unilact} & \paperlink{https://arxiv.org/abs/2602.20231} & \qquad - & action & GPT2-0.1B & latent action token & embodied manipulation/generalization \\

\bottomrule
\end{tabular}}
\label{tab:representation}
\end{table*}

\subsubsection{Internal}
\label{sec:internal}

In the internal paradigm, $\mathbf{z}$ is derived exclusively from endogenous activations generated during the backbone's standard forward pass, without introducing any additional parameters. Let $\Phi^{\mathrm{back}}$ consist of $L$ transformer blocks decomposed as:
\begin{equation}
    \Phi^{\mathrm{back}}
    = \phi_{\mathrm{head}} \circ \phi_L \circ \cdots \circ
      \phi_l \circ \cdots \circ \phi_1 \circ \phi_{\mathrm{emb}},
\end{equation}
where $\phi_{\mathrm{emb}} : \mathcal{V} \to \mathbb{R}^d$ maps discrete tokens to continuous embeddings via the embedding matrix $\mathbf{E} \in \mathbb{R}^{|\mathcal{V}| \times d}$, each transformer block $\phi_l : \mathbb{R}^{d \times T} \to \mathbb{R}^{d \times T}$ processes the full sequence of $T$ tokens, and $\phi_{\mathrm{head}}$ projects the final hidden states back to the vocabulary space. Let $\mathbf{h}_l^t \in \mathbb{R}^d$ denote the hidden state of token $t$ at layer $l$, and let $\mathbf{H}_l = \bigl[\mathbf{h}_l^1, \dots, \mathbf{h}_l^T\bigr] \in \mathbb{R}^{T \times d}$ collect all token states at layer $l$. The latent representation is then a deterministic readout:
\begin{equation}
    \mathbf{z} = g\!\left(\{\mathbf{H}_l\}_{l\in S}\right),
\end{equation}
where $g(\cdot)$ is a parameter-free aggregation function applied over the set of layer activations $S$ (\textit{e.g.}, index selection, mean pooling across tokens, or a fixed linear combination over layers). Depending on which activation type is extracted, internal representations manifest in
three principal forms: \textbf{Hidden State}, \textbf{Weighted Embedding}, and \textbf{Cache}.

\xhdr{Internal Hidden State} 
As the most prevalent form, intermediate activations are extracted to provide a continuous summary of the model's evolving computation. Common instantiations include taking the last hidden state of position $T$, or computing a mean-pooled representation across all tokens at a particular layer $l$:
\begin{equation}
    \mathbf{z} = \mathbf{h}_L^T \quad \text{or} \quad  \frac{1}{T}\sum_{t=1}^{T} \mathbf{h}_l^t.
\end{equation}

For example, COCONUT~\cite{hao2024training} establishes a foundational pattern for continuous generation by feeding the last hidden state directly back as the next input embedding, bypassing discrete vocabulary projection and forming a recurrent loop of continuous ``thoughts''. SIM-CoT~\cite{wei2025simcot} and LatentMAS~\cite{zou2025latent} adopt this recurrent paradigm. Beyond direct generation, the rich semantic geometry of hidden states proves valuable in downstream applications. For instance, CoE~\cite{wang2025latent} constructs chains of embeddings from pooled hidden states to perform label-free self-evaluation. In multimodal settings, internal activations frequently serve as diagnostic and corrective signals: VTI~\cite{liu2024reducing} and CGC-VTD~\cite{wang2025image} dynamically probe intermediate hidden states to detect and mitigate visual hallucinations. Final hidden states have also been repurposed as compressed visual-semantic summaries: IVT-LR~\cite{chen2025reasoning} fuses them with image embeddings for efficient visual reasoning, while CausalEmbed~\cite{huo2026causalembed} projects them into dense representations for visual document retrieval. Because these states natively encode pre-trained knowledge, CLReg~\cite{tang2026from} further leverages them as a regularizer to mitigate catastrophic forgetting. The high expressivity of endogenous states is, however, a double-edged sword from a security perspective. On the offensive side, LFJ~\cite{xing2025latent} and LatentBreak~\cite{mura2025latentbreak} exploit hidden-state vulnerabilities to mount latent jailbreak attacks that bypass discrete text-level filters. Conversely, JLT~\cite{kadali2026jailbreaking} monitors these same signals defensively, using them to detect malicious intent and improve jailbreak robustness.

\xhdr{Internal Weighted Embedding} 
This form replaces hard token sampling with a soft, probability-weighted combination over the vocabulary embedding matrix $\mathbf{E} \in \mathbb{R}^{|\mathcal{V}| \times d}$. Let $\boldsymbol{\alpha} \in \mathbb{R}^{|\mathcal{V}|}$ denote a weight vector derived from the model's output logits (\textit{i.e.}, via softmax); the latent representation is:
\begin{equation}
    \mathbf{z} = \mathbf{E}^\top \boldsymbol{\alpha}, \qquad
    \boldsymbol{\alpha} = \mathrm{softmax}(\mathbf{o}),
\end{equation}
where $\mathbf{o} \in \mathbb{R}^{|\mathcal{V}|}$ denotes the pre-softmax logits. Because $\boldsymbol{\alpha}$ lies in the probability simplex, $\mathbf{z}$ is constrained to the convex hull of the vocabulary embedding vectors.

This differentiable relaxation forms a superposition of candidate token embeddings \cite{zhang2025soft, gozeten2025continuous}, enabling gradient flow through the discrete generation step and facilitating parallel exploration of the reasoning space. LT-Tuning~\cite{liu2026latent} builds on this property with a context-prediction fusion mechanism that exploits predictive semantic guidance from the vocabulary embedding space to construct robust latent thoughts. ThinkRouter~\cite{xu2026thinkrouter} further leverages the continuous nature of $\mathbf{z}$ for confidence-aware routing, dynamically switching between continuous latent thinking and discrete token generation based on model uncertainty.

\xhdr{Internal Cache} 
The third form treats accumulated key-value pairs as structured latent memory, enabling efficient context reuse without recomputation. Given the hidden states $\mathbf{H}_l \in \mathbb{R}^{T \times d}$, the cache representations are computed via learned projection matrices:
\begin{equation}
    \mathbf{k}_l = \mathbf{H}_l \mathbf{W}_l^k, \qquad
    \mathbf{v}_l = \mathbf{H}_l \mathbf{W}_l^v,
\end{equation}
where $\mathbf{W}_l^k, \mathbf{W}_l^v \in \mathbb{R}^{d \times d_k}$ project the sequence into the key and value spaces of the attention heads.

By treating these compressed tensors as operational memory, SALS~\cite{mu2025sals} exploits sparse attention patterns over the KV cache to accelerate inter-step computation and improve inference efficiency. LatentMAS~\cite{zou2025latent} extends this to multi-agent collaboration, using the KV cache as a shared, continuous working memory that enables agents to coordinate without explicit textual communication. In the multimodal domain, LRP~\cite{yao2025reading} combines attention weights and hidden states to capture richer visual semantics, demonstrating improved abstraction and generalization on complex visual tasks.

\xhdr{Summary} 
The internal paradigm converts endogenous model activations into parameter-free latent representations, entirely bypassing the discrete vocabulary bottleneck. By treating standard computational byproducts as versatile semantic assets, it demonstrates that intrinsic model states carry sufficient representational capacity to support continuous reasoning, accelerate inference, and enable robust downstream analysis.

\subsubsection{External}
\label{sec:external}

In the external paradigm, $\mathbf{z}$ originates from an auxiliary encoder $\Phi^{\mathrm{aux}}$ that is structurally independent of the backbone. Given an auxiliary input $\mathbf{x}_{\mathrm{aux}}$, which may coincide with $\mathbf{x}$ or belong to a different modality:
\begin{equation}
    \mathbf{z} = \Phi^{\mathrm{aux}}(\mathbf{x}_{\mathrm{aux}}),
\end{equation}
where $\Phi^{\mathrm{aux}}$ is kept frozen during backbone training. Because $\Phi^{\mathrm{aux}}$ and $\Phi^{\mathrm{back}}$ operate in distinct latent manifolds with potentially mismatched dimensionalities ($d_{\mathrm{aux}} \neq d_{\mathrm{back}}$), $\mathbf{z}$ requires explicit structural alignment before integration. Depending on its functional role, this paradigm operates in two modes.

First, as a conditioning input to guide backbone generation, $\mathbf{z}$ is first aligned to the backbone's native latent space via a learned projection $\psi : \mathbb{R}^{d_{\mathrm{aux}}} \to \mathbb{R}^{d_{\mathrm{back}}}$ (\textit{e.g.}, a linear map or MLP), and the backbone conditions its output on the projected signal:
\begin{equation}
    \hat{\mathbf{z}} = \psi(\mathbf{z}), \qquad
    \mathbf{y} \sim \Phi^{\mathrm{back}}\!\left(\cdot \mid [\hat{\mathbf{z}};\, \mathbf{x}]\right),
\end{equation}
where $[\hat{\mathbf{z}};\, \mathbf{x}]$ denotes a latent integration mechanism such as prefix concatenation or cross-attention injection.

Second, as a supervision target for representation alignment or knowledge distillation, $\mathbf{z}$ serves as a dense, non-verbalized supervision target. The backbone's internal state $\mathbf{h}$ is trained to match the projected external signal. This continuous supervision transfers rich semantic priors from $\Phi^{\mathrm{aux}}$ to the backbone without the information bottleneck imposed by discrete token generation.
Based on the modality and functional role of $\Phi^{\mathrm{aux}}$, we identify three directions: \textbf{Reasoning Priors}, \textbf{Perceptual Priors}, and \textbf{Embodied Priors}.

\xhdr{External Reasoning Priors}
Here, the auxiliary source operates in the semantic domain to supply the backbone with structured logic and cognitive traces. A prevalent strategy is knowledge distillation from an expert reasoning model. CODI~\cite{shen2025codi} formalizes this via a self-distillation loop in which the frozen teacher's hidden states serve as continuous supervision targets, guiding the implicit reasoning of the student. SoftCoT~\cite{xu2025softcot} extends this concept to explicit injection: a lightweight assistant model generates speculative reasoning chains that are projected into soft-token embeddings and prepended to the backbone input, creating a gradient-compatible conditioning signal that substantially improves zero-shot generalization. KaVa~\cite{kuzina2025kava} bypasses intermediate output tokens entirely by distilling the teacher's compressed KV cache, directly transferring structured attention states as
compact latent priors.

\xhdr{External Perceptual Priors}
Here, pre-trained vision encoders or specialized multimodal auxiliary models supply spatially, temporally, or structurally rich feature maps to augment a primarily textual backbone. 3DThinker~\cite{chen2025think} injects spatially grounded 3D tokens from a specialized
auxiliary network to supply geometric priors that the base model cannot derive from
two-dimensional inputs alone. SkiLa~\cite{tong2025sketchinlatents} leverages pre-trained sketch tokens to interleave intermediate visual thoughts with textual reasoning, explicitly grounding the model's spatial understanding. VL-JEPA~\cite{chen2025vljepa} employs pre-trained embeddings from a predictive-coding model in an abstract representation space to improve cross-modal classification and
retrieval. COVT~\cite{qin2025chainofvisualthought} uses an auxiliary vision model to iteratively
extract visual tokens and fuse them into the backbone to construct continuous visual thought chains. OneLatent~\cite{lv2026onelatent} demonstrates that hidden states from a strong vision-language model can condense rich perceptual and OCR context into a single latent token for highly efficient visual reasoning. Mull-Tokens~\cite{ray2025mulltokens} generalizes this injection with modality-agnostic latent tokens derived from an auxiliary system, enabling uniform latent reasoning across both linguistic and visual substrates.

\xhdr{External Embodied Priors}
This direction, prevalent in embodied models and autonomous driving, relies on auxiliary models to generate structured latent representations of environmental dynamics, 3D geometry, and future actions. OccVLA~\cite{liu2025occvla} applies pre-trained 3D occupancy tokens as a supervisory signal to impart fine-grained spatial understanding without requiring explicit 3D inputs at inference time. LCDrive~\cite{tan2025latent} incorporates an external world model to generate action and scene tokens that inject temporally consistent representations of scene dynamics, grounding the host model's navigational reasoning. LaRA-VLA~\cite{bai2026latent} bridges continuous perception and motor control by utilizing pre-trained visual and action tokens, enabling a smooth transition from explicit multimodal supervision to internalized latent reasoning for complex embodied manipulation.

\xhdr{Summary}
The external paradigm provides a principled approach to bridging modality gaps and
circumventing the discrete token bottleneck by injecting structured knowledge from
independent auxiliary systems. By aligning reasoning, perceptual, and embodied priors as either dynamic conditioning signals or dense supervision targets, it endows the backbone with continuous reasoning capabilities and complex structural constraints — without modifying the backbone's parameters. This demonstrates that a foundation model's representational scope is not strictly bounded by its native architecture: leveraging specialized latent priors offers a scalable pathway to expand semantic reach while keeping the backbone largely frozen.

\subsubsection{Learnable}
\label{sec:learnable}

In the learnable paradigm, $\mathbf{z}$ is actively constructed by a parameterized module $\Phi^{\mathrm{comp}}$ with learnable parameters $\theta$ that is directly embedded \textit{within} the backbone architecture (\textit{e.g.}, continuous virtual tokens or lightweight adapters). Let $\mathbf{c}$ denote an optional conditioning context, such as the input $\mathbf{x}$, intermediate hidden states $\mathbf{h}$, or the empty set ($\mathbf{c} = \emptyset$). The latent representation is formulated as:
\begin{equation}
    \mathbf{z} = \Phi^{\mathrm{comp}}(\mathbf{c};\, \theta).
\end{equation}
Unlike the external paradigm, $\Phi^{\mathrm{comp}}$ is structurally coupled with the backbone. The parameters $\theta$ are optimized end-to-end driven by specific task objectives (either independently with a frozen backbone or jointly with $\theta^{\mathrm{back}}$). Depending on the targeted optimization objective, learnable representations manifest in three primary forms: \textbf{Compression Learning}, \textbf{Distribution Learning}, and \textbf{Alignment Learning}.

\xhdr{Compression Learning}
The first class is driven by information-bottleneck principles, optimizing $\Phi^{\mathrm{comp}}$ to compress explicit data into dense, continuous vectors. CoLaR~\cite{tan2025think} learns to aggregate consecutive reasoning tokens into compressed embeddings using a variance-preserving scaling factor. CoLT~\cite{zhu2026colt} employs supervised learning to condense long reasoning trajectories into continuous seed tokens for parametric tool calls. In the multimodal domain, LIVR~\cite{li2025latent1} imposes a visual bottleneck during training, requiring the model to learn implicit spatial compressions without relying on explicit textual descriptions. At the memory level, DeltaKV~\cite{hao2026deltakv} encodes residual differences between successive cache states to substantially compress long-term reasoning overhead.

\xhdr{Distribution Learning}
The second class abandons deterministic point mappings, instead optimizing $\Phi^{\mathrm{comp}}$ to capture the underlying stochastic distributions and structural manifolds of reasoning. CTRLS~\cite{wu2025ctrls} formulates reasoning as a Markov decision process and models state transitions via Dirichlet distributions to capture epistemic uncertainty. MARCOS~\cite{liu2025marcos} employs a conditional hidden Markov model with step-level latent variables, relying on variational training to learn stochastic continuous representations. UniCog~\cite{liu2026unicog} formulates cognitive distributions as a latent variable model, optimizing an evidence lower bound to project activations into a high-dimensional sparse space. LatentGuard~\cite{shu2025latentguard} learns to model the latent space using a VAE, manipulating the resulting semantic distributions to enable robust refusal of adversarial inputs.

\xhdr{Alignment Learning}
The third class optimizes $\Phi^{\mathrm{comp}}$ to construct cross-space projections, bridging rigid boundaries such as distinct modalities or heterogeneous agent architectures. KVCA~\cite{dery2026latent} learns a globally shared latent manifold $\Sigma$ via cross-attention to translate KV caches between architecturally heterogeneous models. C2C~\cite{fu2025cache} trains a neural MLP to directly project KV caches between specific models by aligning their terminal-layer representations. Interlat~\cite{du2025enabling} optimizes communication adapters via a weighted Jensen-Shannon divergence to align transmitted hidden states with the receiving agent's internal representations.
This paradigm also bridges modalities: MCOUT~\cite{pham2025multimodal} learns cross-modal attention fusion while penalizing entropy collapse; MAS4TS~\cite{ruan2026visual} learns to reconstruct predictive numerical trajectories from visual time-series plots; and TCLA~\cite{nikulin2026visionlanguage} aligns noisy observational signals with clean, VLM-prompted latent action targets for embodied manipulation.

\xhdr{Summary} 
The learnable paradigm excels at acquiring latent structures explicitly optimized for
specific downstream objectives, breaking free from the rigid semantic boundaries imposed
by textual pre-training. This flexibility allows models to encode non-verbal modalities (\textit{e.g.}, complex spatial layouts and embodied actions) and support high-bandwidth inter-agent collaboration. However, granting unconstrained optimization access to continuous spaces introduces the risk of manifold overfitting: modules may memorize the noise in supervision signals, yielding highly specialized representations that sacrifice zero-shot generalization. Rigorous regularization, including enforced structural sparsity, variance-preserving normalizations, and controlled belief-shift tuning, which is therefore essential to maintain broad utility.

\subsubsection{Hybrid}
\label{sec:hybrid}

The hybrid paradigm utilizes a dedicated, structurally \textit{independent} module $\Phi^{\mathrm{comp}}$ with learnable parameters $\theta$ to construct a structured latent representation. Let $\mathbf{c}$ denote a conditioning context drawn from the input $\mathbf{x}$ or auxiliary features. The hybrid representation is formulated as:
\begin{equation}
    \mathbf{z} = \Phi^{\mathrm{comp}}(\mathbf{c};\, \theta).
\end{equation}
Crucially, unlike the Learnable paradigm, $\Phi^{\mathrm{comp}}$ remains architecturally disjoint from the backbone $\Phi^{\mathrm{back}}$. Once constructed, $\mathbf{z}$ is deployed in the same manner as the External paradigm: it acts either as an exogenous conditioning signal injected into a typically frozen backbone (serving as a learned alignment bridge), or as a rich, optimized supervision target for latent distillation. This approach can be summarized into three functional categories: \textbf{Traces}, \textbf{Grounding}, and \textbf{Augmentation}.

\xhdr{Traces}
The first branch distills discrete reasoning trajectories into compact continuous vectors that guide the backbone without the latency of explicit chain-of-thought generation. HCoT~\cite{liu2024expediting} compresses multi-step reasoning into a specialized thought token injected into a frozen decoder to accelerate inference. Assorted~\cite{su2025token} combines latent and text tokens by compressing reasoning segments via a VQ-VAE codebook. Latent-SFT~\cite{deng2025latent} restricts the latent space to the column space of the pre-trained vocabulary matrix via induction-supervision masking. EBM-CoT~\cite{chen2025think1} refines thought trajectories toward lower-energy regions via Langevin dynamics calibration. GainRouter~\cite{zheng2025fast} learns a codebook of discrete strategy priors to condition single-pass decoding on continuous thinking vectors. ThoughtComm~\cite{zheng2025thought} employs a sparsity-regularized autoencoder to transmit latent thoughts between agents. For specialized domains, LatentChem~\cite{ye2026latentchem} aligns projected molecular representations with linguistic spaces, while iCLP~\cite{chen2025iclp} and RB-CoT~\cite{he2026reasoning} generate prompt-sensitive latent plans to guide  multi-step reasoning.

\xhdr{Grounding}
The second branch translates continuous sensory or control signals into structured latent tokens, grounding the frozen backbone in physical reality. In the visual domain, AURORA~\cite{perception2025bigverdi} trains a VQ-VAE to produce discrete visual latent codes that enhance spatial perception, while Monet~\cite{wang2025monet} generates continuous embeddings as intermediate visual thoughts via multi-stage distillation. In embodied robotics, UniVLA~\cite{bu2025univla} derives task-centric latent actions from heterogeneous videos and projects them into a unified action space. LatBot~\cite{li2025latbot} decomposes latent actions into discrete motion and scene tokens; Motus~\cite{bi2025motus} extracts pixel-level delta actions via optical flow; UniLACT~\cite{govind2026unilact} constructs depth-aware latent action representations from RGB-D frames; and VITA~\cite{ma2025unifying} maps visual-action dynamics into unified codebooks.

\xhdr{Augmentation}
The third branch condenses large contextual histories into compressed soft prompts or dynamically augmented cache states, extending effective context beyond standard window limits. DCA~\cite{liu2025deliberation} generates latent embeddings via an offline coprocessor and appends them directly to the KV cache, enriching context without altering the decoding architecture. CLaRa~\cite{he2025clara} encodes lengthy documents into compact memory tokens via a salient compressor trained on synthetic data, enabling end-to-end optimization of the retrieval representation space. DEP~\cite{qiu2025latent} isolates user-specific interaction patterns via a sparse autoencoder and injects them as soft prompts for personalized generation. In multimodal settings, VisMem~\cite{yu2025vismem} and CoMEM~\cite{wu2025towards} construct dedicated visual memory tokens that distill episodic experiences for long-horizon planning, while MemGen~\cite{zhang2025memgen} integrates episodic representations into computation for self-evolving behavior.
 
\xhdr{Summary}
Hybrid representations are particularly suited to complex, multimodal, or domain-specialized settings in which the gap between raw inputs and the backbone's textual manifold creates bottlenecks that neither purely internal nor purely learnable approaches can adequately resolve. By first constructing a task-specific latent representations via targeted learnable acquisition and subsequently deploying it as a structured external conditioning signal, hybrid approaches achieve a dual advantage: semantic fidelity, grounding representations in the raw input modality; and computational efficiency, sparing the backbone from processing raw, high-dimensional inputs at inference time.
\subsection{Computation}
\label{sec:computation}

In recent years, research on latent-level computation has advanced rapidly, reflecting a broader departure from the conventional paradigm in which language models perform inference through fixed-depth, token-by-token generation~\cite{survey2025zhu,feng2025efficient}. Instead, a growing body of work aims to equip models with more flexible, efficient, and scalable computational mechanisms by shifting part of the reasoning process to the latent space.

\begin{figure*}[t]
  \centering
    \includegraphics[width=0.9\linewidth]{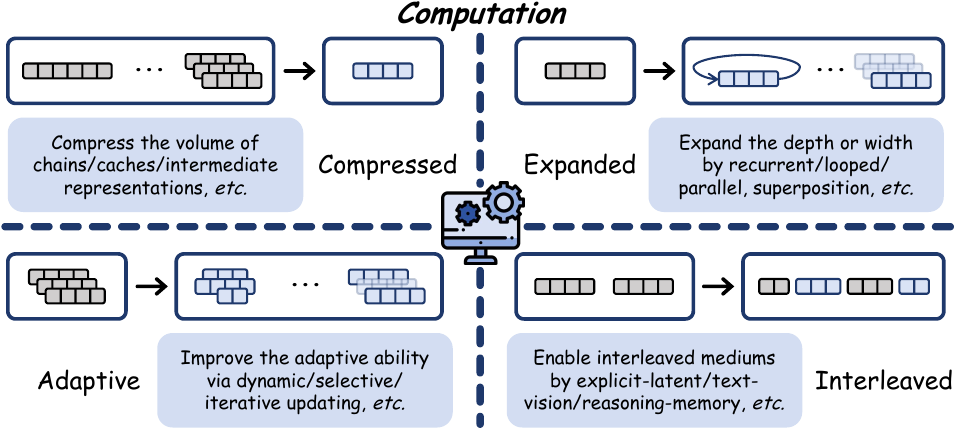}
    \caption{The schematic diagram of \textbf{Computation} mechanism, including four sub-types: \textbf{Compressed} (Section~\ref{sec:compressed}), \textbf{Expanded} (Section~\ref{sec:expanded}), \textbf{Adaptive} (Section~\ref{sec:adaptive}), and \textbf{Interleaved} (Section~\ref{sec:interleaved}).}
    \label{fig:computation}
\end{figure*}

To enable a systematic analysis, a central question to consider is: \textit{what kinds of computational operations are performed in the latent space?} From this perspective, existing methods can be organized along the dimension of \textbf{operation}, yielding four major categories. Specifically, based on their underlying computational operations, as illustrated in Figure~\ref{fig:computation} and Table~\ref{tab:computation}, we classify existing approaches into four representative types:
\vspace{5pt}
\begin{tcolorbox}[
  colback=secblue!5,
  colframe=secblue!50,
  colbacktitle=secblue!50,
  coltitle=black,
  title={\textbf{\textcolor{secblue}{Mechanism:} Computation}},
  boxrule=5pt,
  arc=5pt,
  drop shadow,
  parbox=false,
  before skip=5pt,
  after skip=10pt,
  left=5pt,   
  right=20pt,
]
\begin{itemize}
    \item \textbf{Compressed} (Section~\ref{sec:compressed}): reduces the volume of explicit traces, internal states and cross-modal features, enhancing the efficiency while preserving expressiveness. 
    \item \textbf{Expanded} (Section~\ref{sec:expanded}): increases the effective computation by expanding computation along depth or width by recurrent, looped, parallel, superposition, and structural designs, enabling higher information bandwidth.
    \item \textbf{Adaptive} (Section~\ref{sec:adaptive}): allocates computation adaptively instead of a fixed budget by dynamic depth, width, shortcuts, halting, semantic-unit boundaries adaptation, and control adaptation, balancing capacity and efficiency flexibly.
    \item \textbf{Interleaved} (Section~\ref{sec:interleaved}): interleave heterogeneous generation medias to bridge explicit-latent, language-vision, reasoning-memory, or planning-perception.
\end{itemize}
\end{tcolorbox}

\begin{table*}[!t]
\centering
\caption{Overview of the \textbf{Compressed} (Section~\ref{sec:compressed}), \textbf{Expanded} (Section~\ref{sec:expanded}), \textbf{Adaptive} (Section~\ref{sec:adaptive}), and \textbf{Interleaved} (Section~\ref{sec:interleaved}) computation. We compare the modality, backbone, computational operation, and scenario.}
\setlength{\tabcolsep}{0.9mm}
\resizebox{1\textwidth}{!}{
\begin{tabular}{l|l|ll|llll}
\toprule
\textbf{Date} & \textbf{Method} & \textbf{Paper} & \textbf{Code} & \textbf{Modality} & \textbf{Backbone}  & \textbf{Operation} & \textbf{Scenario}   \\ \midrule
\multicolumn{7}{l}{\textbf{Compressed}} \\\midrule
09/24 & HCoT~\cite{liu2024expediting} & \paperlink{https://arxiv.org/abs/2409.08561} & \qquad - & text & LLaMA2-7B/13B & semantic alignment & efficiency/generalization \\
12/24 & CCoT~\cite{cheng2024compressed} & \paperlink{https://arxiv.org/abs/2412.13171} & \qquad - & text & LLaMA2-7B & variable sequence & latent reasoning/efficiency \\
02/25 & SoftCoT~\cite{xu2025softcot} & \paperlink{https://arxiv.org/abs/2502.12134} & \githublink{https://github.com/xuyige/SoftCoT} & text & LLaMA3.1-8B & semantic projection & efficiency/generalization/zero-shot \\
02/25 & CODI~\cite{shen2025codi} & \paperlink{https://arxiv.org/abs/2502.21074} & \githublink{https://github.com/zhenyi4/codi} & text & GPT2-0.1B/LLaMA3.2-1B & self distillation & latent reasoning/efficiency/generalization  \\
05/25 &  CoLaR~\cite{tan2025think} & \paperlink{https://arxiv.org/abs/2505.16552} & \githublink{https://github.com/xiaomi-research/colar} & text & LLaMA3.2-1B & dynamic compaction & dynamic reasoning/efficiency \\
10/25 & KaVa~\cite{kuzina2025kava} & \paperlink{https://arxiv.org/abs/2510.02312} & \qquad - & text & LLaMA3.2-1B/other 2 & cache distillation & latent reasoning/efficiency \\
10/25 & SALS~\cite{mu2025sals} & \paperlink{https://arxiv.org/abs/2510.24273} & \qquad - & text & LLaMA2-7B/Mistral-7B & cache projection & accelerating/efficiency \\
01/26 & LatentVLA~\cite{xie2026latentvla} & \paperlink{https://arxiv.org/abs/2601.05611} & \qquad - & action & Qwen2.5-VL-3B & vision projection & planning/autonomous driving \\
01/26 & RoT~\cite{wang2026renderofthought} & \paperlink{https://arxiv.org/abs/2601.14750} & \githublink{https://github.com/TencentBAC/RoT} & vision & Qwen3-VL-4B/other 2 & text-to-vision rendering & visual reasoning/accelerating/efficiency \\
02/26 & DeltaKV~\cite{hao2026deltakv} & \paperlink{https://arxiv.org/abs/2602.08005} & \githublink{https://github.com/CURRENTF/Sparse-vLLM} & text & LLaMA3.1-8B/other 3 & semantic residual encoding & complex reasoning/efficiency \\
02/26 & OneLatent~\cite{lv2026onelatent} & \paperlink{https://arxiv.org/abs/2602.13738} & \qquad - & vision & DeepSeek-MoE-3B & text-to-vision rendering & visual reasoning/efficiency \\
02/26 & Future-VLA~\cite{fan2026future} & \paperlink{https://arxiv.org/abs/2602.15882} & \qquad - & action & Qwen3-VL-4B & information densification & embodied planning/efficiency \\
\midrule

\multicolumn{7}{l}{\textbf{Expanded}} \\\midrule
02/25 & Huginn~\cite{jonas2025scaling} & \paperlink{https://arxiv.org/abs/2502.05171} & \githublink{https://github.com/seal-rg/recurrent-pretraining} & text & Huginn-3.5B & recurrent depth & latent reasoning/scaling \\ 
02/25 & Loop~\cite{saunshi2025reasoning} & \paperlink{https://arxiv.org/abs/2502.17416 } & \qquad - & text & decoder-only-1.5B & looped transformer & exploration/multi-step reasoning \\
05/25 & SoftCoT++~\cite{xu2025softcot1} & \paperlink{https://arxiv.org/abs/2505.11484} & \githublink{https://github.com/xuyige/SoftCoT} & text & LLaMA3.1-8B & parallel paths & exploration/zero-shot/scaling \\
05/25 & CoT2~\cite{gozeten2025continuous} & \paperlink{https://arxiv.org/abs/2505.23648} & \githublink{https://github.com/alperengozeten/CoT2} & text & GPT2-0.1B & parallel sampling & exploration/complex reasoning \\
06/25 & PCCoT~\cite{wu25parallel} & \paperlink{https://arxiv.org/abs/2506.18582} & \githublink{https://github.com/whyNLP/PCCoT} & text & GPT2-0.1B/LLaMA3.2-1B & parallel iteration & latent reasoning/efficiency \\
10/25 & Bubbles~\cite{liu2025thoughtbubbles} & \paperlink{https://arxiv.org/abs/2510.00219} & \githublink{https://github.com/stanfordnlp/thoughtbubbles} & text & decoder-only-1.9B/other 3 & parallel width & scaling/zero-shot/unsupervised learning\\
10/25 & ETD~\cite{koishekenov2025encode} & \paperlink{https://arxiv.org/abs/2510.07358} & \qquad - & text &  OLMo-2-1B & recursive iteration & multi-step reasoning/zero-shot \\
10/25 & LatentTTS~\cite{you2025parallel} & \paperlink{https://arxiv.org/abs/2510.07745} & \githublink{https://github.com/YRYangang/LatentTTS} & text &  GPT2-0.1B/LLaMA3.2-1B & parallel sampling & latent reasoning/exploration/scaling \\
10/25 & Latent-SFT~\cite{deng2025latent} & \paperlink{https://arxiv.org/abs/2510.15522} & \githublink{https://github.com/DJC-GO-SOLO/Latent-SFT} & text & LLaMA3.2-1B/other 2 & chain superposition & latent reasoning/efficiency \\
10/25 & Ouro~\cite{zhu2025scaling} & \paperlink{https://arxiv.org/abs/2510.25741} & \qquad - & text &  LoopLM-1.4B/2.6B & looped model pretraining & scaling/efficiency/faithful reasoning \\
12/25 & ColaVLA~\cite{peng2025colavla} & \paperlink{https://arxiv.org/abs/2512.22939} & \githublink{https://github.com/pqh22/ColaVLA} & action & LLaVA-v1.5-7B & hierarchical parallel & planning/autonomous driving \\
01/26 & PLR~\cite{tang2026parallel} & \paperlink{https://arxiv.org/abs/2601.03153} & \qquad - & text & transformer-based & parallel width & sequential recommendation \\
01/26 & Laser~\cite{wang2026forest} & \paperlink{https://arxiv.org/abs/2601.06803} & \githublink{https://github.com/ybb6/laser} & vision & Qwen2.5-VL-7B & feature superposition & visual reasoning/efficiency/generalization \\
02/26 & RD-VLA~\cite{tur2026recurrent} & \paperlink{https://arxiv.org/abs/2602.07845} & \qquad - & action & MiniVLA-0.5B & recurrent-depth & embodied planning/decision \\
02/26 & LoopFormer~\cite{jeddi2026loopformer} & \paperlink{https://arxiv.org/abs/2602.11451} & \githublink{https://github.com/armenjeddi/loopformer} & text & decoder-only & elastic looped transformer & latent reasoning/scaling \\
\midrule

\multicolumn{7}{l}{\textbf{Adaptive}} \\\midrule
05/25 & System-1.5~\cite{wang2025system} & \paperlink{https://arxiv.org/abs/2505.18962} & \qquad - & text &  LLaMA3.2-1B & dynamic shortcuts & cognitive reasoning/efficiency \\
09/25 & FR-Ponder~\cite{he2025learning} & \paperlink{https://arxiv.org/abs/2509.24238} & \qquad - & text & LLaMA3-8B/other 4 &  instance-adaptive steering & dynamic reasoning/efficiency/generalization \\
11/25 & TaH~\cite{fu2025tah} & \paperlink{https://arxiv.org/abs/2511.08577} & \githublink{https://github.com/thu-nics/TaH} & text & Qwen3-0.6B/1.7B & selective iterations & dynamic reasoning/complex reasoning \\
11/25 & LWS~\cite{ning2025learning} & \paperlink{https://arxiv.org/abs/2511.21581} & \githublink{https://github.com/apning/adaptive-latent-reasoning} & text & LLaMA3.2-1B & adaptive halting & dynamic reasoning/efficiency \\
12/25 & DLCM~\cite{qu2025dynamic} & \paperlink{https://arxiv.org/abs/2512.24617} & \qquad - & text & DLCM-2.3B & concept boundaries & concept-level/zero-shot/scaling \\
01/26 & PRE~\cite{liu2026layerorder} & \paperlink{https://arxiv.org/abs/2601.03542} & \githublink{https://github.com/laquabe/Layer-Order-Inversion} & text & GPT-J-6B/LLaMA3.2-8B & selective extraction & latent reasoning/multi-hop reasoning \\
01/26 & I2B-LPO~\cite{deng2026iiblpo} & \paperlink{https://arxiv.org/abs/2601.05870} & \githublink{https://github.com/denghuilin-cyber/IIB-LPO} & text & Qwen3-14B/Qwen2.5-37B  & dynamic branching & exploration/policy optimization \\
01/26 & RISER~\cite{ye2026riser} & \paperlink{https://arxiv.org/abs/2601.09269} & \githublink{https://github.com/gooogleshanghai/RISER-Orchestrating-Latent-Reasoning-Skills-for-Adaptive-Activation-Steering} & text & Qwen2.5-7B/other 3 & adaptive steering & zero-shot/generalization \\
01/26 & PLaT~\cite{wang2026latent} & \paperlink{https://arxiv.org/abs/2601.21358} & \githublink{https://github.com/yunsaijc/PLaT} & text & GPT2-0.1B & dynamic termination & exploration/efficiency/generalization \\
01/26 & Dreamer~\cite{knupp2026depthrecurrent} & \paperlink{https://arxiv.org/abs/2601.21582} & \qquad - & text & Dreamer-1B/2B & depth-recurrent attention & latent reasoning/scaling/efficiency \\
02/26 & AL-CoT~\cite{zeng2025pretraining} & \paperlink{https://arxiv.org/abs/2602.08220} & \qquad - & text & PonderLM2-0.5B/1.4B & token-level adaption & latent reasoning/dynamic reasoning \\
02/26 & SpiralFormer~\cite{yu2026spiralformer} & \paperlink{https://arxiv.org/abs/2602.11698} & \qquad - & text & Pythia-1.4B/other 3 & looped transformers & efficiency/scaling \\
03/26 & PonderLM-3~\cite{li2026ponderlm} & \paperlink{https://arxiv.org/abs/2603.02023} & \qquad - & text & PonderLM2-0.5B & token-wise adaption & latent reasoning/efficiency \\
03/26 & AdaPonderLM~\cite{song2026adaponderlm} & \paperlink{https://arxiv.org/abs/2603.01914} & \qquad - & text & Pythia-1.4B/other 4 & token-wise adaptive depth & latent reasoning/efficiency \\
\midrule

\multicolumn{7}{l}{\textbf{Interleaved}} \\\midrule
12/24 & AURORA~\cite{perception2025bigverdi} & \paperlink{https://arxiv.org/abs/2412.03548} & \githublink{https://github.com/mahtabbigverdi/Aurora-perception} & vision & LLaVA1.5-13B & text/perception & visual perception \\
02/25 & Assorted~\cite{su2025token} & \paperlink{https://arxiv.org/abs/2502.03275} & \qquad - & text & LLaMA3.2-1B/3B/8B & text/latent & latent reasoning/efficiency \\
06/25 & Mirage~\cite{yang2025machine} & \paperlink{https://arxiv.org/abs/2506.17218} & \githublink{https://github.com/UMass-Embodied-AGI/Mirage} & vision & Qwen2.5-VL-7B & text/vision latent & visual imagination/generalization \\
08/25 & LCR-SER~\cite{shi2025bridging} & \paperlink{https://arxiv.org/abs/2508.04152} & \qquad - & text & transformer-based & history/reasoning & working memory/search\&recommendation \\
09/25 & LVR~\cite{li2025latent} & \paperlink{https://arxiv.org/abs/2509.24251} & \githublink{https://github.com/VincentLeebang/lvr} & vision & Qwen2.5-VL-3B/7B & text/vision latent & visual understanding/zero-shot \\
09/25 & MemGen~\cite{zhang2025memgen} & \paperlink{https://arxiv.org/abs/2509.24704} & \githublink{https://github.com/bingreeky/MemGen} & text & SmolLM3-3B/Qwen3-8B & reasoning/memory &  experimental memory/generalization \\
10/25 & SwiReasoning~\cite{shi2025swireasoning} & \paperlink{https://arxiv.org/abs/2510.05069} & \githublink{https://github.com/sdc17/SwiReasoning} & text & Qwen3-8B/other 3 & explicit/latent reasoning & latent reasoning/efficiency\\
10/25 & IVT-LR~\cite{chen2025reasoning} & \paperlink{https://arxiv.org/abs/2510.12603} & \githublink{https://github.com/FYYDCC/IVT-LR} & vision & Qwen2-VL-7B/Chameleon-7B & text latent/vision latent & visual reasoning/efficiency \\
10/25 & 3DThinker~\cite{chen2025think} & \paperlink{https://arxiv.org/abs/2510.18632} & \githublink{https://github.com/zhangquanchen/3DThinker} & vision & Qwen2.5-VL-7B/other 7 & text/3D latent & spatial understanding/visual reasoning \\
10/25 & LS~\cite{zhang2025latentsketchpad} & \paperlink{https://arxiv.org/abs/2510.24514} & \githublink{https://github.com/hwanyu112/Latent-Sketchpad} & vision & Gemma3-12B/other 2 & text/vision latent & visual imagination/visual understanding \\
11/25 & SpiralThinker~\cite{piao2025spiralthinker} & \paperlink{https://arxiv.org/abs/2511.08983} & \qquad - & text & LLaMA3.2-7B & text/latent & latent reasoning/complex reasoning \\
11/25 & VisMem~\cite{yu2025vismem} & \paperlink{https://arxiv.org/abs/2511.11007} & \githublink{https://github.com/YU-deep/VisMem} & vision & Qwen2.5-VL-7B/other 8 & short/long vision memory & visual understanding/visual reasoning \\
11/25 & CLaRa~\cite{he2025clara} & \paperlink{https://arxiv.org/abs/2511.18659} & \githublink{https://github.com/apple/ml-clara} & text & Mistral-7B/Phi4-Mini-3.8B & retrieval/generation & retrieval/efficiency \\
11/25 & Monet~\cite{wang2025monet} & \paperlink{https://arxiv.org/abs/2511.21395} & \githublink{https://github.com/NOVAglow646/} & vision & Qwen2.5-VL-7B & text/vision latent & visual reasoning/generalization \\
12/25 & LiteReason~\cite{gurung2025lightweight} & \paperlink{https://arxiv.org/abs/2512.02240} & \qquad - & text & Qwen2.5-7B & reasoning/latent & latent reasoning/efficiency \\
12/25 & ILVR~\cite{dong2025interleaved} & \paperlink{https://arxiv.org/abs/2512.05665} & \githublink{https://github.com/XD111ds/ILVR} & vision & Qwen2.5-VL-7B/Qwen3-VL-8B & text/vision latent & visual understanding/efficiency \\
12/25 & DMLR~\cite{liu2025reason} & \paperlink{https://arxiv.org/abs/2512.12623} & \githublink{https://github.com/eric-ai-lab/DMLR} & vision & Qwen2.5-VL-7B/other 5 & text/perception & visual understanding/grounding \\
12/25 & SkiLa~\cite{tong2025sketchinlatents} & \paperlink{https://arxiv.org/abs/2512.16584} & \githublink{https://github.com/TungChintao/SkiLa} & vision & Qwen2.5-VL-7B & text/vision latent & visual imagination/visual understanding  \\
01/26 & FlashMem~\cite{hou2026flashmem} & \paperlink{https://arxiv.org/abs/2601.05505} & \qquad - & text & Qwen2.5-1.5B/other 3 & reasoning/memory & working memory/efficiency \\
02/26 & L2-VMAS~\cite{yu2026dual} & \paperlink{https://arxiv.org/abs/2602.00471} & \githublink{https://github.com/YU-deep/L2-VMAS} & vision & Qwen3-VL-8B/other 8 & perception/thinking & multi-agent collaboration/long-horizon \\
02/26 & MM-CoT~\cite{shao2026learning} & \paperlink{https://arxiv.org/abs/2602.00574} & \qquad - & vision & Qwen2.5-VL-7B & text/vision latent & visual imagination/visual understanding \\
02/26 & LatentMem~\cite{fu2026latentmem} & \paperlink{https://arxiv.org/abs/2602.03036} & \githublink{https://github.com/KANABOON1/LatentMem} & text & Qwen3-4B/LLaMA3.1-8B & text/memory & multi-agent collaboration \\
02/26 & LT-Tuning~\cite{liu2026latent} & \paperlink{https://arxiv.org/abs/2602.10229} & \githublink{https://github.com/NeosKnight233/Latent-Thoughts-Tuning} & text &  LLaMA3.2-1B/3B/8B & text/latent & latent reasoning/dynamic reasoning \\
02/26 & ThinkRouter~\cite{xu2026thinkrouter} & \paperlink{https://arxiv.org/abs/2602.11683} & \qquad - & text & Qwen3-8B/other 3 & explicit/latent reasoning  & dynamic reasoning/efficiency \\
 \bottomrule
\end{tabular}}
\label{tab:computation}
\end{table*}

\subsubsection{Compressed}
\label{sec:compressed}
This paradigm subsumes approaches that project an explicit 
trajectory $\mathbf{r} \in \mathcal{V}$ onto a semantically dense 
latent representation $\mathbf{z}$. 
A compression operator $\Phi(\cdot)$ acts upon the intermediate 
hidden states $\mathbf{h}$ to yield:
\begin{equation}
    \mathbf{z} = \Phi(\mathbf{h}), \quad |\mathbf{z}| \ll |\mathbf{h}|,
\end{equation}
where $\Phi(\cdot)$ may be realized as both a functional 
component $\Phi^{comp}(\cdot)$ or the backbone $\Phi^{\mathrm{back}}$ itself, reducing autoregressive overhead 
while preserving the semantic fidelity requisite for faithful downstream decoding.

Compressed computation aims to reduce the cost of explicit intermediate computation by mapping verbose reasoning trajectories or high-dimensional hidden states into compact latent representations, while preserving the semantic information necessary for accurate downstream decoding. Rather than eliminating reasoning altogether, this paradigm seeks to retain inferential content in a more information-dense form, thereby improving efficiency across textual, internal, and cross-modal reasoning processes in three forms: \textbf{Traces Compression}, \textbf{States Compression}, and \textbf{Features Compression}.

\xhdr{Traces Compression} Recent work on this direction can be organized around a shared goal: preserving inferential fidelity while reducing explicit trace lengths. HCoT~\cite{liu2024expediting} and SoftCoT~\cite{xu2025softcot} emphasize semantic alignment across abstraction levels, with HCoT~\cite{liu2024expediting} using hierarchical compression to skip redundant intermediate tokens and SoftCoT~\cite{xu2025softcot} projecting reasoning into a continuous soft-token space that decouples reasoning depth from discrete trace length, thereby supporting zero-shot transfer without task-specific retraining. In parallel, CCoT~\cite{cheng2024compressed} models compression as variable-length latent allocation, assigning shorter representations to simpler inferences and richer ones to more complex deductions, while CODI~\cite{shen2025codi} and CoLaR~\cite{tan2025think} focus on adaptive compaction through self-distillation and dynamic token-level control, respectively.

\xhdr{States Compression} A complementary and potentially deeper form of compression targets not the decoded token sequence but the internal cache, whose linear growth with sequence length has become a major bottleneck for long-context and reasoning-intensive inference. Recent work explores three main approaches. KaVa~\cite{kuzina2025kava} formulates KV-cache compression as a knowledge distillation problem, training a student model to match the output distribution of a teacher operating over the full cache. SALS~\cite{mu2025sals} instead adopts a lightweight, training-free strategy by projecting the cache into a low-rank principal subspace, arguing that the dominant directions are sufficient to preserve attention patterns with bounded error. DeltaKV~\cite{hao2026deltakv} exploits the high correlation between KV-caches across successive reasoning steps, storing semantically compressed residuals rather than full cache states, and reports particularly strong gains on multi-step reasoning tasks where inter-step redundancy is greatest.

\xhdr{Features Compression} Recent work on latent-space compression has moved beyond text to multimodal and embodied settings, where the main bottleneck is the high-dimensional visual and action representations used in perception–action loops such as autonomous driving, robot manipulation, and embodied navigation. In visual reasoning, RoT~\cite{wang2026renderofthought} renders intermediate reasoning states as low-resolution image patches rather than token sequences, enabling subsequent steps to operate over compressed visual structure, OneLatent~\cite{lv2026onelatent} learns a single latent visual token that distills the image context needed for downstream reasoning, achieving competitive performance with a drastically reduced visual token budget. In embodied control, LatentVLA~\cite{xie2026latentvla} projects visual inputs into compact action latents for real-time planning, showing that low-dimensional representations can retain the information necessary for closed-loop decision making, while Future-VLA~\cite{fan2026future} extends this idea by learning future-conditioned latents that incorporate anticipated states without increasing dimensionality.

\xhdr{Summary} Overall, compressed reasoning can be understood as a unifying paradigm for trading explicit length for latent density. Across explicit traces, internal states, and cross-modal representations, existing methods share the same central objective: to preserve reasoning fidelity while reducing the computational and memory overhead associated with verbose intermediate representations. This suggests that effective reasoning need not always remain fully tokenized or fully materialized, and that semantically compact representations may provide a scalable foundation for efficient inference in increasingly complex reasoning systems.

\subsubsection{Expanded}
\label{sec:expanded}
In this category, it augments the effective computational capacity of the model by extending latent computation along 
the depth or width dimensions. Formally, the model iterates over a 
step-$T$, width-$K$:
\begin{equation}
    \mathbf{h}_{t+1}^{(k)} = \Phi\!\left(
    \bigl\{\mathbf{h}_{t}^{(k)}\bigr\}_{k=1}^{K}
    \right), \quad t = 1, \ldots, T,\;\; k = 1, \ldots, K,
\end{equation}
where $\mathbf{h}_{t}^{(k)}$ denotes the $k$-th latent trajectory at step $t$, and all $K$ paths share a common initialization 
$\mathbf{h}^{(0)}$.

Expanded methods increase the effective computational capacity by enlarging the latent process along different dimensions. Rather than relying on a single fixed forward pass, they enable the model to trade extra latent computation for stronger ability, improved faithfulness, and better adaptability across tasks, which could be classified into three types of  expansion: \textbf{Depth Expansion}, \textbf{Width Expansion}, and \textbf{Structural Expansion}.

\xhdr{Depth Expansion} Depth-expanding approaches increase effective compute by reusing the same or structurally similar layers across multiple recurrent passes, allowing the model to iteratively refine a latent representation before producing an output. In recurrent pretraining, Huginn~\cite{jonas2025scaling} introduces a lightweight model trained from scratch with recurrent depth, where a fixed transformer block is applied for a variable number of thinking steps, thereby decoupling parameter count from inference-time compute and enabling test-time scaling without adding parameters. RD-VLA~\cite{tur2026recurrent} extends this paradigm to embodied models and shows that recurrent latent refinement improves long-horizon manipulation planning. In looped transformers, Loop~\cite{saunshi2025reasoning} demonstrates that repeatedly applying the full decoder stack can induce genuinely multi-step reasoning rather than simply approximating a deeper fixed network , while LoopFormer~\cite{jeddi2026loopformer} adds input-adaptive looping to allocate depth dynamically according to task complexity . From a pretraining perspective, Ouro~\cite{zhu2025scaling} shows favorable scaling behavior along the looping dimension, with gains in faithfulness and efficiency . Relatedly, ETD~\cite{koishekenov2025encode} proposes an encode-think-decode framework in which a latent thought state is iteratively refined before decoding, yielding strong zero-shot generalization on arithmetic and commonsense tasks without explicit chain-of-thought supervision.

\xhdr{Width Expansion} These methods contrast with other approaches by allocating compute across multiple parallel hypotheses or latent trajectories, thereby prioritizing broad exploration over sequential refinement. This paradigm appears in several recent frameworks, SoftCoT++~\cite{xu2025softcot1} instantiates multiple parallel reasoning paths in continuous embedding space and aggregates them before decoding, achieving zero-shot performance gains that scale monotonically with the number of paths without requiring path-specific supervision. A similar principle underlies LatentTTS~\cite{you2025parallel}, which performs concurrent latent tree search and prunes partial hypotheses with a learned value function. PCCoT~\cite{wu25parallel} further extends this idea by enabling multiple latent chains to run in parallel and exchange information at synchronization points, improving efficiency while preserving solution diversity; likewise, CoT2~\cite{gozeten2025continuous} shows that parallel sampling in continuous thought space substantially benefits even GPT2-scale models on compositional reasoning tasks. In specialized settings, Bubbles~\cite{liu2025thoughtbubbles} formulates width expansion as parallel ``bubbles" whose pooled representations support zero-shot and unsupervised learning at scale, while PLR~\cite{tang2026parallel} applies parallel latent reasoning to sequential recommendation, where diverse latent representations better capture user preference uncertainty than a single deterministic pass.

\xhdr{Structural Expansion} A smaller but conceptually distinct line of work departs from simply increasing the depth or width of the computation graph and instead introduces new topological primitives for composing latent information across positions, modalities, or levels of control. Latent-SFT~\cite{deng2025latent} supervises models on superposed latent chains—continuous compressions of multi-step reasoning trajectories—rather than their token-level realizations, thereby reducing decoding cost while preserving reasoning performance. Extending a similar superposition principle to visual reasoning, Laser~\cite{wang2026forest} fuses multi-scale visual features into a shared latent representation instead of processing each resolution independently. At the system level, ColaVLA~\cite{peng2025colavla} decomposes VLA modeling into parallel but interactive high-level semantic reasoning and low-level motor planning streams, linked through cross-attention across temporal scales, which better aligns linguistic abstraction with continuous control.

\xhdr{Summary} In this part, approaches enrich latent reasoning by scaling computation along three complementary dimensions: depth, width, and structure. Depth expansion emphasizes iterative refinement through recurrent or looped computation, width expansion promotes parallel exploration over multiple latent hypotheses, and structural expansion introduces richer topologies for composing information across trajectories, positions, or modalities. Together, these methods share a common goal of improving reasoning performance by increasing effective calculation in latent space, while avoiding a proportional increase in model parameters.

\subsubsection{Adaptive}
\label{sec:adaptive}
In this part, methods generalize the Expanded framework by conditioning both 
recurrence depth $T$ and trajectory width $K$ on the input complexity 
of $\mathbf{x}$, rather than prescribing them as fixed hyperparameters. 
A learned halting functional $\mathcal{T}$ governs instance-specific 
termination, yielding a variable-horizon computation:
\begin{equation}
    \mathbf{h}_{t+1}^{(k)} = \Phi\!\left(
    \bigl\{\mathbf{h}_{t}^{(k)}\bigr\}_{k=1}^{K}
    \right),
\end{equation}
where $t\in [1,T]$ and $k\in[1,K]$ may be determined at the granularity of 
tokens boundaries, or other adaptive strategies based on the current hidden states by a halting function $\mathcal{T}(\cdot)$:
\begin{equation}
    (T,\, K) = \mathcal{T}(\mathbf{h}_{t};\,\mathbf{x}),
    \qquad
\end{equation}
where the halting function could be a explicit component, or be internalized the model.
This input-conditioned 
allocation concentrates reasoning depth where warranted by task difficulty, 
realizing a principled trade-off between computational frugality and 
expressive capacity.

Adaptive computation allows the model to allocate computational resources dynamically according to the complexity of the input, rather than relying on a fixed recurrence depth or trajectory width. In this setting, the amount, structure, and locus of computation become input-dependent, enabling the model to balance efficiency and capacity more flexibly. Existing methods can be broadly understood from three perspectives: \textbf{Depth/Width Adaptation}, \textbf{Semantic Adaptation}, and \textbf{Control Adaptation}.

\xhdr{Depth/Width Adaptation} A natural axis for adaptive computation is depth, that is, the number of steps applied. Building on this principle, TaH~\cite{fu2025tah} proposes a think-and-halt mechanism that allows a shared-weight iterative model to exit early on easy tokens while allocating additional refinement steps to harder ones. LWS~\cite{ning2025learning} similarly treats halting as a learned policy, jointly optimizing an auxiliary stopping variable with the language modeling objective. Moving from depth-wise iteration to latent-chain guidance, PLaT~\cite{wang2026latent} learns when to terminate latent token generation before decoding, explicitly trading off depth against efficiency. At the architectural level, Dreamer~\cite{knupp2026depthrecurrent} and SpiralFormer~\cite{yu2026spiralformer} replace fixed-depth stacks with weight-tied recurrent transformers, demonstrating that depth adaptivity can be achieved not only through learned halting policies but also through recurrent designs.

While depth adaptation modulates the length of the computational chain, width adaptation modulates its branching structure, \textit{i.e.}, the parallel hypotheses or policy trajectories explored at each step. 
I2B-LPO~\cite{deng2026iiblpo} learns to expand its branching factor in regions of high uncertainty and to collapse branches where confidence is sufficient, demonstrating that width adaptation is particularly impactful in long-horizon reasoning tasks where premature commitment to a single reasoning path leads to systematic failure.

\xhdr{Semantic Adaptation} This paradigm allocates computation at the finest granularity by assigning different budgets to individual tokens or semantic units within a single forward pass. For example, AL-CoT~\cite{zeng2025pretraining} applies token-level adaptation within the semantic, assigning more refinement steps to difficult tokens. A similar idea is developed in PonderLM-3~\cite{li2026ponderlm} and AdaPonderLM~\cite{song2026adaponderlm}, which endow individual tokens with learned depth or halting decisions, enabling heterogeneous per-token computation and stronger semantic consistency.
In contrast, DLCM~\cite{qu2025dynamic} shifts adaptation from tokens to semantically coherent concepts that may span multiple tokens.

\xhdr{Control Adaptation} This part refers to methods that regulate computation through control signals, \textit{e.g.}, learned activation vectors, routing policies, or extraction mechanisms. System-1.5~\cite{wang2025system} introduces cognitive shortcuts that allow to bypass intermediate transformer blocks according to estimated input difficulty. FR-Ponder~\cite{he2025learning} extends this idea through instance-adaptive activation steering, routing each input through a dynamically selected subset of representational subspaces based on early-layer features. RISER~\cite{ye2026riser} similarly treats adaptation as the composition of latent reasoning skills encoded as steering directions in activation space, enabling zero-shot generalization to unseen tasks without gradient-based updating. Complementarily, PRE~\cite{liu2026layerorder} shows that selective extraction of intermediate-layer representations as reversed signals, outperforming reliance on final-layer outputs.

\xhdr{Summary} In summary, adaptive methods generalize fixed-budget recurrent reasoning into an input-conditioned computation paradigm. Across depth/width adaptation, semantic adaptation, and control adaptation, the common goal is to allocate computation more selectively: spending more resources on difficult inputs, uncertain reasoning branches, or critical semantic units, while preserving efficiency on simpler cases. Taken together, these approaches suggest that the future of adaptive methods lie not merely in increasing computation, but in learning where, when, and how to use it most effectively.

\subsubsection{Interleaved}
\label{sec:interleaved}
This paradigm constructs a heterogeneous generation sequence by 
alternating discrete token embeddings in $\mathcal{V}$ with 
continuous latent in $\mathcal{H}$, yielding interleaved generation:
\begin{equation}
    \mathbf{r} = \bigl[r_1,\, z_1,\, r_2,\, z_2,\, \ldots,\, 
    r_M,\, z_N\bigr], 
    \quad r_i \in \mathcal{V},\;\; z_j \in \mathcal{H},
\end{equation}
where the $\mathbf{r}$ generation trajectory, achieving 
a synergistic coupling of explicit symbolic reasoning and implicit 
neural computation.

Interleaved computation views reasoning as a heterogeneous sequential process in which discrete tokens and continuous latents are alternated within a unified trajectory. Compared with purely token-based or purely latent reasoning, this paradigm provides a more flexible interface for allocating computation across explicit symbolic steps and implicit neural operations. Existing work can be broadly grouped into three categories: \textbf{Explicit-latent Interleaving}, \textbf{Modality Interleaving}, and \textbf{Task Interleaving}.

\xhdr{Explicit-latent Interleaving} Hybrid generation interleaves natural-language tokens with latent internal states within a single trajectory, motivated by the observation that many intermediate steps need not be fully verbalised and can therefore be executed more efficiently without degrading performance. Early empirical support comes from Assorted~\cite{su2025token}, which shows that replacing selected verbal chain-of-thought steps with latent activations preserves downstream accuracy while substantially reducing token usage. Subsequent work extends this idea in two main directions. One line studies fixed or curriculum-based interleaving schemes: SpiralThinker~\cite{piao2025spiralthinker} progressively internalises explicit reasoning steps through a spiral curriculum, and LiteReason~\cite{gurung2025lightweight} demonstrates that even a lightweight model can learn a form of mixed explicit–latent reasoning. A second line makes the explicit–latent boundary itself adaptive: SwiReasoning~\cite{shi2025swireasoning} learns a step-level switching policy via reinforcement learning, LT-Tuning~\cite{liu2026latent} incorporates latent-thought positions into instruction tuning, and ThinkRouter~\cite{xu2026thinkrouter} routes queries to different reasoning depths at the system level to balance efficiency and accuracy.

\xhdr{Modality Interleaving} When generation over non-textual inputs such as visual imagination, or perceptual feature maps, latent representations are not merely an efficiency mechanism but a representational requirement, since the underlying information cannot be fully captured in discrete text. Modality-interleaving methods therefore integrate continuous perceptual latents with linguistic tokens, allowing models to perform intermediate reasoning in the latent domain. Early work such as AURORA~\cite{perception2025bigverdi} established this paradigm by inserting explicit perceptual reasoning steps. Subsequent studies, including Mirage~\cite{yang2025machine}, LVR~\cite{li2025latent}, VisMem~\cite{yu2025vismem}, and Monet~\cite{wang2025monet}, showed that directly interleaving vision latents into architectures improves visual tasks without relying on textual scene descriptions.

A related line of work extends interleaving to multiple latent types: IVT-LR~\cite{chen2025reasoning} couples text and vision activations through cross-attention, while SkiLa~\cite{tong2025sketchinlatents}, LS~\cite{zhang2025latentsketchpad}, and MM-CoT~\cite{shao2026learning} use visual latents as an internal sketchpad that supports subsequent language reasoning. Beyond two-dimensional perception, 3DThinker~\cite{chen2025think} incorporates spatial latents for geometric and spatial reasoning, and DMLR~\cite{liu2025reason} together with ILVR~\cite{dong2025interleaved} shows that sparse latent interleaving can preserve strong grounding performance while reducing computational cost.

\xhdr{Task Interleaving} A third  family views interleaving not as a mechanism for mixing token types or modalities, but as a means of coordinating heterogeneous functional modules such as memory stores, retrieval components, generative heads, and agent sub-processes. Its defining feature is that the interleaved latent stream carries task-specific structured signals, \textit{e.g.},  retrieved evidence, episodic memory traces, or inter-agent communications, rather than general-purpose reasoning states. In single-model settings, this idea primarily appears as memory–reasoning integration: LCR-SER explicitly interleaves a compressed history buffer with ongoing inference for sequential recommendation~\cite{shi2025bridging}; MemGen~\cite{zhang2025memgen}, VisMem~\cite{yu2025vismem} and FlashMem~\cite{hou2026flashmem} equip models with persistent working memory through, respectively, generated memory tokens; and CLaRa~\cite{he2025clara} interleaves retrieval and generation in a jointly trained latent lookup framework, rather than treating retrieval as an external pipeline. In multi-agent settings, interleaving further serves as a coordination protocol across asynchronously operating agents: LatentMem~\cite{fu2026latentmem} introduces a shared latent memory that supports implicit communication during collaborative problem solving, while L2-VMAS~\cite{yu2026dual} alternates perceptual and planning streams across agents to maintain a coherent joint world model in long-horizon embodied tasks.

\xhdr{Summary} Interleaved computation generalises the generation process beyond a homogeneous token stream by allowing models to alternate between symbolic outputs and latent computations. This design improves efficiency when some intermediate steps need not be verbalised, expands representational capacity when generation must operate over non-textual modalities, and enables tighter coordination when multiple functional modules or agents interact within a shared trajectory. Taken together, these studies suggest that interleaving is not merely a decoding strategy, but a general principle for organising hybrid computation systems that combine explicit interpretability with implicit computational power.
\subsection{Optimization}
\label{sec:optimization}

Latent space optimization generally happens at three stages: pre-training, post-training, and inference. The three stages differ mainly in what is optimized: during pre-training and post-training, the optimized variable is typically model parameters $\theta \in \mathcal{W}$ (or a subset of them);
during inference, the model parameters are mostly fixed (sometimes also be trained), and the optimized variable is instead an inference-time state, such as a latent representation $\mathbf{z}\in\mathcal{H}$ or a generation trajectory $\mathbf{r}\in\mathcal{V}$.
For each stage, we categorize methods based on two aspects: \textbf{supervision}, that specifies what provides the learning signal; \textbf{objective}, that captures the purpose behind each loss component for latent optimization.
Based on underlying computational operations, we categorize existing methods into three representative types:
\begin{tcolorbox}[
    colback=secblue!5,
    colframe=secblue!50,
    colbacktitle=secblue!50,
    coltitle=black,
    title={\textbf{\textcolor{secblue}{Mechanism:} Optimization}},
    boxrule=5pt,
    arc=5pt,
    drop shadow,
    parbox=false,
    before skip=5pt,
    after skip=10pt,
    left=5pt,   
    right=20pt,
  ]
  \begin{itemize}
      \item \textbf{Pre-training} (Section~\ref{sec:pre_training}): starts with a randomly initialized model and trains it from the scratch, enabling native latent-level abilities.
      \item \textbf{Post-training} (Section~\ref{sec:post_training}): enhances the ability of pre-trained models, with diverse supervision signals and objectives, learning the latent space.
      \item \textbf{Inference} (Section~\ref{sec:inference}): focuses on inference manipulation of latent states, allowing dynamic adjustment.
  \end{itemize}
\end{tcolorbox}

\subsubsection{Pre-training}
\label{sec:pre_training}
During the pre-training stage, $\mathbf{z}$ is trained jointly with the base model from scratch, the objective is to learn model parameters from large-scale pre-training data $\mathcal{D}$.
Formally, the optimization variable is $\theta \in \Phi$, and the objective can be written as:
\begin{equation}
    \theta^{\star}
    =
    \arg\min_{\theta \in \Phi}
    \mathbb{E}_{(\mathbf{x},\mathbf{y})\sim\mathcal{D}}
    \left[
        \mathcal{L}(\mathbf{x}, \mathbf{y}, \mathbf{z}; \Phi_\theta)
    \right],
\end{equation}
where $\theta \in \Phi \subseteq\mathcal{W}$ denotes the set of trainable parameters in pre-training, and $\mathcal{L}$ trains the whole system end-to-end.

Optimization at this stage mostly relies on simple, scalable supervision already well-established for explicit-space training, rather than more sophisticated supervision that requires human annotation or elaborate processing. To stabilize training, optional auxiliary losses are designed to regulate the latent space. In total, these methods fall into three types: \textbf{Autoregressive Supervision}, \textbf{Auxiliary Supervision}, and \textbf{Reinforcement}.

\xhdr{Autoregressive Supervision Pre-Training} This category focuses on internalizing reasoning as a natural byproduct of next-token prediction or recurrent looping, without explicitly matching intermediate representations. Implementing Jacobi-iteration-based parallel training, PonderLM-2~\cite{zeng2025pretraining} efficiently computes recursive latent thoughts without strict causal blocking. Exploring recurrent depth scaling through looped transformer layers, Looped Trans.~\cite{saunshi2025reasoning} natively simulates complex reasoning via continuous latent updates. Scaling looped language models up to 2.6 billion parameters, Ouro~\cite{zhu2025scaling} utilizes an entropy-regularized objective for consistent latent reasoning. To address scaling efficiency, PHD-Trans.~\cite{wu2025efficient} optimizes predictions via dynamic exploration of continuous states. To maintain stability in dynamic scenarios, AL-CoT~\cite{zeng2026pretraining} employs continuous state prediction objectives.

\xhdr{Auxiliary Supervision Pre-Training} This direction explicitly guides the formation of the latent space geometry through auxiliary semantics, contrastive learning, or reconstruction objectives. Optimizing continuous semantic concepts via cross-entropy and reconstruction losses, CoCoMix~\cite{jihoon2025llm} implicitly guides multi-task textual predictions. Incorporating InfoNCE and cross-entropy, LARES~\cite{liu2025lares} aligns latent representations for sequential recommendation. Introducing latent action pretraining via VQ-VAE, LAPA~\cite{ye2024lapa} learns discretized latent actions directly from unannotated videos. Aligning visual latent spaces from videos with proprioceptive action spaces, CLAP~\cite{zhang2026clap} utilizes contrastive pretraining for robust embodied manipulation. Bypassing full visual dynamic reconstruction, JALA~\cite{jala2026joint} derives predictive action embeddings jointly aligned with inverse dynamics and sparse physical actions. Focusing on mean squared error and task losses, CARE~\cite{shi2026care} scales multi-task prediction for embodied manipulation. Utilizing cross-entropy and mean squared error, ConceptLM~\cite{liu2026next} learns latent representations natively for efficient multi-task prediction.

\xhdr{Reinforcement Pre-Training} This category integrates reward signals directly into the foundational pre-training phase to actively shape latent thought trajectories. Integrating reinforcement signals, LoopRPT~\cite{tang2026looprpt} reframes next-token prediction as a reasoning task via noisy latent rollouts, optimizing intermediate representations to compress effective reasoning into fewer iterations from the very beginning of the lifecycle.

\xhdr{Summary} Methods at this stage embed latent reasoning capacity directly into model parameters during large-scale training, foregoing the need for human annotation or elaborate supervision. The dominant approach is autoregressive prediction over continuous states, where looped or recurrent architectures naturally develop latent reasoning through standard next-token objectives. Auxiliary losses serve a regularizing role, shaping the geometry of the latent space without disrupting scalability. A nascent thread introduces reinforcement signals at pre-training time, aiming to instill efficient reasoning trajectories before any fine-tuning occurs.

\begin{table*}[!t]
\centering
\caption{Overview of the \textbf{Pre-training} (Section~\ref{sec:pre_training}), \textbf{Post-training} (Section~\ref{sec:post_training}), and \textbf{Inference} (Section~\ref{sec:inference}) optimization. We compare the modality, backbone, computational operation, and scenario. 
Here, Recon., CE, InfoNCE, KL, MSE, and CL denote reconstruction, cross-entropy, information noise-contrastive estimation, Kullback–Leibler divergence, mean squared error, and contrastive learning, respectively.}
\label{tab:optimization}
\setlength{\tabcolsep}{0.9mm}
\renewcommand{\arraystretch}{0.9}
\resizebox{1\textwidth}{!}{
\begin{tabular}{l|l|ll|lllll}
\toprule
\textbf{Date} & \textbf{Method} & \textbf{Paper} & \textbf{Code} & \textbf{Modality} & \textbf{Backbone} & \textbf{Supervision} & \textbf{Objective} & \textbf{Scenario} \\ \midrule
\multicolumn{8}{l}{\textbf{Pre-training}} \\\midrule
10/24 & LAPA~\cite{ye2024lapa} & \paperlink{https://arxiv.org/abs/2410.11758} & \qquad - & action & LWM-7B & Recon. & action quantization/prediction & embodied manipulation/transferring \\
02/25 & CoCoMix~\cite{jihoon2025llm} & \paperlink{https://arxiv.org/abs/2502.08524} & \githublink{https://github.com/facebookresearch/RAM/tree/main/projects/cocomix} & text & GPT2-0.1B/0.4B/1.4B & CE/Recon. & multi-task prediction & interpretability/efficiency \\
02/25 & Looped Trans.~\cite{saunshi2025reasoning} & \paperlink{https://arxiv.org/abs/2502.17416 } & \qquad - & text & decoder-only-1.5B & CE/regularization & next-token prediction & latent reasoning/complex reasoning  \\
04/25 & PHD-Trans.~\cite{wu2025efficient} & \paperlink{https://arxiv.org/abs/2504.14992} & \qquad - & text & decoder-only-1.2B &  CE & next-token prediction & scaling/efficiency/exploration \\
05/25 & LARES~\cite{liu2025lares} & \paperlink{https://arxiv.org/abs/2505.16865} & \qquad - & text &  transformer-based & InfoNCE/CE/reward & alignment/next-token prediction & sequential recommendation \\
09/25 & PonderLM-2~\cite{zeng2025pretraining} & \paperlink{https://arxiv.org/abs/2509.23184} & \githublink{https://github.com/LUMIA-Group/PonderLM-2} & text & PonderLM-2-0.5B/1.4B & joint CE & next-token prediction & latent reasoning/efficiency \\
10/25 & Ouro~\cite{zhu2025scaling} & \paperlink{https://arxiv.org/abs/2510.25741} & \qquad - & text & Ouro-1.4B/2.6B & CE/KL/task loss & prediction/stability & efficiency/consistency/exploration \\
01/26 & CLAP~\cite{zhang2026clap} & \paperlink{https://arxiv.org/abs/2601.04061} & \qquad - & action & Qwen3-VL-4B & CE/Recon./task loss & quantization/alignment/prediction & embodied manipulation/transferring \\
01/26 & CARE~\cite{shi2026care} & \paperlink{https://arxiv.org/abs/2601.22467} & \qquad - & action & Prismatic-7B & MSE/task loss & multi-task prediction & embodied manipulation/scaling \\
02/26 & AL-CoT~\cite{zeng2026pretraining} & \paperlink{https://arxiv.org/abs/2602.08220} & \qquad - & text & PonderLM2-0.5B/1.4B & CE/task loss & prediction/stability & latent reasoning/dynamic reasoning \\
02/26 & ConceptLM~\cite{liu2026next} & \paperlink{https://arxiv.org/abs/2602.08984} & \githublink{https://github.com/LUMIA-Group/ConceptLM} & text & GPT2-0.1B/other 5 & CE/MSE & multi-task prediction & latent reasoning/efficiency \\
02/26 & JALA~\cite{jala2026joint} & \paperlink{https://arxiv.org/abs/2602.21736}
& \qquad - & action & InternVL3-2B & task loss & alignment/prediction & embodied manipulation/scaling \\ 
03/26 & LoopRPT~\cite{tang2026looprpt} & \paperlink{https://arxiv.org/abs/2603.19714} & \qquad - & text & Ouro-1.4B/2.6B & reward/KL/task loss & next-token prediction/sampling & complex reasoning/efficiency/scaling \\
\midrule

\multicolumn{8}{l}{\textbf{Post-training}} \\\midrule
11/24 & LaTRO~\cite{chen2024language} & \paperlink{https://arxiv.org/abs/2411.04282} & \githublink{https://github.com/SalesforceAIResearch/LaTRO} & text & Mistral-7B/other 2 & self reward/KL & finetuning/sampling & latent reasoning/zero-shot \\
01/25 & LATPC~\cite{yi2026latentspace} & \paperlink{https://arxiv.org/abs/2501.10639} & \githublink{https://github.com/xinykou/Against_Jailbreak} & text & LLaMA3-8B/other 3 & task loss & finetuning  & safety/roubustness/jailbreak attack \\
03/25 & ReaRec~\cite{tang2025think} & \paperlink{https://arxiv.org/abs/2503.22675} & \qquad - & text & LLaMA3.1-8B & CE/KL/InfoNCE & prediction/sampling & sequential recommendation \\
05/25 & CoLaR~\cite{tan2025think} & \paperlink{https://arxiv.org/abs/2505.16552} & \githublink{https://github.com/xiaomi-research/colar} & text & LLaMA3.2-1B & MSE/reward/task loss & prediction/sampling/finetuning & dynamic reasoning/efficiency \\
05/25 & LARES~\cite{liu2025lares} & \paperlink{https://arxiv.org/abs/2505.16865} & \qquad - & text & transformer-based & CE/reward/KL & alignment/sampling & sequential recommendation \\
05/25 & HRPO~\cite{yue2025hybrid} & \paperlink{https://arxiv.org/abs/2505.18454} & \githublink{https://github.com/Yueeeeeeee/HRPO} & text & Qwen2.5-7B/other 2 & reward/KL & sampling &  latent reasoning/complex reasoning \\
05/25 & LatentR3~\cite{zhang2025reinforced} & \paperlink{https://arxiv.org/abs/2505.19092} & \githublink{https://github.com/xuwenxinedu/R3} & text & Qwen2.5-1.5B & CE/reward & next-token prediction/sampling &  recommendation/efficiency \\ 
06/25 & EPR-Latent~\cite{wang2025efficient} & \paperlink{https://arxiv.org/abs/2506.08552} & \githublink{https://github.com/anord-wang/Lateng-Reasoning} & text & GPT2-0.1B/other 2 & MSE/task loss & contrast/finetuning & latent reasoning/efficiency \\
06/25 & Mirage~\cite{yang2025machine} & \paperlink{https://arxiv.org/abs/2506.17218} & \githublink{https://github.com/UMass-Embodied-AGI/Mirage} & vision & Qwen2.5-VL-7B & CE/reward/task loss & next-token prediction/sampling & visual imagination/generalization \\
09/25 & LTA-Thinker~\cite{wang2025ltathinker} & \paperlink{https://arxiv.org/abs/2509.12875} & \githublink{https://github.com/wangjiaqi886/LTA-Thinker} & text & Qwen2.5-7B/Qwen3-8B & CE/KL/task loss & finetuning/alignment/contrast & latent reasoning/scaling \\
09/25 & GainRouter~\cite{zheng2025fast} & \paperlink{https://arxiv.org/abs/2509.23633} & \qquad - & text & Qwen3-4B & task loss & alignment/finetuning & latent reasoning/adaptive reasoning\\
09/25 & MemGen~\cite{zhang2025memgen} & \paperlink{https://arxiv.org/abs/2509.24704} & \githublink{https://github.com/bingreeky/MemGen}  & text & SmolLM3-3B/Qwen3-8B & CE/reward & finetuning/sampling  & experimental memory/generalization \\
10/25 & Latent-SFT~\cite{deng2025latent} & \paperlink{https://arxiv.org/abs/2510.15522} & \githublink{https://github.com/DJC-GO-SOLO/Latent-SFT} & text & LLaMA3.2-1B/other 2 & KL/CE & finetuning & latent reasoning/efficiency \\
10/25 & 3DThinker~\cite{chen2025think} & \paperlink{https://arxiv.org/abs/2510.18632} & \githublink{https://github.com/zhangquanchen/3DThinker} & vision & Qwen2.5-VL-7B/other 7 & CE/reward/task loss & finetuning/sampling/alignment & spatial understanding/visual reasoning \\
10/25 & SemCoT~\cite{he2025semcot} & \paperlink{https://arxiv.org/abs/2510.24940} & \githublink{https://github.com/YinhanHe123/SemCoT} & text &  LLaMA2-7B/Mistral-7B & CE/CL & contrast/alignment & complex reasoning/efficiency  \\
11/25 & SofT-GRPO~\cite{zheng2025softgrpo} & \paperlink{https://arxiv.org/abs/2511.06411} & \githublink{https://github.com/zz1358m/SofT-GRPO-master} & text & DeepSeek-R1-1.5B/other 2 & KL/reward & sampling/stability & latent reasoning/generalization \\
11/25 & EBM-CoT~\cite{chen2025think1} & \paperlink{https://arxiv.org/abs/2511.07124} & \qquad - & text & Qwen2.5-7B/other 5 & CE/regularization & next-token prediction & latent reasoning/efficiency \\
11/25 & VisMem~\cite{yu2025vismem} & \paperlink{https://arxiv.org/abs/2511.11007} & \githublink{https://github.com/YU-deep/VisMem} & vision & Qwen2.5-VL-7B/other 8 & reward & sampling/alignment & visual understanding/visual reasoning \\
11/25 & SCM~\cite{wang2025improving} & \paperlink{https://arxiv.org/abs/2511.16885} & \qquad - & text & DeepSeek-R1-7B/other 3 & KL/reward & sampling & latent reasoning \\
11/25 & LWS~\cite{ning2025learning} & \paperlink{https://arxiv.org/abs/2511.21581} & \githublink{https://github.com/apning/adaptive-latent-reasoning} & text & LLaMA3.2-1B & CE/reward/task loss & prediction/alignment/sampling & dynamic reasoning/efficiency \\
11/25 & Monet~\cite{wang2025monet} & \paperlink{https://arxiv.org/abs/2511.21395} & \githublink{https://github.com/NOVAglow646/Monet} & vision & Qwen2.5-VL-7B & CE/reward/task loss & finetuning/alignment/sampling & visual reasoning/generalization \\
12/25 & ReLaX~\cite{zhang2025relax} & \paperlink{https://arxiv.org/abs/2512.07558} & \qquad - & text & Qwen2.5-3B/other 4 & reward/regularization & sampling & exploration/generalization \\  
12/25 & RL-Latent~\cite{enes2025reinforcement} & \paperlink{https://arxiv.org/abs/2512.11816} & \githublink{https://github.com/enesozeren/latent-space-thinking-model} & text & Qwen2.5-1.5B & CE/reward/KL & prediction/finetuning/sampling & latent reasoning/complex reasoning \\
01/26 & I2B-LPO~\cite{deng2026iiblpo} & \paperlink{https://arxiv.org/abs/2601.05870} & \githublink{https://github.com/denghuilin-cyber/IIB-LPO} & text & Qwen3-14B/Qwen2.5-37B  & self-reward/KL & sampling & exploration/policy optimization \\
01/26 & GeoSteer~\cite{kazama2025geosteer} & \paperlink{https://arxiv.org/abs/2601.10229} & \qquad - & text & Qwen3-8B /other 3 & Recon./KL/task loss & alignment & complex reasoning/consistency\\
01/26 & DLR~\cite{shan2026latentspace} & \paperlink{https://arxiv.org/abs/2601.17275} & \qquad - & text & LLaMA2-1B/7B & reward/regularization & sampling/stability/contrast & exploration/efficiency \\
01/26 & PILOT~\cite{zheng2026pilot} & \paperlink{https://arxiv.org/abs/2601.19917} & \qquad - & text & Qwen2.5-1.5B/other 2 & CE/regularization & finetuning/alignment & robustness/generalization \\
01/26 & ATP-Latent~\cite{zheng2026beyond} & \paperlink{https://arxiv.org/abs/2601.21598} & \githublink{https://github.com/zz1358m/ATP-Latent-master} & text & LaMA3.2-1B &  CE/KL/reward & finetuning/sampling & latent reasoning/generalization \\
01/26 & GANPO~\cite{jiang2026latent} & \paperlink{https://arxiv.org/abs/2601.22083} & \qquad - & text & Gemma2-2B/LLaMA3-8B & CE/regularization & alignment & preference optimizatio  \\ 
01/26 & BNPO~\cite{li2026controlling} & \paperlink{https://arxiv.org/abs/2601.07516} & \qquad - & action & Qwen2.5-VL-3B/7B & Recon./task loss & alignment/sampling & efficiency/generalization \\
01/26 & LaViT~\cite{wu2026lavit} & \paperlink{https://arxiv.org/abs/2601.10129} & \githublink{https://github.com/Svardfox/LaViT} & vision & Qwen2.5-VL-3B & CE/Recon./task loss & next-token prediction/alignment & visual reasoning/efficiency/robustness \\
01/26 & RoT~\cite{wang2026renderofthought} & \paperlink{https://arxiv.org/abs/2601.14750} & \githublink{https://github.com/TencentBAC/RoT} & vision & Qwen3-VL-4B/other 2 & MSE/CE & finetuning/prediction & visual reasoning/accelerating/efficiency \\
02/26 & CogSense~\cite{li2026toward} & \paperlink{https://arxiv.org/abs/2602.01541} & \qquad - & vision & Qwen3-VL-8B & MSE/CE/task loss & finetuning/prediction/alignment & visual imagination/generalization \\
02/26 & MaD-Mix~\cite{xie2026madmix} & \paperlink{https://arxiv.org/abs/2602.07790} & \qquad - & vision & LLaVA-OneVision-0.5B/7B & task loss & alignment & visual reasoning/efficiency \\
02/26 & TS~\cite{amos2026latent} & \paperlink{https://arxiv.org/abs/2602.08332} & \qquad - & text & Qwen2.5-0.5B/1.5B & CE & next-token prediction/alignment & latent reasoning/efficiency  \\
02/26 & Reason-IAD~\cite{chen2026reason} & \paperlink{https://arxiv.org/abs/2602.09850} & \githublink{https://github.com/chenpeng052/Reason-IAD} & vision & Qwen2.5-
VL-7B/other 3 & self-reward & sampling/alignment & visual detection \\ 
02/26 & RLTT~\cite{williams2026prioritize} & \paperlink{https://arxiv.org/abs/2602.10520} & \qquad - & text & Ouro-2.6B & reward/regularization & sampling/alignment & latent reasoning/generalization \\
03/26 & SPOT~\cite{chu2026spot} & \paperlink{https://arxiv.org/abs/2603.06222} & \qquad - & text & DeepSeek-R1-7B & CE/task loss & prediction/alignment/finetuning & latent reasoning/efficiency \\

\midrule
\multicolumn{8}{l}{\textbf{Inference}} \\\midrule
10/24 & VTI~\cite{liu2025reducing} & \paperlink{https://arxiv.org/abs/2410.15778} & \githublink{https://github.com/shengliu66/VTI} & vision & LLaVA1.5-7B/other 2 & CL & contrast & hallucination mitigation \\
05/25 & SoftCoT++~\cite{xu2025softcot1} & \paperlink{https://arxiv.org/abs/2505.11484} & \githublink{https://github.com/xuyige/SoftCoT} & text & LLaMA3.1-8B & CL & contrast/sampling & exploration/zero-shot/scaling \\
05/25 & LatentSeek~\cite{li2025latentseek} & \paperlink{https://arxiv.org/abs/2505.13308} & \githublink{https://github.com/bigai-nlco/LatentSeek} & text & LLaMA3.1-8B & self-reward & sampling & forgetting alleviation/scaling \\
05/25 & SR~\cite{zhu2025soft} & \paperlink{https://arxiv.org/abs/2505.24688} & \githublink{https://github.com/alickzhu/Soft-Reasoning} & text & LLaMA3.1-8B/other 4 & self-reward & sampling/alignment & exploration/efficiency \\
08/25 & LatentPrompt~\cite{bystronski2025latentprompt} & \paperlink{https://arxiv.org/abs/2508.02452} & \qquad - & text & Mistral-7B & self-reward/task loss & sampling & prompt optimization \\
09/25 & LatentEvolve~\cite{zhang2025latentevolve} & \paperlink{https://arxiv.org/abs/2509.24771} & \githublink{https://github.com/jins7/LatentEvolve} &  text & LLaMA3.2-3B/other 3 & self-reward/Recon. & sampling/finetuning &  experimental memory \\
09/25 & LTO~\cite{du2025latent} & \paperlink{https://arxiv.org/abs/2509.26314} & \qquad - & text & Huginn-3.5B/other 4 & reward/KL & sampling/finetuning/alignment & efficiency/scaling \\
10/25 & LTPO~\cite{ye2025thinking} & \paperlink{https://arxiv.org/abs/2510.04182} & \githublink{https://github.com/ltpo2025/LTPO} & text & LLaMA3.1-8B/other 3 & self reward & sampling/alignment & generalization/unsupervised learning \\
11/25 & L2V-CoT~\cite{zhan2025l2vcot} & \paperlink{https://arxiv.org/abs/2511.17910} & \qquad - & vision & LLaVA-Next-8B/other 4 & task loss & contrast/alignment & transferring/visual reasoning \\
12/25 & DMLR~\cite{liu2025reason} & \paperlink{https://arxiv.org/abs/2512.12623} & \githublink{https://github.com/eric-ai-lab/DMLR} & vision & Qwen2.5-VL-7B/other 5 & self-reward & sampling/alignment & visual understanding/grounding \\
01/26 & TGR~\cite{zhuang2026geometric} & \paperlink{https://arxiv.org/abs/2601.18832} & \qquad - & text & Qwen3-8B/other 3 & self-reward & sampling & exploration/long-horizon \\
02/26 & STIR~\cite{shi2026internalizing} & \paperlink{https://arxiv.org/abs/2602.04925} & \githublink{https://github.com/sznnzs/LLM-Latent-Action} & text & DeepSeek-R1-7B/other 3 & CL/self-reward & sampling/alignment/contrast & complex reasoning/efficiency \\
02/26 & ITR~\cite{kong2026inference} & \paperlink{https://arxiv.org/abs/2602.06584} & \qquad - & text & LLaMA2-0.2B & KL/task loss & sampling & complex reasoning/efficiency \\ 
02/26 & REVIS~\cite{wu2026revis} & \paperlink{https://arxiv.org/abs/2602.11824} & \githublink{https://github.com/antgroup/Revis} & text & Qwen2.5-VL-7B/other 4 & CL & contrast/alignment & hallucination mitigation \\
02/26 & GTS~\cite{wang2026gts} & \paperlink{https://arxiv.org/abs/2602.14077} & \qquad - & text & GPT2-0.1B/LLaMA3.2-1B & reward/KL & sampling/ailgnment & exploration/latent reasoning \\
02/26 & Control++~\cite{wu2026test} & \paperlink{https://arxiv.org/abs/2602.19505} & \qquad - & vision & LLaVA1.5-7B & task loss & alignment & hallucination mitigation \\
03/26 & $\nabla$-Reasoner~\cite{wang2026nabla} & \paperlink{https://arxiv.org/abs/2603.04948} & \qquad - & text & Qwen-2.5-7B/other 2 & reward/KL & sampling/alignment & complex reasoning/exploration \\
\bottomrule
\end{tabular}
}
\end{table*}

\subsubsection{Post-training}
\label{sec:post_training}
During the post-training stage, $\mathbf{z}$ is further optimized via fine-tuning on a pre-trained model.
Thanks to the reduced data requirements compared to pre-training, post-training supervision can involve more sophisticated data augmentation or specialized loss designs.
Many methods focus on defining an additional signal to supervise the latent variable, as supervision on explicit variables alone may not be sufficient.
Again, the optimized variable is still model parameters, but the optimization is now restricted to a post-training objective:
\begin{equation}
    \theta^{\star}
    =
    \arg\min_{\theta \in \mathcal{W}}
    \mathbb{E}_{(\mathbf{x},\mathbf{y})\sim\mathcal{D}}
    \left[
        \mathcal{L}(\mathbf{x}, \mathbf{y}, \mathbf{z}; \Phi_\theta)
    \right]
    -
    \beta\,
    \mathbb{E}_{\mathbf{r}\sim\Phi_\theta(\cdot\mid\mathbf{x})}
    \left[
        R(\mathbf{x}, \mathbf{r}, \mathbf{z})
    \right],
\end{equation}
where the formulation subsumes supervised fine-tuning, preference optimization, and reinforcement-learning-based alignment.
In latent-space methods, post-training often specializes in refining how latent variables are induced, controlled, or aligned with downstream objectives. Based on the supervision signals, there are three types, including: \textbf{Explicit Supervision}, \textbf{Implicit Supervision}, and \textbf{Reinforcement Learning}.

\xhdr{Explicit Supervision Fine-Tuning} This category optimizes latent reasoning by applying loss functions solely to the final human-readable output, without providing step-by-step targets for the continuous representations. Fine-tuning continuous variables via specific task losses, LATPC~\cite{yi2026latentspace} enhances safety and robustness against jailbreak attacks. Using task loss alignment, GainRouter~\cite{zheng2025fast} dynamically routes features to enable fast and adaptive latent reasoning. Steering hidden states on a learned latent manifold, GeoSteer~\cite{kazama2025geosteer} ensures faithful and logically consistent reasoning trajectories. Fine-tuning models with cross-entropy and regularization, PILOT~\cite{zheng2026pilot} internalizes the strategic oversight of large models into intrinsic latent guidance. Focusing on next-token prediction and alignment, TS~\cite{amos2026latent} utilizes cross-entropy to streamline latent reasoning.

\xhdr{Implicit Supervision Fine-Tuning} This approach provides explicit gold targets for latent representations using knowledge distillation, contrastive alignment, or step-wise reconstruction signals. 

Distillation-based methods anchor latent states to teacher-provided signals. SPOT~\cite{chu2026spot} compresses explicit reasoning trajectories into compact latent tokens by anchoring them to corresponding teacher spans, while SemCoT~\cite{he2025semcot} distills ground-truth reasoning into semantically aligned implicit tokens to accelerate computation. Latent-SFT~\cite{deng2025latent} takes a different angle, redefining latent tokens as vocabulary-space superpositions and training them with KL divergence and cross-entropy losses.

Contrastive learning constitutes another prominent technique. LTA-Thinker~\cite{wang2025ltathinker} and EPR-Latent~\cite{wang2025efficient} employ contrastive objectives to optimize latent thought distributions and refine intermediate representations, respectively, while ReaRec~\cite{tang2025think} applies the same principle to sequential recommendation. EBM-CoT~\cite{chen2025think1} further combines contrastive signals with regularization to improve overall latent reasoning efficiency.

Reconstruction objectives are especially prevalent in multimodal settings, where aligning across modalities demands explicit feature-level supervision. LaViT~\cite{wu2026lavit} jointly predicts next tokens and reconstructs visual features to align cross-modal reasoning, and RoT~\cite{wang2026renderofthought} takes a distinctive approach by rendering chain-of-thought as images and applying MSE together with cross-entropy. BNPO~\cite{li2026controlling} uses reconstruction and task losses to align embodied action spaces. Methods such as Mirage~\cite{yang2025machine}, 3DThinker~\cite{chen2025think}, VisMem~\cite{yu2025vismem}, Monet~\cite{wang2025monet}, and CogSense~\cite{li2026toward} further demonstrate the breadth of this paradigm across visual imagination, spatial reasoning, and generalization tasks.

\xhdr{Reinforcement Learning} To mitigate geometric drift, this sub-category leverages policy gradients, rewards, and preference signals to autonomously discover efficient latent trajectories. 

Self-rewarding mechanisms form one important thread, where the model itself provides the training signal without external annotation. LaTRO~\cite{chen2024language} formulates reasoning as a variational sampling process optimized via the model's own probability estimates, and I2B-LPO~\cite{deng2026iiblpo} employs an iterative information bottleneck with self-rewards for thorough latent policy exploration. MemGen~\cite{zhang2025memgen} similarly relies on targeted self-rewards to enhance generalization.

A further set of methods introduces specialized regularizers to stabilize or enrich the optimization landscape. SofT-GRPO~\cite{zheng2025softgrpo} applies Gumbel reparameterization for stable group relative policy optimization, GANPO~\cite{jiang2026latent} introduces an adversarial regularizer to robustify preference optimization, and DLR~\cite{shan2026latentspace} combines contrastive stability constraints with reward signals for directed latent exploration. HRPO~\cite{yue2025hybrid} takes a hybrid approach, using a learnable gate to progressively integrate continuous hidden states with discrete tokens. CoLaR~\cite{tan2025think}, LWS~\cite{ning2025learning}, LatentR3~\cite{zhang2025reinforced}, and Reason-IAD~\cite{chen2026reason} round out the landscape by applying reward-driven optimization to dynamic trajectory prediction, sequential recommendation, and visual spatial understanding. KL divergence regularization also serves as a recurring stabilization mechanism across many methods. LARES~\cite{liu2025lares}, RL-Latent~\cite{enes2025reinforcement}, SCM~\cite{wang2025improving}, and ATP-Latent~\cite{zheng2026beyond} all incorporate KL penalties alongside reward objectives to prevent excessive deviation from the reference policy during latent space optimization. ReLaX~\cite{zhang2025relax} further addresses premature convergence through strict reward regularization, and RLTT~\cite{williams2026prioritize} distributes rewards across the full trajectory to improve alignment.

\xhdr{Summary} Compared to pre-training, post-training affords greater flexibility in supervision design, enabling richer signals such as distillation, contrastive alignment, and reward-based feedback to refine latent representations. A central tension in this stage is whether to supervise latent variables explicitly through gold targets or implicitly through output-level task losses alone. Reinforcement learning methods go further by treating latent trajectory discovery as a policy optimization problem, allowing models to autonomously identify compact and efficient reasoning paths without relying on predetermined supervision.

\subsubsection{Inference}
\label{sec:inference}
For inference-time latent optimization, the model weights $\theta$ are usually frozen (also could be trained) and the latent states $\mathbf{z}$ are directly manipulated at test time.
Unlike pre-training and post-training, inference-time methods treat $\mathbf{z}$ itself as the optimization variable.
Formally, let $\theta$ denote the trained parameters, then the optimized variable becomes $\omega$, with:
\begin{equation}
    \mathbf{z}^{\star}
    =
    \arg\min_{\omega\in\Omega}
    \mathcal{J}
    \!\left(
        \mathbf{z};
        \mathbf{x},
        \Phi_{\theta}
    \right).
\end{equation}
where $\Omega$ is the feasible region of the variable, and $\mathcal{J}$ is the inference-time objective, final output is then generated conditioned on the optimized inference-time state. The part includes \textbf{Scaling}, \textbf{Tuning}, and \textbf{Guidance}.

\xhdr{Inference Scaling} This category focuses on exploring the latent space by generating multiple candidate trajectories and employing reward-based heuristics to identify the optimal reasoning path. Employing a continuous-space classifier as a latent reward model, LTO~\cite{du2025latent} aggressively prunes incorrect thinking patterns during inference. Performing manifold-informed latent foresight search, TGR~\cite{zhuang2026geometric} scores candidate latent anchors to encourage smooth trajectories and diverse long-horizon exploration. Reformulating latent exploration as conditional sampling over continuous thought representations, GTS~\cite{wang2026gts} utilizes a Gaussian Thought Sampler to predict context-dependent perturbation distributions. Applying self-reward sampling, LatentSeek~\cite{li2025latentseek} effectively alleviates catastrophic forgetting during continuous generation. Employing self-rewards, SR~\cite{zhu2025soft} samples and aligns soft reasoning explorations to maximize efficiency. Leveraging the latent semantic space, LatentPrompt~\cite{bystronski2025latentprompt} automatically evaluates and optimizes prompts via intrinsic self-rewards. Utilizing KL divergence and task losses, ITR~\cite{kong2026inference} guides inference-time sampling to greatly improve complex reasoning efficiency. Enhancing visual understanding and grounding, DMLR~\cite{liu2025reason} implements continuous self-reward sampling and alignment.

\xhdr{Inference Tuning} This approach shifts from trial-and-error stochastic search to continuous, gradient-based optimization executed directly on the latent variables or policy during the forward pass. Optimizing intermediate latent thought vectors on the fly, LTPO~\cite{ye2025thinking} utilizes an online policy gradient method guided by an intrinsic confidence-based reward. Integrating differentiable optimization over token logits, $\nabla$-Reasoner~\cite{wang2026nabla} refines the policy directly within the decoding loop via gradient descent in the continuous sample space. Combining self-reward and reconstruction losses, LatentEvolve~\cite{zhang2025latentevolve} dynamically samples and fine-tunes experimental memory states through test-time scaling.

\xhdr{Inference Guidance} This category applies targeted structural constraints, contrastive logic, or sparse interventions to dynamically guide latent representations and prevent hallucinations. Steering internal activations via sparse interventions, REVIS~\cite{wu2026revis} mitigates object hallucination in large vision-language models by extracting pure visual information vectors. Internalizing contrastive learning and self-rewards, STIR~\cite{shi2026internalizing} introduces a value-modulated trajectory intervention that dynamically injects context-specific impulses via anchor-based gating. Perturbing latent thoughts via specialized initial tokens, SoftCoT++~\cite{xu2025softcot1} uses contrastive learning to enforce diversity among soft representations. Utilizing contrastive learning at inference time, VTI~\cite{liu2025reducing} successfully mitigates hallucinations in massive visual models. Contrasting visual features at test time, L2V-CoT~\cite{zhan2025l2vcot} tightly aligns transferring and visual reasoning capabilities through latent intervention. Preventing visual hallucinations dynamically, Control++~\cite{wu2026test} applies highly targeted task losses during the alignment generation process.

\xhdr{Summary} Unlike parameter-level optimization, inference-time methods treat latent states themselves as the optimization variable while keeping model weights fixed. The key distinction among approaches lies in the search strategy: scaling methods explore the latent space stochastically through reward-guided trajectory selection, optimization methods apply gradient updates directly to latent variables during decoding, and guidance methods impose structural or contrastive constraints to correct latent representations on the fly, particularly to suppress hallucinations.

\section{\textcolor{secyellow}{Ability:} What Does Latent Space Enable?}
\label{sec:ability}
\begin{figure*}[t]
  \centering
    \includegraphics[width=0.85\linewidth]{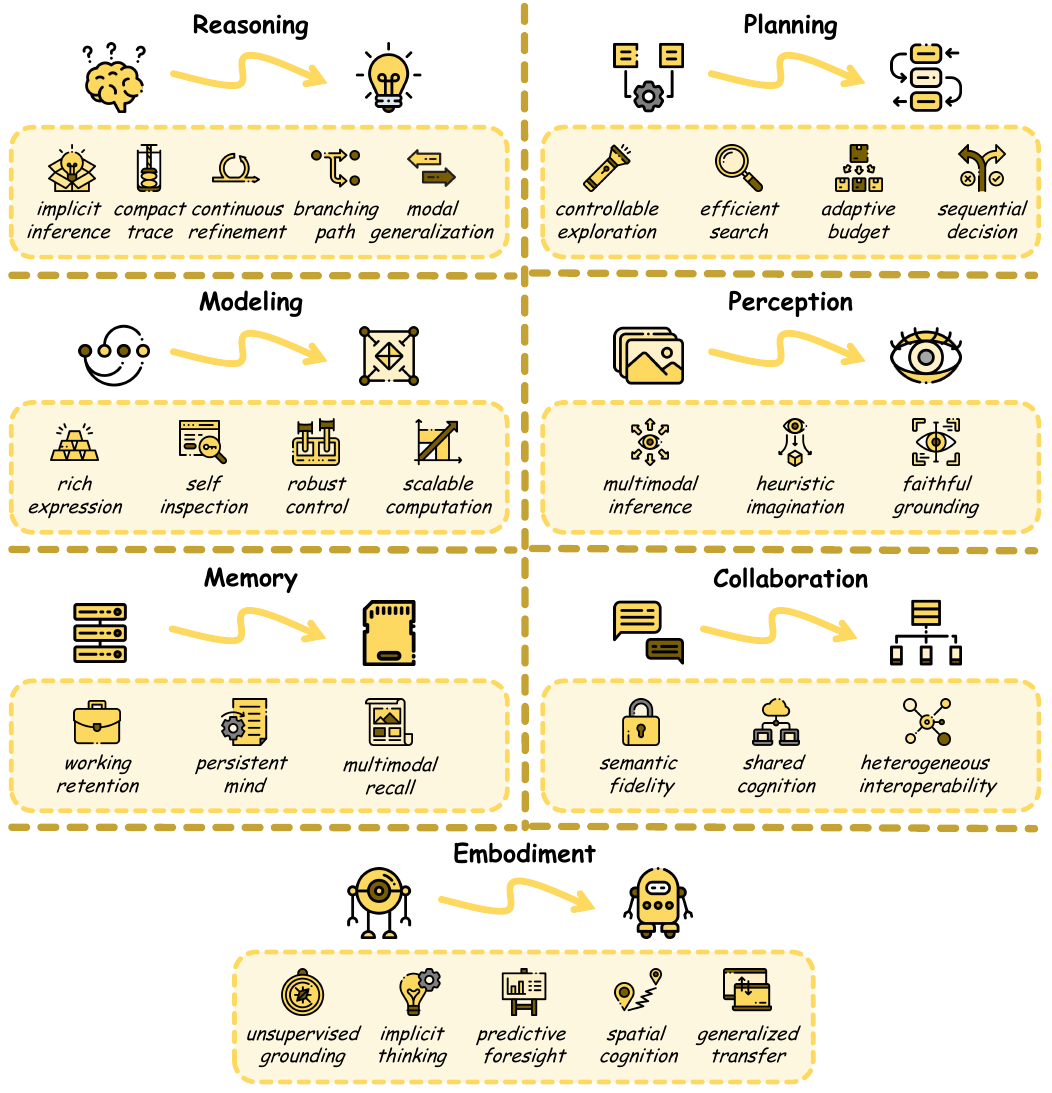}
    \caption{Core abilities brought by the latent space, including: \textbf{Reasoning} (Section~\ref{sec:reasoning}), \textbf{Planning} (Section~\ref{sec:planning}), \textbf{Modeling} (Section~\ref{sec:modeling}), \textbf{Perception} (Section~\ref{sec:perception}), \textbf{Memory} (Section~\ref{sec:memory}), \textbf{Collaboration} (Section~\ref{sec:collaboration}), and \textbf{Embodiment} (Section~\ref{sec:embodiment}).}
    \label{fig:ability}
\end{figure*}

The latent space, as a machine-native representational substrate within large models, unlocks a spectrum of emergent capabilities that transcend the boundaries of explicit token-level processing.
In this section, we systematically examine these capabilities along seven dimensions: \textbf{Reasoning} (Section~\ref{sec:reasoning}) concerns the ability to carry out deduction and relational computation through continuous internal states; \textbf{Planning} (Section~\ref{sec:planning}) emphasizes the prospective organization of solution trajectories, resource allocation, and multi-step decision-making; \textbf{Modeling} (Section~\ref{sec:modeling}) focuses on the characterization, interpretability, controllability, and scalable depth of latent representations themselves; \textbf{Perception} (Section~\ref{sec:perception}) enables models to preserve and manipulate rich, spatially structured information for more faithful visual understanding; \textbf{Memory} (Section~\ref{sec:memory}) supports compact, persistent, and adaptive knowledge retention across contexts; \textbf{Collaboration} (Section~\ref{sec:collaboration}) allows multiple agents to exchange semantic content directly through latent channels rather than discrete language alone; and \textbf{Embodiment} (Section~\ref{sec:embodiment}) extends latent computation into physical interaction, supporting grounded action, predictive foresight, spatial imagination, and transfer across heterogeneous bodies. 
As depicted in Figure~\ref{fig:ability}, each dimension reflects a distinct facet of intelligence that latent representations uniquely empower, ranging from internal logical deduction to physical interaction with the environment.

\subsection{Reasoning}
\label{sec:reasoning}

Reasoning in latent space refers to the capacity of large models to perform logical deduction, relational computation, and conclusion generation through internal continuous representations rather than through explicit token-by-token verbalization. 
The shift from explicit CoT reasoning~\cite{wei2022chain} to latent reasoning represents a fundamental paradigm change: instead of articulating every intermediate step in natural language, models learn to think within a continuous, high-dimensional latent manifold~\cite{hao2024training,zhu2025reasoning}.

This paradigm offers substantial advantages in computational efficiency, representational capacity, and the ability to encode superpositions of multiple reasoning paths simultaneously. We organize this rapidly expanding landscape along six abilities: \textbf{Implicit Inference} without full verbalization, \textbf{Compact Trace} that condenses long chains into compact states, \textbf{Continuous Refinement} that sustains and revises thought in latent form, \textbf{Branching Path} over multiple candidates, and \textbf{Modal Generalization} beyond text-only settings.

\xhdr{Implicit Inference}
A central motivation for moving reasoning into latent space is the growing evidence that explicit CoT, while interpretable, is often redundant and fundamentally constrained by the discrete, sequential nature of language. Converging evidence from CoT compression~\cite{liu2024expediting,zhang2024uncovering}, implicit capability elicitation~\cite{chen2024language,implicit2025li}, and latent self-evaluation~\cite{wang2025latent} shows that reasoning-like behavior is already substantially encoded in the continuous activation spaces of pretrained models.
Further analyses establish latent reasoning as a distinct computational mode encompassing richer, non-sequential inference~\cite{he2026reasoning,reasoning2025chen}, collectively challenging the assumption that reasoning must be externalized in tokens.
Building on these insights, COCONUT~\cite{hao2024training} demonstrated that continuous thought vectors can encode superpositions of multiple reasoning paths, enabling emergent breadth-first search and showing that models can infer internally before committing to language.

\xhdr{Compact Trace}
One major ability unlocked by latent space is compressing explicit CoT into far more compact internal states without losing its problem-solving value.
A broad range of supervision and post-training studies~\cite{cheng2024compressed,shen2025codi,wei2025simcot,he2025semcot,wang2025efficient,shen2025efficient,zheng2025fast} shows that long reasoning traces can be absorbed into compact latent states while preserving much of their functional value.
This evidence suggests that already-trained models can acquire compressed reasoning ability with only modest additional overhead.
The collective evidence indicates that reasoning chains can be preserved in representations orders of magnitude more compact, revealing fundamental redundancy in token-level reasoning.

\xhdr{Continuous Refinement}
Beyond compressing existing CoT, latent space also supports the ability to sustain, blend, and iteratively revise thought as a continuous state.
Soft token methods~\cite{zhang2025soft,xu2025softcot,wang2025improving} replace discrete sampling with probability-weighted embedding mixtures or learned concept-level blending, while stochastic refinement through diffusion~\cite{kang2025ladir} and Markov dynamics~\cite{liu2025marcos} enables revision of earlier reasoning decisions.
Energy-based consistency enforcement~\cite{chen2025think1} and theoretical analyses of vocabulary-space superposition~\cite{deng2025latent,gozeten2025continuous} further show that such latent states can preserve coherence while remaining amenable to simultaneous optimization.
Thought augmentation and contextualization~\cite{wang2025ltathinker,liu2026latent,sun2025enhancing,ruan2025reasoning,long2025reasoning} enrich this refinement ability by fusing task-relevant and background knowledge and modulating reasoning through targeted interventions.

\xhdr{Branching Path}
The continuous nature of latent space enables fundamentally new reasoning topologies that let models explore several candidate trajectories at once.
Parallel latent reasoning~\cite{xu2025softcot1,liu2025thoughtbubbles,you2025parallel,wu25parallel} demonstrates simultaneous search through soft path sampling, stochastic width, and Jacobi iteration, reducing wall-clock latency while maintaining quality.
Hybrid latent-explicit systems~\cite{piao2025spiralthinker,su2025token,shi2025swireasoning,xu2026thinkrouter} further suggest that models can flexibly alternate between compact internal search and selective externalization when beneficial.
Studies of dual-system orchestration~\cite{codaforno2025exploring} reinforce this view, indicating that strong reasoners benefit from coordinating multiple reasoning paths and modes rather than committing to a single linear trace.

\xhdr{Modal Generalization}
A key indicator of the maturity of latent reasoning is its ability to generalize beyond text-only settings.
Modality-agnostic continuous thought~\cite{pham2025multimodal,ray2025mulltokens} demonstrates that the latent reasoning paradigm applies across linguistic, visual, and heterogeneous substrates, while cross-modal transfer and efficiency techniques~\cite{zhan2025l2vcot,shao2026learning,wang2026renderofthought,lv2026onelatent} show that this capability can move across modalities with increasing compactness.
We defer detailed treatment of visual latent reasoning to the Perception subsection (Section~\ref{sec:perception}).

Beyond vision, domain-specific applications spanning chemical synthesis~\cite{armstrong2025synthstrategy}, narrative generation~\cite{gurung2025lightweight}, novelty discovery~\cite{bystronski2025large}, and joint-embedding prediction~\cite{liu2025jepareasoner} demonstrate broad applicability.
Structured spatial-temporal reasoning~\cite{xue2025reasoning,vilas2025tracing} extends this generalization to geometric and temporal domains, while faithfulness and controllability studies~\cite{lukas2025activationreasoning,zhang2026silence,ye2026riser,wang2026regular} show that latent reasoning can remain reliable as it moves across settings.

\subsection{Planning}
\label{sec:planning}

Planning concerns the search for optimal trajectories through the solution landscape, where the continuous, differentiable nature of the latent manifold admits gradient-based policy optimization and iterative trajectory refinement. 
Unlike reasoning, which focuses on logical deduction within a given context, planning emphasizes the prospective organization of computation, determining where to allocate resources, how to explore the solution space, and when to terminate search~\cite{wang2026latent,xu2026no}.
We examine latent planning through four abilities: \textbf{Controllable Exploration} over internal solution paths, \textbf{Search Efficiency} in navigating the latent manifold, \textbf{Adaptive Budget} allocation that matches compute to difficulty, and \textbf{Sequential Decision} in downstream interactive tasks.

\xhdr{Controllable Exploration}
A central ability in latent planning is controlling internal solution trajectories rather than merely generating the next token greedily.
RL-based trajectory optimization~\cite{du2025latent,yue2025hybrid,zheng2025softgrpo,ye2025thinking} shows that continuous thought representations can be directly improved via policy gradients, Gumbel reparameterization, and test-time refinement.
Training stability and diversity remain active challenges, addressed through exploration collapse prevention~\cite{deng2026iiblpo}, systematic design analysis~\cite{enes2025reinforcement}, and contrastive reward shaping~\cite{shan2026latentspace}.
These works collectively establish that latent geometry enables deliberate trajectory improvement that is fundamentally difficult in discrete token space.

\xhdr{Efficient Search}
The latent manifold provides a natural substrate where geometric smoothness and continuity can be exploited for efficient navigation.
Exploration restoration and trajectory diversification~\cite{zhang2025relax,xiaomi2026led,zhu2025soft} maintain reasoning variety through entropy exploitation, latent decoding, and controlled embedding exploration, while geometry-guided search~\cite{li2025latentseek,zhuang2026geometric,shi2026internalizing} leverages intrinsic manifold properties for instance-level targeting and long-context foresight.

\xhdr{Adaptive Budget}
A defining characteristic of latent planning is dynamic, input-dependent resource allocation.
Adaptive depth and horizon determination~\cite{he2025learning,ning2025learning,wang2026latent,zheng2026beyond} adjusts reasoning depth through instance-level steering, RL-based stopping, dynamic termination, and active budget determination, investing computation proportionally to problem complexity.

\xhdr{Sequential Decision}
Latent planning has been productively deployed in sequential decision-making domains, where the temporal structure of user behavior or system states naturally maps onto trajectory optimization in latent space.In recommendation, retrieval,and cross-domain adaptation~\cite{tan2025think,liu2025lares,zhang2025reinforced, tang2026parallel,guo2026s2gr,zhang2026reasoning,shi2025bridging, wang2025decoding}, latent planning improves sequential 
prediction, re-ranking, and transfer by maintaining and refining internal trajectories over time. In multi-step planning and tool use~\cite{wu2025ctrls, chen2025iclp,zhu2026colt,zheng2026pilot,li2026controlling,
zeng2026latent}, it supports sustained state tracking, optimization over intermediate decisions, and tighter control of multimodal agents, and further extends to reparameterizing the action space itself---compressing recurrent low-entropy scaffolds into compact latent action units to directly reduce the effective decision horizon while preserving executability. The breadth of these applications, spanning recommendation systems, information retrieval, tool use, action representation learning, and conversational AI, establishes latent planning as a versatile paradigm that extends well beyond the scope of traditional reasoning benchmarks.

\subsection{Modeling}
\label{sec:modeling}
Modeling encompasses the ability to characterize, inspect, and shape latent representations within large language models.
While reasoning and planning concern what models compute in latent space, modeling focuses on what latent representations let us understand and control about the computation itself. We structure this dimension into four abilities: \textbf{Rich Expression} to encode complex computation, \textbf{Self Inspection} that makes internal states analyzable, \textbf{Robust Control} over risky or unstable behavior, and \textbf{Scalable Computation} that expands capacity through latent recurrence.

\xhdr{Rich Expression}
Rigorous analysis increasingly shows that latent space supports a richer computational capacity than explicit token-only reasoning.
Expressiveness analyses~\cite{zhu2025reasoning,xu2025formal} prove that continuous thought vectors encode multiple search frontiers simultaneously and achieve provably greater expressiveness than CoT, while monitorability analysis~\cite{korbak2025chain} reveals the associated trade-offs between efficiency and interpretability.
Fundamental limitation results~\cite{zou2026capabilities,sahoo2026shallow} show that exploration and execution cannot be simultaneously optimized within fixed budgets, and that reasoning depth correlates weakly with correctness under certain conditions.
Cognitive and domain-specific frameworks~\cite{hu2024understanding,sharma2025analyzing} frame reasoning as transitions across representation spaces, extending insights from neuroscience to program understanding, together clarifying both the power and the boundaries of latent reasoning.

\xhdr{Self Inspection}
Understanding the internal dynamics of latent reasoning is critical because latent space increasingly supports direct inspection of what the model is representing and how those states evolve.
Validity probing~\cite{zhang2025do,liang2026do} examines whether latent tokens encode genuine reasoning or exploit artifacts and whether models truly reason step-by-step or develop qualitatively different strategies.
Transparency and visualization methods~\cite{chen2025latent,ning2025visualizing,liu2026unicog} make internal representations interpretable through latent debate, geometry visualization, polarity-aware probing, and cognitive analysis.
Representation-level analysis~\cite{giorgos2025language,reichman2025emotions} demonstrates information preservation and reveals rich semantic dimensions beyond task performance, while information flow studies~\cite{liu2026layerorder,wezuke2025transfer} uncover multi-hop reasoning paths and cross-lingual transition mechanisms.
Practical steering~\cite{kazama2025geosteer,kamai2026talking} further suggests that inspection is not merely descriptive, but can directly support targeted improvements.

\xhdr{Robust Control}
Latent space provides a powerful but double-edged lever for model safety, because the same representations that enable strong performance can also be manipulated for both attack and defense.
On the one hand, attack vectors~\cite{xing2025latent,mura2025latentbreak} exploit latent fusion and feedback-based gradients, alongside backdoor triggers embedded in latent CoT.
On the other hand, layered defense mechanisms~\cite{shu2025latentguard,yi2026latentspace,yang2025lfsteering,li2025kelp} provide controllable steering, adversarial training, feature activation steering, and streaming risk detection.
Training-phase safety~\cite{jiang2026latent,tang2026from,ball2025human} addresses adversarial regularization, contrastive unlearning, and human preference modeling, underscoring that robust latent control is essential for building safe systems.

\xhdr{Scalable Computation}
Modeling also highlights the ability of latent systems to expand effective depth and capacity without proportionally expanding explicit token generation.
Formal expressiveness results~\cite{saunshi2025reasoning} confirm that looped transformers with latent iterations express strictly more complex computations than feedforward counterparts.
Recurrent-block scaling~\cite{jonas2025scaling,zhu2025scaling,koishekenov2025encode,altabaa2025unlocking} iterates shared blocks for test-time compute, scales looped pretraining, introduces encode-think-decode architectures, and demonstrates out-of-distribution generalization.
Progressive refinement and architectural variants~\cite{lu2025latent,knupp2026depthrecurrent,jeddi2026loopformer,yu2026spiralformer} extend this flexibility through depth-recurrent attention, elastic depth, and multi-resolution recursion.
Adaptive depth allocation~\cite{fu2025tah,wang2025system,li2026ponderlm,song2026adaponderlm} further enables dynamic computation through selective iteration, dual-process routing, and token-wise pondering.

Beyond recurrence, concept-level and efficiency innovations~\cite{qu2025dynamic,liu2026multi} operate on adaptive semantic boundaries and achieve KV cache efficiency, while representation-level design~\cite{bystronski2025latentprompt,wu2025efficient,wang2025decoding,qiu2025latent} bypasses discrete bottlenecks in prompt optimization, pretraining scaling, and deployed system efficiency.

\subsection{Perception}
\label{sec:perception}
Perception in latent space addresses the fundamental challenge of enabling large models, particularly VLMs, to understand, represent, and process visual information in continuous, high-fidelity latent spaces.
Current VLMs still face a critical bottleneck: converting rich visual content into discrete text tokens inevitably discards spatial structure, fine-grained detail, and relational geometry~\cite{perception2025bigverdi,li2025latent}.
Latent perception overcomes this limitation by preserving dense, spatially-structured information that discrete tokenization necessarily destroys, enabling models to reason about visual content with the richness and nuance of human perception.
We organize latent perception into three progressively deeper abilities: \textbf{Multimodal Inference} over internal visual representations, \textbf{Heuristic Imagination} for generative manipulation and 3D understanding, and \textbf{Faithful Grounding} that improves output faithfulness through representation-level intervention.

\xhdr{Multimodal Inference}
A primary thrust is enabling VLMs to reason about visual content through internal latent representations rather than text-mediated descriptions.
Foundational latent visual reasoning~\cite{li2025latent,wang2025monet} demonstrated that generating and updating latent visual states alongside text enables fine-grained visual understanding unattainable through text-only reasoning.
Follow-up work~\cite{sun2025latent,dong2025interleaved,li2025latent1,wang2026forest,chen2025vljepa} shows that this visual inference ability can remain both accurate and efficient through selective computation, coarse-to-fine processing, and joint embedding prediction without pixel-level reconstruction.
Multimodal coordination and alignment studies~\cite{chen2025reasoning,liu2025reason,jeon2026vision,zhang2026multimodal,li2026toward,yang2026crystal,serussi2026pregen,wu2026revis} further show that structured visual latents can be enriched, stabilized, and generalized to temporal domains.

\xhdr{Heuristic Imagination}
Latent perception also enables VLMs to perform visual heuristic imagination, generating and manipulating internal visual representations as part of the reasoning process, analogous to human mental imagery.
This capability is valuable for tasks requiring spatial reasoning, 3D understanding, or visual planning that cannot be adequately expressed in text.
Internal visual imagination~\cite{yang2025machine,chen2025think} empower latent manipulation and align VLM features with 3D foundation models, while visual scratchpads~\cite{zhang2025latentsketchpad,tong2025sketchinlatents,qin2025chainofvisualthought} introduce sketching mechanisms that capture dense spatial and geometric information through continuous visual tokens.
Perceptual fidelity preservation~\cite{wu2026lavit,ma2025multimodal} ensures that latent visual representations maintain fidelity during distillation and dynamically refocuses attention across modality gaps.

\xhdr{Faithful Grounding}
A critical application of latent perception is improving the faithfulness of VLM outputs by intervening at the representation level, addressing the pervasive problem of hallucination.
Hallucination mitigation via latent steering and architectural alignment~\cite{liu2025reducing,ahmed2025alignvlm} corrects visual-textual misalignments and addresses root causes, while perceptual grounding tokens~\cite{perception2025bigverdi} provide auxiliary signals through depth maps and detection outputs.
Representation analysis and diagnostics~\cite{berasi2025not,benno2026latentlens,yao2025reading,xu2026thinking} reveal when perceptual failures occur and support calibrated uncertainty estimation.
Domain-specific deployment in industrial and video anomaly detection~\cite{chen2026reason,cai2026steering} demonstrates practical reliability improvements. These methods collectively establish latent perception as a powerful mechanism for reducing the gap between what models see and what they report, with direct implications for the reliability of deployed vision-language systems.

\subsection{Memory}
\label{sec:memory}
Memory has emerged as a necessary complement to LLMs, whose stateless architecture needs external mechanisms to retain knowledge across inference steps~\cite{hu2025memory}. Yet token-based memory introduces its own bottleneck: representing accumulated context as discrete sequences inflates prompt length, degrades retrieval fidelity, and prevents the gradient-based optimization needed for adaptive memory consolidation. Latent memory resolves this by encoding persistent knowledge as continuous vectors, enabling compact cross-context retention with superior fidelity and adaptability. We organize latent memory into three progressively broader abilities: \textbf{Working Retention} for cache intervention, \textbf{Persistent Mind} evolution for self-evolving knowledge stores, and \textbf{Multimodal Recall} grounding across visual and embodied modalities.

\xhdr{Working Retention}
Continuous latent representations transform the KV cache from a passive record into an actively managed working memory that can be augmented, compressed, and consolidated far beyond what discrete token sequences allow. Differentiable cache injection and selective token retention schemes~\cite{liu2025deliberation,wu2025efficient} demonstrate that models can deliberate asynchronously and scale to longer sequences without proportional memory growth, while low-rank and cross-layer compression~\cite{mu2025sals,hao2026deltakv} confirm that latent key-vector structure can be exploited to reduce cache footprint without sacrificing representational fidelity. Intrinsic consolidation~\cite{hou2026flashmem} further shows that working memory can be synthesized directly from the model's own transient reasoning states, where uncertainty-driven triggering eliminates the need for auxiliary encoders.

\xhdr{Persistent Mind}
Latent representations unlock a qualitatively different memory regime where knowledge stores persist across context resets, update selectively, and differentiate into specialized functions through experience alone. Gated and generative approaches~\cite{xu2026gmemllm,zhang2025memgen,zhang2026nextmem} establish that latent slots can be durably maintained through differentiable selective writing while being dynamically synthesized on demand, with planning, procedural, and working memory faculties emerging without explicit cognitive supervision. Self-evolving and retrieval-unified methods~\cite{zhang2025latentevolve,he2025clara} further demonstrate that these stores improve across queries through dual-phase episodic and procedural consolidation, and can collapse external document retrieval into the same continuous space as generation, enabling end-to-end optimization; this persistent memory regime extends naturally to multi-agent settings~\cite{fu2026latentmem}, where role-conditioned composition resolves homogenization and token-overhead bottlenecks inherent to shared context.

\xhdr{Multimodal Recall}
Latent space imposes a structural memory barrier for visual and embodied agents: spatial layout, widget details, and temporal continuity are lost in conversion, rendering token-based memory incapable of sustaining perceptual grounding during extended generation~\cite{yu2025vismem,wu2025towards}. Continuous encoding methods~\cite{wu2025towards,wu2025autoscaling} prove that VLMs can compress multimodal knowledge into fixed-length embeddings compatible with frozen backbones. Unlike text-based memory under long prompts, these methods scale performance monotonically with memory depth. Cognitively structured variants~\cite{yu2025vismem} further reveal that organizing latent stores into complementary perceptual and semantic modules prevents systematic drift, establishing continuous latent memory as the essential substrate for anchoring complex tasks.

\subsection{Collaboration}
\label{sec:collaboration}
Collective intelligence in agent systems has traditionally been mediated by natural language~\cite{he2025llm}. Yet language constitutes an inherent bottleneck: compressing internal representations into discrete tokens discards semantic nuance, increases communication latency, and breaks the gradient pathways required for joint optimization~\cite{zou2025latent,fu2025cache}. Latent collaboration addresses these limitations by enabling agents to exchange continuous representations, preserving richer internal states and supporting a more expressive form of collective collaboration.
We organize latent collaboration into three progressively broader abilities:
\textbf{Semantic Fidelity} for lossless inter-agent state transfer via latent channels, \textbf{Shared Cognition} for identifying and evolving shared latent thought structures across agents, and \textbf{Heterogeneous Interoperability} for extending latent
collaboration across diverse model families and modalities without architectural coupling.

\xhdr{Semantic Fidelity}
Continuous representations unlock the most structurally immediate advance in multi-agent collaboration: replacing token-based message passing with direct latent state transfer that preserves the full semantic content of each agent's internal representations. Work on KV-cache and hidden-state communication~\cite{fu2025cache,du2025enabling,zou2025latent}
demonstrates that agents can exchange internal states without intermediate decoding, with theoretical analyses confirming that it has strictly higher expressiveness and lower complexity than text-based counterparts.
Complementary alignment approaches~\cite{dery2026latent} further show that a shared latent space can be learned across heterogeneous models, establishing a unified high-bandwidth collaboration channel without modifying any pre-trained parameters.

\xhdr{Shared Cognition}
Latent representations further make the structure of shared cognition between agents identifiable and continuously evolvable, a capability that text communication fundamentally lacks. Formal latent variable analyses~\cite{zheng2025thought,yu2026dual,fu2026latentmem} prove that shared and private thought components between agents can be identified from observable outputs nonparametrically, with the global topology of thought-sharing relationships also theoretically recoverable. Strategy evolution methods~\cite{tang2025learning} demonstrate that agents can update collaborative strategies by reflecting on text embeddings and propagating these reflections into external latent vectors, where stable, disentangled strategic styles emerge over long horizons without model fine-tuning.

\xhdr{Heterogeneous Interoperability}
Latent space also enables coordination across agents of different architectures, specializations, and modalities through a shared continuous substrate rather than task-specific natural language protocols. Agent Primitives~\cite{wang2026primitives} show that recurring MAS interaction patterns can be abstracted into reusable latent building blocks that generalize across tasks without manual role engineering. Visual latent frameworks~\cite{yu2026dual,liu2026vision} further demonstrate that perceptual and reasoning trajectories in multimodal systems can be decoupled into complementary latent memories to overcome the performance-degrading scaling wall of text-centric communication, and that the visual interface of VLMs can serve as a model-agnostic port for injecting heterogeneous reasoning traces directly into a receiver's perceptual pathway, enabling training-free cross-family collaboration without pair-specific translators.

\subsection{Embodiment}
\label{sec:embodiment}
Embodied agents confront a data bottleneck that no purely linguistic domain faces as acutely: every increment in physical diversity, \textit{e.g.}, new hardware morphologies, viewpoints, and task environments, invalidates existing labeled demonstrations and forces platform-specific re-collection that does not transfer~\cite{liu2025embodied,ni2025swiftvla,bu2025univla}. Action in latent space compounds this by severing the continuous geometric and causal structure that manipulation requires, collapsing spatially rich dynamics into a symbolic bottleneck that discards depth, temporal continuity, and cross-embodiment correspondence. Latent representations dissolve all three failure modes simultaneously, enabling action semantics to emerge from unlabeled video, deliberate reasoning to be internalized as continuous state trajectories, and spatial priors to be distilled directly into policy backbones without instrumentation or re-annotation. We organize latent embodiment into five progressively broader abilities: \textbf{Unsupervised Grounding} for deriving transferable action representations from unlabeled video without embodiment-specific labels, \textbf{Implicit Thinking} for internalizing multi-step planning as continuous latent computation without explicit chain-of-thought generation, \textbf{Predictive Foresight} for simulating future states in latent space to generate dense training signals and guide real-time decision-making, \textbf{Spatial Cognition} for reconstructing 3D/4D geometric structure from 2D observations within the policy latent space, and \textbf{Generalized Transfer} for bridging heterogeneous hardware morphologies through a shared body-agnostic latent substrate.

\xhdr{Unsupervised Grounding}
The most consequential advantage latent representations confer on embodied AI is the ability to ground action semantics from internet-scale video without any teleoperation labels, converting the scarcity of robot demonstrations from a structural ceiling into a surmountable bottleneck.
Quantization and continuous codebook approaches establish that inter-frame visual transitions can be compressed into latent action tokens that generalize across embodiments and outperform label-supervised models in low-data regimes~\cite{ye2024lapa,bu2025univla,tharwat2025latent}, while grounding fidelity improves further when latent objectives are constrained by physical trajectory priors, contrastive proprioceptive alignment, or disentangled structure-and-motion decomposition~\cite{li2025latbot,zhang2026clap,dai2026conla,jala2026joint,fx2026cowvla,bi2025motus}. Spatial and temporal structure serve as auxiliary grounding signals that sharpen action-relevance and suppress task-irrelevant distractors, as confirmed by frameworks
that jointly address geometric and temporal bottlenecks~\cite{cai2025seeing} and by pretraining pipelines that interleave cross-viewpoint alignment with latent action learning~\cite{jeong2026learning,jiang2025wholebodyvla}, collectively demonstrating that the quality of the latent action space, not the quantity of labeled data, is the primary determinant of downstream manipulation generalization.

\xhdr{Implicit Thinking}
Continuous representations enable a qualitatively different mode of embodied cognition: replacing the latency-dominated, linguistically bottlenecked chain-of-thought with compact latent trajectories that carry multi-step deliberation directly into the action generation pathway.
Reinforcement-learned visual plan latents and their distilled
successors~\cite{huang2025thinkact,huang2026fastthinkact} show that embodied reasoning can be grounded in action-aligned visual rewards and compressed into a handful of continuous tokens, achieving long-horizon planning and few-shot adaptation at a fraction of the
inference cost of textual alternatives. Curriculum-based approaches that progressively internalize explicit chain-of-thought
supervision into pure latent computation~\cite{bai2026latent}, architectures that iterate action predictions through recurrent latent refinement at constant memory~\cite{tur2026recurrent}, and frameworks that unify visual dynamics, spatial priors, and proprioceptive states within a single token-efficient reasoning space~\cite{liu2026last0,ma2025unifying,luo2026last,tan2025latent,peng2025colavla,xie2026latentvla} collectively establish that latent reasoning is not merely more efficient than symbolic alternatives but strictly more expressive for the continuous, spatial domain of physical control.

\xhdr{Predictive Foresight}
Latent representations uniquely enable embodied agents to simulate future states without generating pixels, allowing imagined outcomes to serve as training supervision and real-time decision guidance rather than expensive auxiliary outputs. JEPA-style pretraining~\cite{ginwind2026vlajepa} demonstrates that predicting target-encoder latents of future frames in a leakage-free regime yields dynamics abstractions robust to camera motion and background variation, while world-model-derived latent distances~\cite{fei2025srpo} provide dense progress rewards that resolve the sparsity bottleneck of VLA reinforcement learning, achieving near-complete task mastery within two hundred RL steps from a sparse-reward baseline. Frameworks that generate foresight and action within the same latent autoregressive pass~\cite{huang2025thinkact,fan2026future} further demonstrate that action-aligned visual rewards and reviewable visual look-aheads can be produced in a single forward pass, establishing latent-space future simulation as a training and deployment mechanism that discrete token prediction fundamentally cannot replicate.

\xhdr{Spatial Cognition}
Physical manipulation imposes a 3D geometric demand on policies trained from 2D observations, and latent representations are uniquely positioned to reconstruct spatial structure without requiring explicit depth sensors or 3D annotations. Knowledge distillation of frozen geometry-aware encoders into LLM visual token representations~\cite{guo2025glad,govind2026unilact,ni2025swiftvla} demonstrates that aligning policy latents with 3D and 4D structural features injects spatial priors into the backbone without architectural modification, enabling precise contact-rich manipulation where appearance-only latents systematically fail. Treating dense 3D occupancy as both a latent predictive output and a supervisory signal~\cite{liu2025occvla} shows that volumetric spatial awareness can be developed from auto-annotated 2D data alone, while jointly addressing geometric and temporal grounding bottlenecks~\cite{cai2025seeing,jeong2026learning} confirms that geometry-aware encoding and cross-viewpoint alignment are complementary and mutually necessary, establishing latent spatial imagination as the representational prerequisite for precise physical interaction under realistic sensor constraints.

\xhdr{Generalized Transfer}
Cross-hardware deployment is the structural bottleneck that prevents generalist embodied intelligence from scaling: every morphological change invalidates action spaces and demands platform-specific retraining that cannot amortize across the growing diversity of robotic hardware. Latent action spaces resolve this by functioning as body-agnostic abstraction layers where semantically equivalent motions from heterogeneous embodiments converge, enabling zero-shot cross-platform deployment and data-efficient adaptation without access to
task rewards or target-platform demonstrations~\cite{zhang2025alignthensteer,shi2026care,jiang2025wholebodyvla}. Language-action disentanglement via Bayesian decomposition~\cite{lian2026langforce} and
state-aware latent re-representation~\cite{wang2025lola} further demonstrate that the same latent substrate simultaneously resolves instruction-following collapse and supports multi-modal cross-task generalization, while navigation-oriented latent alignment~\cite{subedi2026lcla} confirms that the body-agnostic latent principle extends from manipulation to embodied navigation, establishing continuous latent action spaces as the necessary foundation for embodied intelligence that must generalize across hardware morphology, task distribution, and physical environment simultaneously.

\section{\textcolor{secgray}{Outlook:} What is Next?}
\label{sec:outlook}
The preceding sections survey latent space in large models from multiple angles: its foundational definition and comparison with explicit space, its evolutionary trajectory from early exploration to a full-fledged research paradigm, the technical mechanisms spanning architecture, representation, computation, and optimization that govern latent processing, and the diverse abilities it unlocks across reasoning, planning, modeling, perception, memory, collaboration, and embodiment. Together, these advances demonstrate both the breadth and the momentum of the latent-space paradigm, while also revealing structural limitations and open questions. This section synthesizes these observations into a set of perspectives, challenges, and future directions.

\subsection{Perspective}

The rise of latent space signals a fundamental reorientation in the study of language-based intelligence. Rather than treating latent space as a incidental byproduct of computation, recent research increasingly positions them as a primary substrate. In this perspective, we formulate this survey in four sequential lenses: \textbf{Foundation} (Section~\ref{sec:foundation_sec2}) intrinsically clarifies what latent space is; \textbf{Evolution} (Section~\ref{sec:evolution}) reveals how it has grown from an emerging idea into a broad research paradigm; \textbf{Mechanism} (Section~\ref{sec:mechanism}) explains through what technical designs is realized; and \textbf{Ability} (Section~\ref{sec:ability}) shows what it enables within the latent space. Viewed as a whole, these dimensions suggest that latent space is a promising candidate basis for the next generation of general-purpose intelligent systems.

\xhdr{Foundation}
From a foundational standpoint, latent space should be understood not merely as an auxiliary hidden representation, but as a machine-native space that redefines where language-based autoregressive models compute with the semantic information~\cite{hu2024understanding,xu2025formal,zou2026capabilities}. Relative to explicit space (or verbal space)~\cite{xi2025rise,hu2025landscape,wu2025evolutionary,humze2025comprehensive}, its central value lies in four representational properties: \textbf{machine-native}, \textbf{continuous}, \textbf{flexible \& efficient}, and \textbf{high-fidelity}, and four functional capabilities: \textbf{operable}, \textbf{expressive}, \textbf{scalability}, and \textbf{generalized}, which reduces the redundancy, discretization bottleneck, inefficiency, and semantic loss inherent in verbalized computation. In this sense, the latent paradigm marks a shift from human-aligned generation to machine-optimal computation. At the same time, this shift introduces a constitutive tension: the gains in efficiency, expressiveness. Thus, the foundational significance of latent space is not only technical but epistemic: it changes both what models compute and how that computation can be understood, laying the groundwork for next-generation models to transcend token-centric operation.

\xhdr{Evolution}
The evolutionary trajectory of latent-space research suggests that the field is moving from sprout toward \textbf{all-round outbreak}, in terms of not only the number of the works but also the diversity of paradigms. The early stage established feasibility by showing that reasoning-relevant structure already resides in internal activations and can, in part, bypass explicit verbalization. The subsequent foundation stage supplied theoretical justification, and initial multimodal extensions, thereby converting isolated demonstrations into a coherent research agenda. The later expansion and outbreak stages reveal a further transition: latent space is no longer treated as a compression trick for textual reasoning alone~\cite{reasoning2025chen,implicit2025li,survey2025zhu}, but as a general framework spanning visual cognition, memory, collaboration, and embodied action. This progression indicates that latent space is best viewed not as a transient optimization of autoregressive models, but as an emerging systems principle for next-generation general intelligence.

\xhdr{Mechanism}
From a mechanistic perspective, the survey implies that progress in latent space is driven by the co-design of four interdependent dimensions: \textbf{architecture}, \textbf{representation}, \textbf{computation}, and \textbf{optimization}. The key issue is not simply whether a model contains latent variables, but \textit{what architectures are utilized}, \textit{what representations are instantiated}, \textit{how computation is operated through them}, and \textit{at how they are optimized}. This taxonomy reveals an important trend: the field is gradually moving from heuristic usage toward systematic principle, from externally integrated to internally enabled, from static and fixed to dynamic and adaptive, and from conventional designs to multiple paradigms. Accordingly, future research is less about adding latent space to existing models than about designing models whose primary core is intrinsically latent.

\xhdr{Ability}
In terms of ability, the most consequential contribution of latent space is that it broadens the functional scope of intelligence beyond explicit-space models. The survey shows that latent space of language-based models supports not only enhanced \textbf{reasoning}, but also \textbf{planning}, \textbf{modeling}, \textbf{perception}, \textbf{memory}, \textbf{communication}, and \textbf{embodiment}. What unifies these capacities is that they all require structures that are difficult, costly, or fundamentally lossy to externalize in natural language. Latent space therefore acts as a common substrate for general integration across modalities, timescales, and agents. Under this view, its ultimate promise lies not in replacing explicit textual tokens, but in enabling models to coordinate heterogeneous forms of information within a shared continuous latent space. The strongest long-term implication is that latent space may become the principal medium of general-purpose models.

Overall, the perspective presented in this survey about latent space suggests that from its foundational properties to its historical expansion, from its underlying mechanisms to its emerging abilities, latent space consistently points toward a common conclusion: future intelligent systems may rely increasingly on latent rather than purely verbal computation as their primary operating principle.

\subsection{Challenge}
\label{sec:challenges}

Despite its growing promise as a machine-native substrate for computation, latent space still faces several fundamental obstacles before it can serve as a reliable foundation for general-purpose intelligent systems. The same properties that make latent representations powerful, their continuity, compression, flexibility, and expressive capacity, also make them difficult to inspect, assess, and govern, resulting relatively low \textbf{Evaluability}, \textbf{Controllability}, and \textbf{Interpretability}. Together, these issues reveal that progress in latent space depends not only on improving capability, but also on making latent computation more observable, steerable, and understandable.

\xhdr{Evaluability} 
A central challenge for latent-space reasoning methods lies in their limited evaluability. Unlike explicit reasoning traces~\cite{miles2023language,temera2023measuring}, latent trajectories are not directly accessible to human inspection, which makes it inherently difficult to determine whether an intermediate computation is correct, complete, or even relevant to the task at hand~\cite{du2025latent,li2026dynamics,zou2026capabilities}. This opacity substantially constrains process-level verification: researchers are often unable to distinguish between genuinely structured intermediate reasoning and representations that merely correlate with the correct output. Consequently, the evaluation of latent reasoning systems still relies predominantly on final-answer accuracy or on post hoc verbalization, both of which offer only indirect and potentially incomplete evidence about the actual reasoning process~\cite{zhang2025do,implicit2025li}.

Although some recent studies have begun to propose benchmarking strategies for latent-space reasoning~\cite{hagendorff2025beyond}, the field still lacks mature and widely accepted protocols for supervision and evaluation~\cite{zou2026capabilities,sahoo2026shallow}. Existing approaches remain fragmented across tasks, datasets, and measurement criteria, and there is as yet no standardized framework for assessing the faithfulness, robustness, or internal consistency of latent reasoning trajectories. Such benchmark fragmentation and metric inconsistency make fair comparison across methods difficult, hinder cumulative progress, and complicate the identification of genuine methodological advances. As a result, improving evaluability remains one of the most pressing open problems for the development of latent reasoning models.

\xhdr{Controllability} 
Although latent space is, in principle, a highly operable substrate for computation and control, achieving reliable and generalizable manipulation of latent representations remains a substantial challenge in practice~\cite{zhang2024survey,lu2025latent,survey2025zhu}. Fine-grained interventions on hidden states can indeed reshape model behavior in useful and sometimes remarkably precise ways; however, such interventions often suffer from relatively low controllability. The central difficulty lies not merely in identifying where and how to intervene, but in determining how high-level semantic intentions should be specified so that they are simultaneously machine-actionable, sufficiently precise, and intelligible to human operators~\cite{kyle2024steering,andy2023representation,sadiekh2025polarity}. This tension exposes a deeper gap between continuous internal dynamics and discrete, interpretable objectives. Consequently, the development of truly controllable latent systems requires more than local steering techniques alone: it calls for principled mechanisms that can map explicit goals, safety requirements, and resource constraints onto internal computational processes in a robust and adaptive manner.

\xhdr{Interpretability}
The difficulty lies in the very nature of latent representations: they are high-dimensional, distributed, and often deeply entangled, so that neither individual dimensions nor their induced trajectories map neatly onto stable semantic concepts~\cite{andy2023representation,van2017neural,yang2025internal}. What emerges is a representational space of immense expressive power, but one whose internal organization is resistant to human understanding. This opacity makes it challenging to explain why a model reaches a particular conclusion, to trace how information is transformed across successive stages of computation, or to determine where error and misalignment first arise~\cite{zou2026capabilities,berasi2025not,sahoo2026shallow,li2026imagination}. More importantly, the problem is not simply epistemic but institutional: systems that reason through inscrutable latent operations may become more powerful while simultaneously becoming less auditable, less diagnosable, and less accountable. Thus, there is still rooms for the interpretability study.

\subsection{Future}

Looking ahead, the next decisive step for latent space research is not merely to produce better latent reasoning methods, but to establish latent space as the native substrate of machine intelligence. What is emerging from current progress is a broader architectural shift: explicit language may remain the interface for instruction, generation, and verification, while latent space increasingly becomes the internal workspace where models think, understand, simulate, remember, and act. In this sense, the future of latent space is not simply about improving efficiency, but about redefining where general-purpose operations happens.

\xhdr{Theory}
A central priority for future research is to move beyond empirical success toward a principled theoretical understanding of latent-space intelligence~\cite{hao2024training}. Rather than merely showing that latent works in practice, the field must explain how and why latent spaces support computation, under what conditions they outperform explicit token-level space, and what forms of reasoning are genuinely native to latent space. In this sense, the key question is not simply whether latent reasoning is more efficient than verbalized reasoning, but when it is more operable, more expressive, more scalable, and more generalizable as a computational substrate. More broadly, the field must progress from scattered theoretical validation to a unified framework for latent representation, latent computation, and the capability gains they enable.

There is therefore an indispensable need for a \textbf{foundational theory} of latent space. Existing work has begun to provide formal accounts of its expressiveness and computational advantages~\cite{hagendorff2025beyond,zhu2025reasoning,yang2025internal}, but a systematic theory remains underdeveloped~\cite{xu2025formal,zhang2025do,zou2026capabilities,liang2026do,sahoo2026shallow}. Future research should clarify not only how latent reasoning operates, but also \textit{why}, \textit{when}, and under \textit{what} constraints it surpasses explicit or verbal reasoning. This requires a deeper account of the geometry, semantics, optimization dynamics, and computational organization of latent trajectories, so that latent reasoning can be understood as more than an effective engineering technique.

Equally important is a principled formulation of explicit and latent spaces as two \textbf{complementary representational regimes} within language-based intelligence~\cite{survey2025zhu,giorgos2025language,xu2025formal,zou2026capabilities}. The central issue is not whether latent computation will replace language, but how symbolic and continuous representations differ in expressive capacity, computational function, and communicative role. Within such a framework, explicit language may remain the externally accessible interface for instruction, generation, and verification, while latent space serves as the internal workspace for reasoning, abstraction, memory, simulation, and planning. The theoretical goal is to explain \textit{how} these two spaces interact, \textit{what} kinds of information and operations are most naturally allocated to each, and \textit{what} trade-offs emerge in their coordination.

Future work must also develop a principled theory of \textbf{trustworthy} latent space. Because latent trajectories are inherently opaque to direct human inspection, progress cannot rely on outcome-level performance alone~\cite{hao2024training,implicit2025li,survey2025zhu}. Instead, the field needs rigorous frameworks for evaluability, controllability, and interpretability, together with standardized benchmarks and supervision protocols, to assess the faithfulness, robustness, and internal consistency of latent computation. Without such foundations, benchmark fragmentation and metric inconsistency will continue to obstruct fair comparison and cumulative progress.

Ultimately, the future of latent-space research lies in transforming latent space from an empirical method into a principled theory: one that formally explains its computational advantages, formalizes its relation to explicit space, and renders its hidden processes evaluable, controllable, and interpretable.

\xhdr{Multimodal}
The survey suggests that the future of multimodal intelligence, \textit{e.g.}, visual models and embodied models, should not be understood as the mere addition of more sensory channels to language models. Rather, the more consequential shift is the emergence of latent space as a shared computational workspace, in which language, vision, action, memory, and inter-agent communication can be jointly processed within continuous representations. In this view, latent space is no longer a secondary device for compressing reasoning traces, but a machine-native substrate that supports the internal organization of general-purpose intelligence across modalities and timescales.

This perspective implies a first major direction for future research: the transition from text-mediated multimodality to \textbf{modality-native multimodality} latent computation. Existing multimodal systems often rely on translating non-linguistic inputs into textual or tokenized forms before higher-level operation takes place~\cite{yin2023survey,tadas2019multimodal,liu2025embodied}. By contrast, the survey highlights that latent representations can preserve high-fidelity information, reduce modality gaps, and support smoother fusion, alignment, and interaction than explicit symbolic channels. The long-term goal, therefore, is not simply to describe perception in language, but to enable models to operate within visual, spatial, and action-oriented latent spaces directly.

A second direction concerns the evolution from isolated multimodal  models to \textbf{integrated multimodal systems}~\cite{ma2025multimodal,li2025latent,li2025latent1,yu2025vismem}.Across its ability-oriented taxonomy, the survey emphasizes that latent space enables not only reasoning, but also planning, modeling, perception, memory, collaboration, and embodiment. What unifies these capabilities is that they all involve structures that are difficult, costly, or lossy to externalize in natural language. For this reason, the future of multimodal AI is likely to depend on architectures in which perception, world modeling, memory formation, communication, and action planning are coordinated through a common latent substrate rather than through loosely coupled modules or additional calculations.

Overall, the survey points toward a broader conclusion: the future of multimodal model lies not in expanding token-centric generation to more modalities, but in establishing a unified latent computational substrate for cross-modal reasoning, memory, communication, and embodied interaction. Under this framework, explicit language is likely to remain the interface for instruction, reporting, and verification, whereas latent space increasingly becomes the internal medium in which models think, simulate, coordinate, and act.

\xhdr{Downstream Task}
Future downstream progress should be understood as a broader shift in which explicit language remains the interface for instruction, supervision, and verification, while latent space increasingly serves as the internal workspace for computation. Under this view, the most promising downstream tasks are those whose \textbf{intermediate states} are poorly captured by discrete verbal traces, including search-intensive reasoning~\cite{survey2025zhu,hao2024training,implicit2025li}, sequential planning~\cite{hagendorff2025beyond,yang2025internal}, visual perception~\cite{yin2023survey,yin2023survey}, long-horizon memory~\cite{hu2025memory}, multi-agent coordination~\cite{he2025llm}, and embodied control~\cite{liu2025embodied}. Across these settings, latent representations offer a unified substrate for compact search, continuous refinement, richer visual grounding, persistent memory, higher-bandwidth coordination, and transferable action abstractions; accordingly, future systems are likely to perform most internal inference, planning, and state evolution in latent space, and externalize only final outputs or strategically selected checkpoints. More broadly, the survey suggests that the central opportunity of latent space lies in moving from text-centric problem solving toward persistent internal state management across heterogeneous tasks and modalities, although realizing this promise will require principled progress in evaluation, control, interpretability, and interface standardization.

\xhdr{Governable}
A promising future direction is to develop latent space into an observable and governable substrate. Although latent representations are continuous, compact, and expressive, these same properties also make them difficult to evaluate, control, and interpret. Future work should therefore go beyond improving latent reasoning accuracy, and instead establish a full methodology for \textbf{trustworthy} latent computation: benchmark suites that assess the faithfulness and robustness of latent trajectories; \textbf{supervision} strategies that provide process-level signals rather than only final-answer feedback; controllable latent interfaces that align internal dynamics with explicit goals, resource budgets, and safety requirements; and \textbf{explainable} frameworks that identify semantic structure, causal pathways, and failure sources in latent representations. Such efforts would help bridge the gap between the efficiency of machine-native computation and the need for human oversight, potentially enabling latent space to serve as a dependable substrate.

\section{Conclusion}
In this survey, we have presented a systematic review of latent space in language-based models from five complementary perspectives: foundation, evolution, mechanism, ability, and outlook. Taken together, these perspectives suggest that latent space should be a substrate that may fundamentally reshape how intelligent language models deal with diverse information.  We further show that the development of this field has rapidly progressed from early explorations of latent reasoning to a broader and increasingly unified research paradigm spanning language, vision, memory, collaboration, and embodied action.

To systematically organize this promising landscape, we propose a taxonomy along two orthogonal axes: mechanism orientation and ability orientation. On the axis of \textbf{Mechanism}, we classify four key types: architecture, representation, computation, and optimization, which defines how latent space is operationalized. On the axis of \textbf{Ability}, we expand the single type in previous surveys to seven main functional categories: reasoning, planning, modeling, perception, memory, collaboration, and embodiment. Across these dimensions, a consistent trend becomes visible: Latent space brings a fundamental transformation to model mechanisms while pushing the boundaries of model capabilities.

At the same time, the promise of latent space must be considered together with its unresolved challenges. As increasingly more cognition is internalized into continuous hidden computation, the resulting processes become harder to evaluate, control, and interpret. Future progress will therefore depend not only on improving empirical performance, but also on establishing stronger theoretical foundations, more reliable benchmarks and supervision protocols, and more transparent as well as controllable latent mechanisms. Overall, the central conclusion of this survey is that latent space holds the potential to become a foundational principle for language-based models. We hope that this survey offers a coherent foundation for future research and serves as a valuable reference for future researchers.


\nocite{xu2025learning,lin2025identity,yan2026breaking,peng2026measuring,lys2026inner,bu2025laof,jiang2025dart}
\bibliographystyle{assets/plainnat}
\bibliography{citation}

\end{document}